\documentclass[journal]{IEEEtran}
\ifCLASSOPTIONcompsoc
  \usepackage[nocompress]{cite}
\else
  \usepackage{cite}
\fi

\usepackage[x11names]{xcolor}
\usepackage{times,graphics,epsfig,amsmath,xspace,endnotes,pifont,multirow,rotating,listings,amssymb,color,nicefrac,adjustbox,tabularx,mathtools}
\usepackage{amssymb,amsfonts}
\usepackage{graphicx}
\usepackage{subfig}
\usepackage{textcomp}
\usepackage{xcolor}
\usepackage{enumitem}   
\usepackage{dsfont}
\usepackage{tikz}
\usetikzlibrary{decorations.pathreplacing}
\usetikzlibrary{shapes.symbols}
\usepackage{todonotes}
\usepackage{siunitx}
\usepackage{booktabs}
\usepackage{makecell}
\usepackage{soul}
\usepackage{algorithm}
\usepackage{algpseudocode}

\usepackage{gnuplot-lua-tikz} % for gnuplot tikz term

\usepackage{pgfplots}

\usetikzlibrary{calc,shapes.geometric,arrows,positioning,intersections,math}
\usetikzlibrary{fit}
\def\BibTeX{{\rm B\kern-.05em{\sc i\kern-.025em b}\kern-.08em
    T\kern-.1667em\lower.7ex\hbox{E}\kern-.125emX}}
    
  \usepackage{enumitem}   
\usepackage{diagbox}
\ifCLASSINFOpdf
\else
\fi
\DeclareMathOperator{\erf}{erf}
\DeclareMathOperator*{\argmin}{argmin} % no space, limits underneath in displays
\DeclareMathOperator*{\argmax}{argmax} % no space, limits underneath in displays

\usepackage[savewrites,nopostdot,toc,acronym,symbols,nogroupskip]{glossaries-extra}
\usepackage{changes}
\setabbreviationstyle[acronym]{long-short}% glossaries-extra.sty

\usepackage{glossaries}
\newacronym{3gpp}{3GPP}{Third Generation Partnership Project}
\newacronym{4g}{4G}{4\textsuperscript{th} generation}
\newacronym{5g}{5G}{5\textsuperscript{th} generation}
\newacronym{6g}{6G}{6\textsuperscript{th} generation}
\newacronym{ai}{AI}{Artificial Intelligence}
\newacronym{a2c}{A2C}{Advantage Actor-Critic}
\newacronym{aoi}{AoI}{Age of Information}
\newacronym{fgan}{FGAN}{Focus Group on Autonomous Networks}
\newacronym{arima}{ARIMA}{AutoRegressive Integrated Moving Average}
\newacronym{bs}{BS}{Base Station}
\newacronym{cots}{COTS}{Custom-Off-The-Shelf}
\newacronym{cpu}{CPU}{Central Processing Unit}
\newacronym{cv2n}{C-V2N}{Cellular Vehicle-to-Network}
\newacronym{cnn}{CNN}{Convolutional Neural Network}
\newacronym{ddpg}{DDPG}{Deep Deterministic Policy Gradient}
\newacronym{dhpg}{DHPG}{Deep Hybrid Policy Gradient}
\newacronym{dod}{DOD}{Deterministic Ordered Discretization}
\newacronym{dz}{DZ}{Dead Zone}
\newacronym{drl}{DRL}{Deep Reinforcement Learning}
\newacronym{dqn}{DQN}{Deep Q-Network}
\newacronym{ecdf}{eCDF}{Empirical Cumulative Distribution Function}
\newacronym{elu}{ELU}{Exponential Linear Unit}
\newacronym{epdf}{ePDF}{Empirical Probability Density Function}
\newacronym{epmf}{ePMF}{Empirical Probability Mass Function}
\newacronym{etsi}{ETSI}{European Telecommunications Standards Institute} \newacronym{eni}{ENI}{Experiential Network Intelligence}
\newacronym{ga}{GA}{Genetic Algorithm}
\newacronym{gnb}{gNB}{Next-Generation Node B}
\newacronym{gnn}{GNN}{Graph Neural Network}
\newacronym{ilp}{ILP}{Integer Linear Programming}
\newacronym{isg}{ISG}{Industry Specification Group}
\newacronym{itu-t}{ITU-T}{International Telecommunication Union - Telecommunication standardization sector}
\newacronym{kpi}{KPI}{Key Performance Indicator}
\newacronym{lstm}{LSTM}{Long Short Term Memory}
\newacronym{mano}{MANO}{Management and Orchestration}
\newacronym{mdp}{MDP}{Markov Decision Process}
\newacronym{mec}{MEC}{Multi-access Edge Computing}
\newacronym{ml}{ML}{Machine Learning}
\newacronym{ng}{NG}{Next Generation}
\newacronym{nfv}{NFV}{Network Function Virtualization}
\newacronym{nfvi}{NFVI}{NFV Infrastructure}
\newacronym{nn}{NN}{Neural Network}
\newacronym{pi}{PI}{Proportional Integral}
\newacronym[longplural={Points of Presence}]{pop}{PoP}{Point of Presence}
\newacronym{ppo}{PPO}{Proximal Policy Optimization}
\newacronym{qoe}{QoE}{Quality of Experience}
\newacronym{ran}{RAN}{Radio Access Network}
\newacronym{rl}{RL}{Reinforcement Learning}
\newacronym{rt}{RT}{Real-Time}
\newacronym{ru}{RU}{Radio Unit}
\newacronym{ric}{RIC}{Radio Intelligent Controller}
\newacronym{sarsa}{SARSA}{State–Action–Reward–State–Action}
\newacronym{sla}{SLA}{Service Level Agreement}
\newacronym{sdn}{SDN}{Software-Defined Networking}
\newacronym{tes}{TES}{Triple Exponential Smoothing}
\newacronym{td}{TD}{Temporal Difference}
\newacronym{v2x}{V2X}{Vehicle-to-Everything}
\newacronym{v2n}{V2N}{Vehicle-to-Network}
\newacronym{v2v}{V2V}{Vehicle-to-Vehicle}
\newacronym{vnf}{VNF}{Virtualized Network Function}
\newacronym{zsm}{ZSM}{Zero touch network \& Service Management}

\usepackage{bm}
\usepackage{amsthm}
\newtheoremstyle{mystyle}%                % Name
  {}%                                     % Space above
  {}%                                     % Space below
  {\itshape}%                                     % Body font
  {}%                                     % Indent amount
  {}%                            % Theorem head font
  {.}%                                    % Punctuation after theorem head
  { }%                                    % Space after theorem head, ' ', or \newline
  {\thmname{{\bfseries#1}} \thmnumber{{\bfseries#2}}\thmnote{ (#3)}}%
\theoremstyle{mystyle}
\newtheorem{problem}{Problem}
\newtheorem{lemma}{Lemma}

\usepackage{url}
\usepackage{changes}
\begin{document}

%\copyrightstatement
%\title{Data-driven Models for Dynamic Placement and Scaling of V2X Services}
% \title{Dynamic Steering and Scaling for V2N traffic}
%\title{Multi-PoP V2N service provisioning:\\a comparison of existing solutions }
%\title{A Deep RL Approach on Task Placement and Scaling of Edge Resources for Cellular Vehicle-to-Network Service Provisioning}
%\title{\color{red}On Task Placement and Scaling of Edge Resources  for Cellular Vehicle-to-Network Service Provisioning}
\title{A Deep RL Approach on Task Placement and Scaling of Edge Resources for Cellular Vehicle-to-Network Service Provisioning}
\author{Cyril Shih-Huan Hsu, Jorge Mart\'in-P\'erez, Danny De Vleeschauwer,\\%Koteswararao Kondepu, 
Luca Valcarenghi, Xi Li, Chrysa Papagianni
\IEEEcompsocitemizethanks{\IEEEcompsocthanksitem Cyril Shih-Huan Hsu and
    Chrysa Papagianni are with University of Amsterdam\protect\\
    E-mail: \{s.h.hsu, c.papagianni\}@uva.nl
\IEEEcompsocthanksitem
Jorge Martín Pérez is with
Departamento de Ingeniería de Sistemas Telemáticos, Universidad Politécnica
de Madrid.\protect\\
E-mail: jorge.martin.perez@upm.es
\IEEEcompsocthanksitem Danny de Vleeschauwer is with Nokia Bell Labs\protect\\
    E-mail: danny.de\_vleeschauwer@nokia-bell-labs.com
%\IEEEcompsocthanksitem  Koteswararao Kondepu is with IIT Dharwad\protect\\
   % E-mail: k.kondepu@iitdh.ac.in
\IEEEcompsocthanksitem  Luca Valcarenghi is with Scuola Superiore Sant'Anna\protect\\
    E-mail: luca.valcarenghi@santannapisa.it
\IEEEcompsocthanksitem  Xi Li is with NEC Laboratories Europe\protect\\
    E-mail: Xi.Li@neclab.eu}% <-this % stops an unwanted space
\thanks{Manuscript sent in 2024}}

%{Shell \MakeLowercase{\textit{et al.}}: Bare Demo of IEEEtran.cls for Computer Society Journals}

\IEEEtitleabstractindextext{
\begin{abstract}

Cellular Vehicle-to-Everything (C-V2X) is currently at the forefront of the digital transformation of our society. By enabling vehicles to communicate with each other and with the traffic environment using cellular networks, we redefine transportation, improving road safety and transportation services, increasing the efficiency of vehicular traffic flows, and reducing environmental impact.
To effectively facilitate the provisioning of Cellular Vehicular-to-Network (C-V2N) services, we tackle the interdependent problems of service task placement and scaling of edge resources. Specifically, we formulate the joint problem and prove that it is not computationally tractable. 
%for provisioning Cellular Vehicular-to-Network (C-V2N) services, addressing the interdependent problems of service task placement and scaling of edge resources. We formalize the joint problem and prove its complexity.
 %To tackle the joint problem, we introduce Deep Hybrid Policy Gradient (\gls{dhpg}), a deep reinforcement learning approach for hybrid action spaces, addresses the interdependent challenges of task placement and resource scaling in C-V2N service provisioning, enabling holistic decision-making and improved performance.
%This paper introduces 
To address its complexity we propose \gls{dhpg}, a new Deep Reinforcement Learning (DRL) approach that operates in hybrid action spaces, enabling holistic decision-making and enhancing overall performance.
%to address the interdependent problems, enabling holistic decision-making and improved performance.
%We formalize the task placement and resource scaling as a joint problem and prove its complexity.
%We further analyze the theoretical complexity of the joint problem.
%We benchmark the performance of our approach, focusing on the scaling agent, against several State-of-the-Art (SoA) scaling approaches via simulations using a real C-V2N traffic data set. The results show that DDPG-based solutions outperform SoA solutions, keeping the latency experienced by the C-V2N service below the target delay while optimizing the use of computing resources. By conducting a complexity analysis, we prove that DDPG-based solutions achieve runtimes in the range of sub-milliseconds, meeting the strict latency requirements of C-V2N services.
%The performance of \gls{dhpg} is evaluated against several state-of-the-art (SoA) solutions through simulations employing a real-world C-V2N traffic dataset.
We evaluated the performance of DHPG using simulations with a real-world C-V2N traffic dataset, comparing it to several state-of-the-art (SoA) solutions.
DHPG outperforms these solutions, guaranteeing the $99^{th}$ percentile of C-V2N service delay target, while simultaneously optimizing the utilization of computing resources.
%{\color{blue}The results demonstrate that \gls{dhpg} outperforms SoA solutions by maintaining the $99^{th}$ percentile of C-V2N service latency within acceptable limits, while simultaneously optimizing the utilization of computing resources.}
Finally, time complexity analysis is conducted to verify that the proposed approach can support real-time C-V2N services. %Additionally, a complexity analysis is conducted to verify that \gls{dhpg} fulfills the stringent latency requirements of C-V2N services and achieves sub-millisecond runtime complexity.

\end{abstract}
}

\maketitle

\IEEEdisplaynontitleabstractindextext

% Note that keywords are not normally used for peer review papers.
%\begin{IEEEkeywords}
%v2X
%\end{IEEEkeywords}

\IEEEpeerreviewmaketitle

\section{Introduction} 
\label{sec:intro}

\IEEEPARstart{T}he term “Cellular Vehicle-to-Everything" (C-V2X) refers to the communication system that utilizes Long-Term Evolution (LTE) and/or 5G cellular technologies to facilitate seamless communication among vehicles, infrastructure, pedestrians, and other road users, establishing a comprehensive intelligent transportation ecosystem. The ``Everything" in C-V2X refers to  vehicles (V2V), infrastructure (V2I), pedestrians (V2P), and the network  (V2N)\cite{3GPPv2xrequirements, tutorial-v2x}. C-V2X technology's primary objective is to improve road safety, enhance transportation system efficiency, and elevate the overall driving experience \cite{rammohan2023revolutionizing}. Standardization of C-V2X technology is conducted by organizations like the 3rd Generation Partnership Project~\cite{3GPPV2X1} and corporate coalitions like the 5G Automotive Association (5GAA)\footnote{https://5gaa.org/about-us/}.

In particular, C-V2N communication allows the
C-V2X device to utilize the cellular network connection through the logical interface between the UE and the base station (Uu interface), to support applications/services such as advanced driver-assistance and collision avoidance; tele-operated driving; platooning (on highways); infotainment (e.g., video streaming to passengers in the vehicles); etc. 
Such applications, also require process-intensive and low-latency, reliable computing capabilities that the current  vehicles with limited computation resources cannot meet. In this sense, offloading of tasks in a vehicular environment, that is transferring a workload to resource-rich computing platforms on external servers, is a viable solution for resource-constrained vehicles \cite{AHMED20224135}. However, outsourcing to data centers can lead to increased delays, including transmission and processing delays due to CPU loads, as well as queuing delays from network congestion. In contrast, Edge Computing (EC) that provides computing resources and services at the edge of the
network, closer to the devices and applications, is an effective approach to overcome delay challenges. % and provide more reliable and secure services. 
Efficient employing resources across the edge-to-cloud continuum for C-V2X applications, is considered a basic requirement for evolving 6G-V2X to provide more user-aware, scalable, and low-latency services for vehicles \cite{noor20226g}.

In this paper, we consider a C-V2N  application supported by edge computing resources spanning multiple \glspl{pop} distributed throughout the metropolitan area of a city. We assume that application tasks per vehicle can be offloaded to any \gls{pop} which are interconnected throughout the city networks. %, which puts an additional load on the network and the processors.
In order to avoid  degradation in the C-V2N application performance % this offloading should be done wisely. The corresponding 
we need to address efficiently the resource allocation problem that consists of:
\vspace{-2pt}
\begin{enumerate}
\item \textit{Task Placement}: deciding where (i.e., in which \gls{pop}) to process the application tasks (e.g., decoding, sensor data processing, etc.) for each vehicle, given the availability of computing resources in each \gls{pop};
\item \textit{Scaling}: vertically scaling application resources  at each \gls{pop}, to support offloaded tasks in a cost-efficient manner.
\end{enumerate}

These decisions are interdependent. On one hand the placement of application tasks determines the %processing load and thus 
computing requirements per \gls{pop}, that drive scaling decisions. On the other hand scaling decisions define the maximum available computing resources per \gls{pop} that is further used as input for deciding on task placement. 
Moreover, the C-V2N application load fluctuates due to the
dynamic flow of vehicle traffic.
%Furthermore, in such a system the vehicle arrival process is dynamic. %cyclostationary. 
%Consequently, the C-V2N application load fluctuates, 
This leads to computing requirements per \gls{pop} that vary over time.
Furthermore, task placement decisions may not always depend on a vehicle's proximity to a \glspl{pop}. As long as delay requirements are met, it may be more cost-effective to process a vehicle's tasks at a \gls{pop} outside its immediate vicinity—particularly in cases where such a decision would lead to resource scaling.

In summary, the key challenges in C-V2N scenarios include: ($i$) the real-time processing requirements of C-V2N applications, often within milliseconds; ($ii$) the dynamic nature of application traffic, which fluctuates over time; and ($iii$) the large-scale network problem involving multiple \gls{pop}s and a high volume of vehicles moving through metropolitan areas. These challenges demand scalable and compute-efficient solutions to solve the joint problems in a near-optimal and time-efficient manner. However, traditional centralized optimization approaches are not practical for addressing them. One reason is that the workload and resource dynamics of \gls{pop}s cannot be updated in a timely and synchronized manner, especially in city-wide scenarios where \gls{pop}s are widely dispersed. Additionally, decisions made centrally may not be communicated or enforced at each \gls{pop} in time, depending on the status of signaling channels. Lastly, traditional optimization methods often struggle with large-scale problems due to the substantial computational efforts required.
Therefore, in this work we propose to leverage advanced AI methods to explore data driven solutions to address the above challenges.

%\tbd{Furthermore, it is almost impossible to the meet real time requirements of C-V2N applications using traditional, centralized optimization solutions. This is mainly due to the fact that: ($i$) the information regarding the workload and resource dynamics of the \glspl{pop} cannot be (centrally) updated in a timely and synchronized manner, especially considering a city-wide application scenario where the \glspl{pop} are fully dispersed throughout the city; ($ii$) centralized-computed decisions may not be communicated and enforced at each PoP on time, depending on the status of signalling channels, and ($iii$) traditional optimization techniques fail to address such problems at large scale.}

\subsection{Contributions}
The main contributions of this work are summarized below:
\begin{enumerate}
    \item We first formulate the task placement and resource scaling decisions as a joint optimization problem, considering the latency constraints of the C-V2N application and cost-efficiency, in terms of the employed computing resources. The problem is initially formulated with the assumption of perfect knowledge of all future vehicle arrivals.
    %    The problem is initially defined under the idealized assumption that all future vehicle arrivals are known. 
    We prove that the problem is $\mathcal{NP}$-hard and that the solution fosters meeting the $99^{th}$ percentile delay requirement.
    \item To account for the stochastic nature of traffic loads and the availability of physical resources, we formulate the joint problem as a~\gls{mdp}, assuming no prior information about future vehicle arrivals. We introduce a new Deep Reinforcement Learning (DRL) approach called Deep Hybrid Policy Gradient (DHPG), to support a hybrid action space that encompasses both discrete and continuous actions, accommodating the different placement and scaling decisions.
        %and propose a new Deep Reinforcement Learning (DRL) approach, termed Deep Hybrid Policy Gradient (DHPG), to support a hybrid action space that encompasses both discrete and continuous actions, accounting for the different placement and scaling decisions.
    %While learning a hybrid policy can be addressed by using multiple agents for the heterogeneous action spaces, a single agent with a hybrid action space is often preferred due to its efficiency, reduced coordination needs, and richer state representation.
    The proposed approach involves (i) a  state encoder that maps the high-dimensional joint state into a compact latent space, providing a noise-reduced representation with rich information;
    (ii) this joint representation is shared across specialized action heads, allowing each of them to make more holistic and well-informed decisions independently; (iii) we introduce the probability-as-action (PAA) approach, which allows for a differentiable representation of the hybrid action
    space, enabling actions from different actors to be optimized jointly with a single critic. This unified optimization results in faster convergence by ensuring more consistent gradient updates throughout the entire network.
    %This unified optimization leads to faster convergence, as more coherent gradient updates are achieved across the entire network. 
    
    %a joint state encoder that maps high-dimensional states into a compact, denoised latent representation. This shared representation is utilized by multiple specialized action heads to support informed and integrated decision-making. Additionally, we introduce the probability-as-action (PAA) approach, which enables a differentiable hybrid action representation that can be inherently optimized through a single critic, resulting in faster convergence and more consistent gradient updates.% This generalizable approach paves the way for future advancements in DRL with diverse action spaces.

    \item We compare the proposed DHPG approach to different SoA methods, introduced in \cite{noms, DeVleeschauwer2021} and \cite{v2n-access}. We further use the ``ideal" exact solution of the $\mathcal{NP}$-hard problem
    to derive optimality gaps for the selected approaches. To assess performance in realistic scenarios, we developed a simulator with real world traffic data. The simulation operates at the timescale of vehicle arrivals (in seconds) while capturing task arrivals at finer timescales (in milliseconds) by employing queuing model approximations for task processing delays. Running the simulation at the granularity of arriving vehicles allows us to investigate performance over extended time spans (hours or days) and examine the differences in performance between busy and off-peak hours.
\end{enumerate}

The proposed DHPG effectively addresses aforementioned key challenges in C-V2N applications. First, its concise design allows for processing speeds in the sub-millisecond range, meeting the stringent real-time requirements of these applications in terms of 99-percentile delay guarantees. Second, by formulating and solving the problem as an MDP, DHPG inherently adapts to the dynamic nature of application traffic, allowing the system to respond to fluctuations over time. 
%At last, the use of a joint latent space and hybrid action space alleviates scalability issues, facilitating time efficient decision-making on a large scale.
 At last, the use of a joint latent space, which captures a compact representation of the global system state, along with the use of a hybrid action space, enables joint decision-making with reduced input dimensionality. The use of a single critic for different types of actions further simplifies training by avoiding the need for multiple critics, leading to more efficient learning and better scalability on a large scale.
To the best of our knowledge, the proposed DHPG is the first end-to-end (E2E) DRL framework that directly considers the full joint task placement and scaling action space in a single step without relying on heuristics or approximations that may reduce the solution space, and is specifically designed to accommodate dynamic vehicle traffic changes in real-time. These features make DHPG a robust and scalable solution for the complexities of C-V2N environments.

The remainder of this paper is organized as follows. 
In Section~\ref{sec:related} we provide a brief review of the related problems and the corresponding solutions.
In Section~\ref{sec:problem} we describe the problem in mathematical terms and provide its formulations. Section~\ref{sec:ddpg} details our proposed approach. Section~\ref{sec:simulation} describes the simulation environment and the supported C-V2N application. 
We evaluate the proposed approaches in Section~\ref{sec:evaluation}. Finally, Section~\ref{sec:conclusion} further discusses the main findings of this paper and points out future research directions.

\section{Related work} 
\label{sec:related}

% In this section, we review the existing literature on resource scaling, task offloading, and the joint optimization with a focus on the context of \gls{v2x} communication.
In this section, we review the literature on resource scaling, task offloading, and the joint optimization problem.
% The review serves as a foundation to understand the state-of-the-art approaches and identify gaps that our work addresses.

%scaling of computing resources
\noindent \textbf{Resource Scaling.}
Resource scaling is crucial in edge computing environments, particularly for services with highly variable traffic demands and sensitivity to latency. The challenge lies in dynamically adjusting the allocated computational resources to meet service demands while optimizing factors such as cost, energy efficiency, and service quality.
% Authors in~\cite{10.1007/s10586-021-03265-9} classify the scaling techniques in the context of IoT-based cloud applications, as either schedule-based or rule-based solutions.
% Specifically, they show that threshold-based policies, queuing theory, control theory, time-series analysis, and \gls{rl} are candidates for rule-based auto scaling.
Authors in~\cite{10.1007/s10586-021-03265-9} classify auto-scaling techniques for IoT-based cloud applications as either schedule-based or rule-based solutions. Specifically, they identify threshold-based policies, queuing theory, control theory, time-series analysis, and \gls{rl} as examples of rule-based approaches. Among these, threshold-based policies define specific upper or lower bounds for performance metrics, triggering scaling actions when these thresholds are crossed.
% However, for rule-based auto scaling solutions, \gls{rl} is not the only AI/ML approach employed -- as pointed out by \cite{8540003}. Specifically, {\color{red}[what else AI/ML methods they are using]}.
In~\cite{8422788}, authors use classification-based \gls{ml} solutions to predict if scaling actions are necessary based on current network conditions.
Moreover, authors in~\cite{8806631} propose an ML-based method that can
assist in proactive auto-scaling by forecasting the number of VNF
instances.
Authors in \cite{8540003} investigate the application of DRL in solving resource management for network slicing scenarios.
% Authors in~\cite{8422788} propose to forecast the future network demand to trigger an~\gls{ilp} solver that takes provisioning and scaling decisions.
Additionally, authors in~\cite{noms} propose to vertically scale V2N services using a variant of Deep Deterministic Policy Gradient (DDPG) that captures the structure of the discrete action space to avoid the curse of dimensionality.

Forecasting traffic demands is also crucial, as it provides insights on the scaling operation.
Related studies use classic
time-series techniques such as \gls{arima} or 
 gls{tes}~\cite{Winters1960TES, LEE201166},
\gls{nn} solutions based on \gls{lstm} cells
\cite{lossleap,https://doi.org/10.1049/iet-its.2016.0208}, convolutional neural networks~\cite{aztec}, or even \gls{gnn}s~\cite{sdgnet}.
% Despite the context where forecasting is used, it is important to keep in mind that AI/ML approaches require large amounts of data and computing resources, whilst traditional time-series approaches do not. Thus, works as~\cite{v2n-access,DeVleeschauwer2021} compare the trade-offs between traditional and AI/ML solutions for forecast-assisted scaling.
While AI/ML approaches offer powerful capabilities, they typically require substantial amounts of data and computational resources, which may be a limiting factor in certain contexts. In contrast, traditional time-series forecasting techniques often demand fewer computational resources and lower energy consumption due to their reliance on simple analytical formulas.
Studies such as~\cite{v2n-access,DeVleeschauwer2021} have compared the trade-offs between traditional and AI/ML solutions for forecast-assisted scaling.

%offloading of computing resources
\noindent \textbf{Task Offloading/Placement.}
Task offloading is a crucial aspect of edge computing, enabling resource-constrained devices to delegate computationally demanding tasks to more powerful edge nodes or cloud servers. The goal is to enhance the overall performance, reduce latency, and improve energy efficiency. The decision on whether or not to offload a task depends on various factors, including the computational requirements of the task, the available resources of the device and the edge nodes, and the latency of the network.
The early approaches to task offloading~\cite{8542668, 8240666, 7553459}, were primarily based on low complexity optimization algorithms, which can be simple and computationally efficient, but often led to suboptimal performance. To address this limitation, researchers have explored dynamic task offloading strategies that employ machine learning techniques~\cite{9046820, ZHAO2019346}.
% Deep reinforcement learning (DRL) has gained significant attention in recent years due to its ability to learn complex decision-making policies from high-dimensional data. DRL algorithms can dynamically evaluate the trade-offs between offloading decisions and system performance metrics, such as latency, throughput, and energy consumption. 
Several DRL-based task offloading frameworks have been proposed in the recent literature. The authors in~\cite{8647611} present the double deep Q-network (DDQN) approach for optimizing resource allocation in heterogeneous networks. Computation offloading in Mobile Edge Computing (MEC) and Internet of Things (IoT) devices with DRL has been explored in several studies, including~\cite{9363256, 8690980, 8761385, 8771176, DDQNEC, 10036357}. These frameworks typically employ deep NNs to represent the offloading policy, which learns to map task characteristics and network conditions to optimal offloading decisions. The frameworks differ in their choice of neural network architecture, reward function design, and training methodology.
%The joint optimization of computation offloading and resource allocation with DDPG-based DRL is studied in~\cite{9435782, chen2020decentralized}.
% ~\cite{8647611, 9363256, 8690980, 8761385, 8771176}. These frameworks typically employ deep neural networks to represent the offloading policy, which learns to map task characteristics and network conditions to optimal offloading decisions. The frameworks differ in their choice of neural network architecture, reward function design, and training methodology.
% % Forecasting solutions
% Forecasting future traffic demand is actually quite effective, since an accurate estimation of the incoming demand provides insights on the scaling operation.
% Works in the literature resort to classic
% time-series techniques such as \gls{arima} or 
%  gls{tes}~\cite{Winters1960TES, LEE201166},
% \gls{nn} solutions based on \gls{lstm} cells
% \cite{lossleap,https://doi.org/10.1049/iet-its.2016.0208}, convolutional neural networks~\cite{aztec}, or even \gls{gnn}s~\cite{sdgnet}.
% Such approaches have been successfully used in the context of effective network
% slicing~\cite{deepcog,pi-road,microscope}
% and cloud dimensioning~\cite{xiao2019NFVdeep}.
% But despite the context where forecasting is used, it is important to keep in mind that AI/ML approaches require large amounts
% of data and computing resources, whilst traditional time-series approaches do not. Thus, works as~\cite{v2n-access,DeVleeschauwer2021} compare the trade-offs between traditional
% and AI/ML solutions for forecast-assisted scaling.

% joint
\noindent \textbf{Joint Task Offloading/Placement and Resource Scaling.}
%The joint optimization of resource scaling and task offloading aims to address the limitations of treating these problems separately. By considering the interdependencies between the two problems, it is expected to enhance the overall system performance.
The joint optimization of resource scaling and task offloading aims to address the limitations of treating these problems separately by considering their interdependencies, thereby enhancing overall system performance. The joint problem of task offloading and resource allocation at the edge, which involves a wider range of resources and resource management strategies, has been widely investigated in literature. 

%%%%% problems offloading and resource allocation, energy, task delay deconoupling

In~\cite{10.1109/TMC.2022.3150432}, authors proposed an online joint offloading and resource allocation framework in an energy-constrained MEC network, explicitly considering the task delay, energy consumption and MEC long-term energy, using Lyapunov optimization. Authors in~\cite{LIU20231399} proposed a computation offloading and resource allocation scheme. The problem is decoupled into two sub-problems for the offloading mode selection and the resource allocation, and solved by a distributed iterative algorithm.
Problem decoupling is also used in~\cite{li2022joint}, where the total energy consumption subject to the service latency requirement is minimized by jointly optimizing the task offloading ratio and resource allocation.
In~\cite{LAI2024110692}, authors addressed the placement and allocation problem in heterogeneous multi-server systems using an efficient approximation algorithm. Their approach focused on maximizing overall utility in static scenarios based on both synthetic and real-world utility functions.

%%%%% problems of offloadimg and CPU frequency ALLOCATION 
Authors in~\cite{7914660}, minimize the total execution latency and energy consumption, by jointly optimizing the task allocation decision and the CPU frequency of mobile devices, using a semi-definite relaxation-based approach. Authors in~\cite{MA} employ Markov approximation to jointly optimize task assignment and CPU frequency scaling of mobile devices.
Similarly,  Authors in~\cite{chen2020decentralized} proposed a DDPG-based framework to adjust the levels of both transmission power and local execution (CPU-cycle frequency) power. In~\cite{9435782}, authors address the joint optimization of computation offloading and resource allocation in a dynamic MEC system, considering constraints like task completion time and energy consumption of mobile devices. Specifically, a DDPG-based method was proposed to optimize the offloading ratio, the computing power on mobile devices, and the uplink transmission power for offloading.

%%%%% problems of offloadimg allocation for V2X
Recent works have started to explore this joint optimization in the context of V2X. Authors in~\cite{peng2020deep} investigated the joint allocation of spectrum, computing, and storage resources in a MEC-based vehicle network using DDPG.
The problem of joint offloading and resource allocation for vehicular edge computing with result feedback delay was studied in~\cite{10048752}, where the task offloading decisions, the uplink bandwidth allocation and the computation resources allocation on the roadside unit are jointly optimized using an approximate algorithm.
In~\cite{WANG2023}, the authors presented a two-stage algorithm. The initial stage employs an optimization-based method to determine offloading decisions, followed by a multi-agent DDPG approach to adjust transmission power for efficient resource scaling. 
A Twin Delayed DDPG (TD3)-based DRL approach was introduced in~\cite{huang2023joint} to derive the optimal decision for task offloading, computation load ratio, and bandwidth allocation ratio.
In \cite{10024868} a game-theoretical approach is introduced for resource allocation and task offloading for vehicular edge networks, where the intra-server resource allocation and inter-server load-balanced task offloading are jointly optimized.

In this study, we assume that tasks are fully offloaded and we further expand the decision space by considering all available edge servers in the area as potential candidates for task placement, provided latency constraints are met. While this larger search space increases the complexity of the problem, it also creates opportunities to improve overall system performance. By dynamically distributing computational loads across multiple servers, we can reduce bottlenecks, optimize resource utilization, and minimize costs by avoiding unnecessary scaling actions and overprovisioning of resources.
%The proposed approach in this paper distinguishes itself from previous works by introducing DHPG method that jointly optimizes both task placement (discrete decisions) and resource scaling (continuous decisions) in vehicular edge computing systems.

% Unlike many existing solutions that address these problems separately or decouple them into sub-problems, DHPG handles the inter-dependencies between task offloading and resource scaling in a single framework.

%To the best of our knowledge, the proposed DHPG is the first E2E DRL framework that explores the full joint action space to simultaneously decide the placement and scaling in a single step, without relying on heuristics or approximations that may reduce the solution space, while considering the dynamic changes of the vehicle traffic in a real time (online) manner.
To the best of our knowledge, the proposed DHPG is the first E2E DRL framework that has direct access to the full joint task placement and scaling action space in a single step without resorting to heuristics or approximations, while designed to handle dynamic vehicle traffic variations in real-time and in a large scale. Unlike existing approaches that often address task offloading/placement and resource scaling in a decoupled or sequential manner (e.g.,~\cite{10.1109/TMC.2022.3150432,LIU20231399,li2022joint, WANG2023, 10024868}), rely on approximations (e.g.,~\cite{LAI2024110692, 7914660, MA, 10048752}), and/or focus on the system's short-term performance or static scenarios (e.g.,~\cite{LAI2024110692,7914660, MA, 10048752, 10024868}), DHPG captures the interdependencies between these tasks within a unified framework, and optimizes the long-term performance of dynamic systems.
While step-wise/heuristic approaches may offer efficient solutions, they often implicitly reduce the search space by solving sub-problems in a pre-defined order (e.g., optimizing scaling before task placement) or introduce additional dependencies between decision variables through heuristic steps (e.g., using a greedy approach to find the placement strategy that minimizes latency based on the current scaling decision). In contrast, DHPG avoids these limitations by jointly optimizing both tasks. %, achieving a better outcome by considering their interdependency.
Furthermore, by using real-world traffic data and addressing stringent latency constraints, such as maintaining the $99^{th}$ percentile of service delay within acceptable limits, the study goes beyond simply improving cost-efficiency to ensure reliable and high-quality service performance.

\section{System model and Problem formulation}
\label{sec:problem}

%\todo[ inline]{*by Luca, How does our work compare with previous works ?: Danny: see section related work}

\begin{table}[b]
 \centering
  \caption{Notation Table}
  \label{tab:symbols}
  \begin{tabular}{c p{0.8\columnwidth}}
   \toprule
   \textbf{Symbol} & \textbf{Definition}\\
   \midrule
   $P$ & number of \glspl{pop} \\
   $V$ & number of vehicles \\
   $p$ & \gls{pop} number\\
   $v$ & vehicle number\\
   $t_v$ & arrival instant of vehicle $v$\\
   $T_v$ & departure instant of vehicle $v$\\
   $p_v$ & \gls{pop} at which vehicle $v$ arrives ($\in \{1, 2, \ldots, P \}$)\\
   $p'_v$ & \gls{pop} processing vehicle $v$ tasks\\
   $N_p(t)$ & number of vehicles processed at \gls{pop} $p$ at time $t$\\
   %$N_{max}$ & maximum number of vehicles that a \gls{pop} can process\\
   $\lambda$ & avg. task arrival rate generated by the \gls{cv2n} application\\
   $C_{p,v}$ & CPUs used by the \gls{cv2n} application at \gls{pop} $p$ right after $t_v$\\
   $C_{p,v}^+$ & \acrshort{cpu} increase/decrease at \gls{pop} $p$ right after $t_v$\\
   $C_p^{\max}$ & maximum processing capacity at \gls{pop} $p$\\
   $\mu(C)$ & service rate at \gls{pop} employing $C$ CPUs\\
   $\rho_{p,v}$ & load in \gls{pop} $p$ at time $t_v$\\
   $\mathbb{E}[d_{p,v}]$ & (average) processing delay in \gls{pop} $p$ at time $t_v$\\
   %$\mathbb{E}[l_{p,p'}]$ & (average) transmission latency between \gls{pop} $p$ and $p'$\\
   %$l_{p,p'}$ & maximum transmission latency between \gls{pop} $p$ and $p'$\\
   $l_{p,p'}$ & transmission delay between \gls{pop} $p$ and $p'$\\
   $d_v$ & total delay experienced by vehicle $v$\\
   %$d_v^\kappa$ & $\kappa$-percentile delay experienced by vehicle $v$\\
   $d_{tgt}$ & target delay\\
   $R(\cdot)$ & reward function to be optimized\\
   %   $C_{p,v}$ & number of \gls{cpus} at \gls{pop} $p$ in interval $[t_v, t_{v+1})$ \\
   \bottomrule
   \end{tabular}
\end{table}
%%%%%%%%%%%%%%%%%%%%%%%%%%%%%%%%%%%%%%%%%%%%%%%%%%%%%%%%%%%%%%%%
In this section we provide the system model and the associated problem formulation ($i$) as a combinatorial optimization problem, and ($ii$) as an MDP.
%In this section we provide the system model and provide the associated problem formulation as. Specifically, in Section~\ref{subsec:system-model} we model the multi-PoP environment, the \gls{cv2n} application, task placement and scaling actions, as well as average task processing delay at each PoP in the context of the \gls{cv2n} application. In Section~\ref{sec:optimization} we formulate the corresponding joint optimization problem for efficient \gls{cv2n} service provisioning under the (unrealistic) assumption that all future vehicle arrivals are known. 
%Moreover, we prove that the problem is~$\mathcal{NP}$-hard. Finally, in Section~\ref{subsec:mdp} we formulate the problem as an ~\gls{mdp}, which does not use future information on vehicle arrivals. %This \gls{mdp} is then tackled in Section~\ref{sec:ddpg}.

%of placing \gls{cv2n} tasks to \glspl{pop}, the scaling of the \glspl{cpu} in the \glspl{pop}, and the delay resulting from the traffic steering and scaling decisions. Table~\ref{tab:symbols} summarizes the most relevant symbols that we use. Moreover, we prove that the problem to tackle is~NP-hard. Then, in Section~\ref{sec:optimization} we formulate an optimization problem to achieve efficient \gls{cv2n} service provisioning under the (unrealistic) assumption that all future vehicle arrivals are known. Finally, in Section~\ref{subsec:mdp} we formulate an ~\gls{mdp}, which does not use future information on vehicle arrivals. This \gls{mdp} is then tackled in Section~\ref{sec:ddpg} with an \gls{ddpg} approach.

\subsection{System Model}
\label{subsec:system-model}

% \begin{figure*}
%     \input{Figures/system-figure.tex}
%     \caption{Vehicles $v$ send C-V2N tasks to ICT infrastructure (PoP) where these tasks are processed, employing cellular  communications. During non-rush-hour (2 am) C-V2N tasks  are processed  locally at the PoP co-located with the base station. During rush-hour (8.30 am) selected C-V2N tasks may be processed by another edge computing node introducing additional network transmission latency $l_{1,2}$. To accommodate the peak demand for the particular C-V2N application in rush hour the \gls{pop} scales up the active CPUs (e.g., \gls{pop}~2 scales from $C_{2,\cdot}=1$ to $C_{2,\cdot}=2$ CPUs).}
%     \label{fig:system}
% \end{figure*}

\noindent
\textbf{Infrastructure.} 
We consider a set of $P$ \glspl{pop} distributed throughout a city.
Each \gls{pop} $p\in\{1,\ldots,P\}$ has a maximum processing capacity of $C_p^{max}$.
Vehicle $v\in\{1,\ldots,V\}$ arrives at time instance $t_v$ at the
vicinity of a \gls{pop} that we denote as $p_v$.
A \gls{bs} (i.e., gNodeB) provides network connectivity to
this vehicle $v$ in the vicinity of the \gls{pop} $p_v$.
The application tasks stemming from the
\gls{cv2n} application running on vehicle $v$ are either
processed locally at the \gls{pop} $p_v$,
or redirected to be processed
at another \gls{pop} $p\neq p_v$.
The vehicle $v$ leaves the vicinity of \gls{pop} $p_v$ at time $T_v > t_v$. 
At that time, vehicle $v$ either leaves the system altogether or enters the vicinity of another \gls{pop}.% (where it is identified with another sequence number).

\vspace{1mm}
\noindent \textbf{Application.} 
We consider a \gls{cv2n} application scenario, where each task is defined as a \textit{basic unit of work to be executed towards the accomplishment of an application service} \cite{islam2021survey}. For example, in this study we employ a \gls{cv2n} application where  each vehicle $v$ produces a video sequence to be processed at the edge; each frame is decoded and analysed at some PoP $p_v\in P$, constituting the application task, as described in detail in Section \ref{sec:simulation}.  We assume that every vehicle $v$ sends $\lambda$ tasks/s to its associated \gls{pop} during the interval $[t_v, T_v]$.  We denote as $d_{p,v}$ the processing delay experienced by tasks in \gls{pop} $p_v$ right after the arrival of vehicle $v$ at time $t_v$.
As we focus on the performance impact of task offloading from the vehicle to edge computing resources, we define the task execution delay, which includes transmission and processing delay. To experience a good application quality, this task execution delay should be bounded by $d_{tgt}$.
%As we focus on task offloading from the vehicle to edge computing resources, the task execution delay, including transmission and processing delay is bounded by $d_{tgt}$.

\begin{figure}
     \includegraphics[width=0.95\columnwidth]{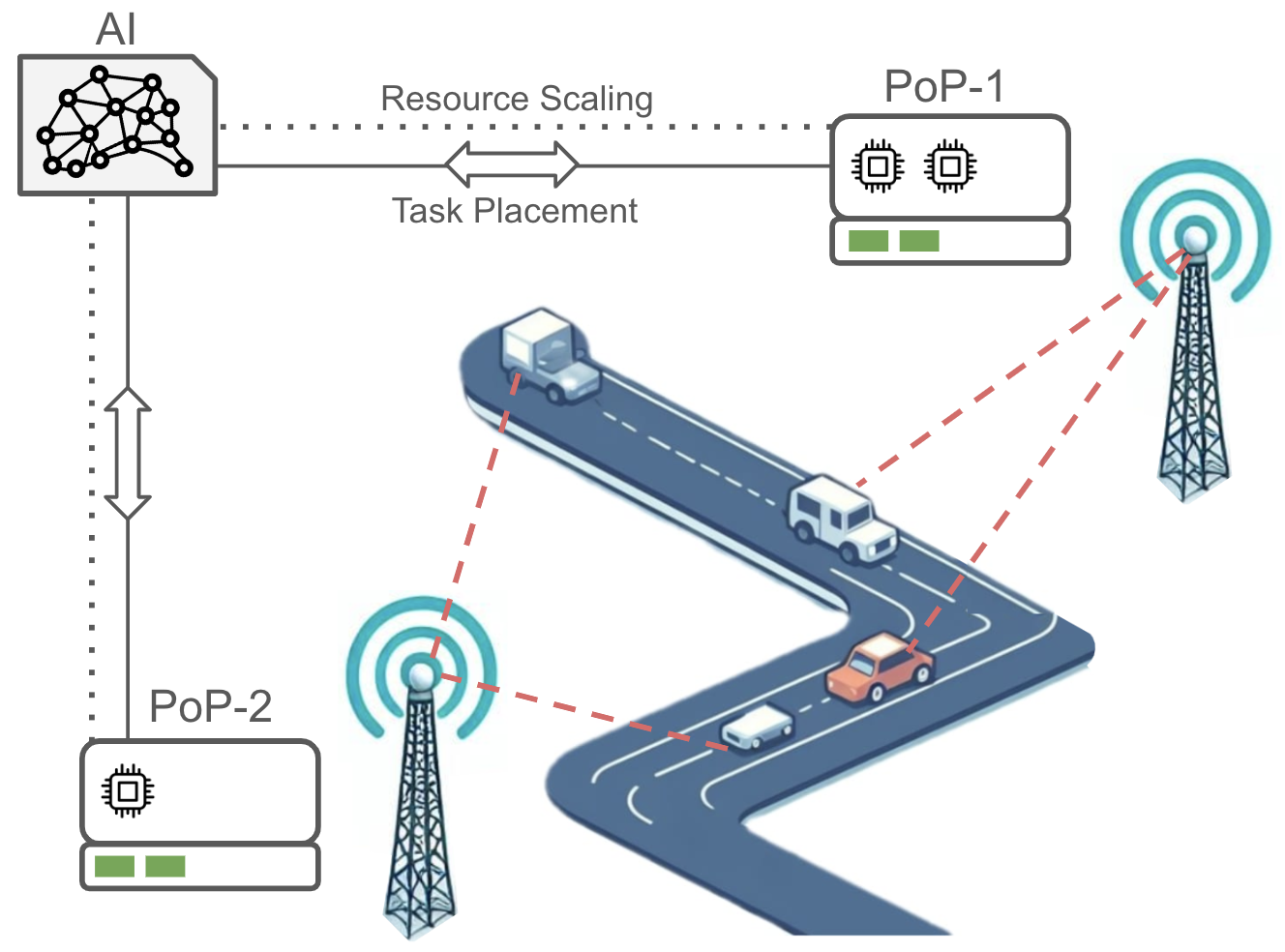}
    \caption{System Overview}
    %Illustration of the system where vehicles are communicating with Points of Presence (\gls{pop}s). These \gls{pop}s are connected to a centralized AI unit for management. Vehicles offload application tasks (green blocks) to the associated \gls{pop} for processing. at the Mobile Edge Computing (MEC) layer. The computational resources allocated to each \gls{pop} are visualized by CPUs. In this example, \gls{pop}-1 has three C-V2N tasks requiring processing supported by two CPUs, while \gls{pop}-2 has two tasks and only one CPU. The AI unit allows re-allocating a task from a \gls{pop} encountering high latency to another \gls{pop} with available resources.}
    \label{fig:system}
    \vspace{-3mm}
\end{figure}
\vspace{1mm}
\noindent
\textbf{Task Placement.} 
When vehicle $v$ enters the vicinity of \gls{pop} $p_v$ at time $t_v$, a placement decision needs to be made  for offloading the task(s) of the particular vehicle. The decision is captured in the decision variable $p'_v$; the \gls{cv2n} tasks of that vehicle $v$ will be processed at the \gls{pop} near the vehicle if $p'_v=p_v$, or at another \gls{pop} when $p'_v\neq p_v$. Note that the placement decision $p_v'$ is made at time $t_v$ and it remains valid until the vehicle $v$ leaves the vicinity of \gls{pop} $p_v$ at time $T_v$.
Redirecting the tasks of vehicle $v$ from \gls{pop} $p_v$ to $p'_v$, %where $p'_v\neq p_v$, 
introduces
an additional transmission latency $l_{p_v,p'_v}$,
%{\color{red}a maximum transmission latency $l_{p_v,p'_v}$},
but allows to tap into resources that could be potentially underutilised at \gls{pop} $p'_v$.
%Note that $p'_v$ is a decision variable that is only defined in the interval $[t_v, T_v]$. 
The decision relies on information that \gls{pop} $p_v$ exchanges with all other \glspl{pop}. In particular, it relies on the task processing delay that other \gls{pop}s experience when the vehicle $v$ arrived, which we denoted as $d_{p,v},\ p \in P$.
%Since it takes some time for that information to propagate between \glspl{pop}, this processing delay value may be slightly outdated. 

%To make that local decision, the  in \gls{pop} $p_v$ relies on information that it exchanges with all other \glspl{pop}. In particular, it relies on the task processing delay that other \gls{pop}s experience when the vehicle $v$ arrived, which we denote as $d_{p,v},\ p \in P$. Since it takes some time for that information to propagate between \glspl{pop}, this processing delay value may be slightly outdated. 
\vspace{1mm}
\noindent
\textbf{Resource Scaling.} 
Scaling decisions determine the computing resources (number of CPUs) $C_p(t)$ required in each \gls{pop} $p \in P$ at regular instants $t$, with $C_{p}(t) \leq C_p^{max},\ \forall t$.
These instants $t$ could be different from the vehicle arrival instants, e.g., they could occur periodically (e.g., every 5 minutes), but in this study we assume that they are aligned with vehicle arrival instants $t_v$, following the placement decisions.
As a result, between consecutive \emph{scaling} instants the decision variables $C_p(t)$ remain constant, i.e.:
\begin{equation}
    C_p(t) = C_p(t'),\quad \forall p,\ \forall t, t' \in [t_v,t_{v+1})
\end{equation}
%{\color{red} Therefore, the decision variable $C_{p,v}$ at \gls{pop} $p$, upon arrival of vehicle $v$ (at time $t_v$) is fixed until the arrival of the next vehicle $v+1$ at time $t_{v+1}$.}
Consequently, from now on we will use $C_{p,v}$ to
denote $C_p(t_v)$.
Remark that in the set of all possible \emph{scaling} decisions, the ``no \emph{scaling}'' option is also included. %, i.e. require no scaling action.  
%% \begin{figure}[htp]
%%     \includegraphics[width=\columnwidth]{Figures/systemSetUp.png}
%%     \caption{{\color{red} Placeholder to be updated:} System set up, vehicle arrivals, \emph{placement} and \emph{scaling} decisions. @Jorge: to update or to replace}
%%     \label{fig:SystemSetUp}
%% \end{figure}

\vspace{1mm}
\noindent
\textbf{Task Processing Delay.}
The average task processing delay $\mathbb{E}[d_{p,v}]$ in \gls{pop} $p$ at time $t_v$ can be estimated using a queuing model, given the task arrival and service processes. 
As each vehicle $v$ sends $\lambda$ tasks/sec to its associated \gls{pop} during the interval $[t_v, T_v]$, the task arrival process has an arrival rate $\lambda N_{p,v}$, where $N_{p,v}$ denotes the number of vehicles assigned to $p$ at $t_v$ as follows:
%when vehicle $v$ arrived at $t_v$:
\begin{equation}
    N_{p,v} = \sum_{v'\in V:\ p'_{v'}=p} \mathds{1}_{[t_{v'},T_{v'}]}(t_v)
    \label{eq:number-vehicles}
\end{equation}
 $\mathds{1}_A(x)$ is an indicator function, which is 1 when \mbox{$x\in A$}, and 0 otherwise. In line with \cite{vnf-sharing, near-optimal-placement, joint-placement-infocom}, we model the task service process at each \gls{pop} with an \mbox{M/G/1-PS} queue, where PS stands for Processor Sharing, with a serving rate $\mu(C_{p,v})$. By modeling a~\gls{pop} with an \mbox{M/G/1-PS} queue we assume that the arrival rate of vehicles' tasks follow a Poisson distribution, and such assumption holds for sufficiently large task arrival rates according to the Palm-Khintchine theorem~\cite{heyman1982stochastic}. Regarding the service rate $\mu(C_{p,v})$, we do not make any assumption about its distribution. Similar to \cite{vnf-sharing, near-optimal-placement, joint-placement-infocom}, we only consider that $\mu(C_{p,v})$ increases as the number of CPUs $C_{p,v}$ increases. %We will further discuss and justify its precise form in Section~\ref{sec:evaluation}.
Under the assumptions above, the average processing delay $\mathbb{E}[d_{p,v}]$ at time $t_v$ in \gls{pop} $p$ is:
\begin{equation}
    \label{eq:avg-delay}
    \mathbb{E}[d_{p,v}] =
    \begin{cases}
        \frac{1}{\mu(C_{p,v}) - \lambda N_{p,v}} & \mbox{if $\mu(C_{p,v}) >\lambda N_{p,v}$} \\
    \infty & \mbox{otherwise} 
    \end{cases}
\end{equation}

\vspace{1mm}
\noindent\textbf{Task Total Delay}.
In the average delay experienced by a task we consider both the
transmission latency $l_{p_v,p_v'}$ and processing
delay $d_{p',v}$. We define the experienced delay $d_v$ as:
\begin{equation}
    d_v = 
    l_{p_v,p'_v} + \mathbb{E}[d_{p_v',v}]
    \label{eq:total-delay}
\end{equation}
i.e. $d_v$ is the sum of the worst transmission latency
and average processing delay, which we compute
using~\eqref{eq:avg-delay}.

However, ensuring $d_v \leq d_{tgt}$ does not guarantee
that C-V2N tasks meet the delay requirement the 99\% of the times.
To that end, we define the $\kappa$-percentile of the
experienced delay as:
\begin{equation}
    d_v^\kappa = 
    l_{p_v,p'_v}^\kappa+ d_{p_v',v}^\kappa
    \label{eq:kappa-delay}
\end{equation}
with $l_{p_v,p'_v}^\kappa,d_{p_v',v}^\kappa$
the $\kappa$-percentile of the transmission and processing
latency, respectively.
The expression for $d_{p_v',v}^\kappa$ depends on how
processing times are distributed, i.e., it will vary
depending on the C-V2N tasks. To remain as generic as
possible, we resort to the well-known
M/G/1-PS average delay~\eqref{eq:avg-delay} expression
and bound the $\kappa$-percentile of the processing
delay as $d_{p,v}^\kappa < K(\kappa) \cdot \mathbb{E}[d_{p,v}]$.
Here, $K(\kappa) \geq 1$ is a bound that depends on the
% {\color{red} reliability
% level $\kappa$.}
percentile $\kappa$.
%For instance, numerical results in Appendix~\ref{app:latency}
%show that $K(\kappa)=2$ with $\kappa=99$ and deterministic
%processing times of video decoding/recognition tasks
%-- i.e. M/D/1-PS queues. 
With the aforementioned, we provide an upper bound denoted as $\overline{d_v^\kappa}$, i.e.,
an upper bound of the $\kappa$-percentile
of the experienced task delay $d_v^\kappa$:
\begin{equation}
    \overline{d_v^\kappa} = l_{p_v,p_v'} +
    K(\kappa) \cdot \mathbb{E}[d_{p_v',v}]
    \label{eq:bound}
\end{equation}

%That is, we resort to the average delay closed-form expression in~\eqref{eq:avg-delay} to bound the $\kappa$ delay percentile of the C-V2N task.

%\bigskip

Fig.~\ref{fig:system} illustrates the placement of vehicles' tasks to \gls{pop}s, and number of CPUs allocated to support their respective requirements at each \gls{pop}. Based on placement and scaling decisions made by the agent, vehicles offload application tasks (green blocks) to the associated PoP for processing, while the computational resources allocated per PoP are visualized by CPUs. In
this example, PoP-1 is processing two tasks
supported by two CPUs, while PoP-2 has two tasks supported by 
one CPU. %The AI unit allows re-allocating a task from a PoP encountering high latency to another PoP with available resources

% In Fig.~\ref{fig:system} we illustrate (i) the \emph{placement} of vehicles' \gls{cv2n} tasks to \gls{pop}s $p_v'$ and (ii) \emph{scaling} decisions in each \gls{pop} during non-rush- and rush-hours.
% Note that we assume that a base station provides cellular connectivity to vehicles using the \gls{cv2n} service. 

 %, and we also consider Layer~3 connectivity in between the~\gls{ru} and the~\gls{pop} that processes the \gls{cv2n} tasks.
%In the next sections we formulate an optimization and~\gls{mdp} to decide the best vehicle-to-\gls{pop} steering (to determine decision variable $p'_v$) and \gls{cpu} scaling (to determine decision variable $C_{p,v}$). The goal is to minimize resource consumption, yet satisfying delay requirements of \gls{cv2n}~services.

%%%%%%%%%%%%%%%%%%%%%%%%%%%%%%%%%%%%%%%%%%%%%%%%%%%%%%%%%%%%%%%%%%%%

\subsection{Optimization Problem}
\label{sec:optimization}

%{\color{red}@Xi + @Chrysa: to review}

In this section we assume that all vehicle arrival instants $t_v$ and the \glspl{pop} $p_v$ the vehicle arrive at are known. We formulate the task placement and scaling actions as an optimisation problem aiming to maximize an objective function $R(\cdot)$ while taking the appropriate \emph{task placement} and \emph{scaling} decisions, i.e.: ($i$) which \gls{pop} $p'_v$ has to process the \gls{cv2n} tasks produced by vehicle $v$; and ($ii$) how many \glspl{cpu} $C_{p,v}$ are required by \gls{pop} $p$ to process the vehicles' \gls{cv2n} tasks (just after the placement decision for vehicle $v$ has been made until the arrival of the next vehicle).

%We resort to an optimization formulation to describe the problem we want to solve. Namely, our optimization problem will aim to maximize an objective function $R(\cdot)$ with the adequate \emph{steering} and \emph{scaling} decisions, i.e.: (i) which \gls{pop} $p'_v$ has to process the \gls{cv2n} tasks produced by vehicle $v$; and (ii) how many \glspl{cpu} $C_{p,v}$ are required by \gls{pop} $p$ to process the vehicles' \gls{cv2n} tasks (just after the steering decision for vehicle $v$ has been made until the arrival of the next vehicle).
%\todo[inline]{*by Luca, Maybe I already asked this. Which is the set of vehicles for which we solve the optimization problem ? That is, which is the set of vehicles $v$. Are the vehicles arriving within a certain interval of the specific vehicle we are considering ? Danny: in principle for all of them; in practice for all the ones that are active at the moment vehicle $v$ arrives.}

\begin{problem}[\emph{Task placement} and \emph{scaling}]
\label{problem}
\begin{align}
    \max_{C_{p,v},\ p_v'} & \ \sum_{v}\sum_p R(C_{p,v}, p'_v)\\
    \text{s.t.:} 
    %& \quad N_{p,v} \leq N_p^{\max},\quad \forall p, v \label{eq:admission-control}\\
    & \quad 
    C_{p,v} \leq C_p^{\max},\quad \forall p, v \label{eq:limit-cpus}\\
    %\sum_p p_v'=1,\qquad \forall v\label{eq:to-one-pop}\\
    & \quad p'_v \in P \label{eq:to-one-pop},\quad \forall v \\
    %&\qquad p_v'\in\{0,1\},\quad \forall p,v\\
    &\quad C_{p,v} \in\mathbb{N},\quad \forall p, v
\end{align}
\end{problem}
\noindent with $R(C_{p,v}, p_v')$ %\todo[inline]{*by Luca, slight different formula here.; Danny: corrected}
being the reward function that we obtain based on the \emph{scaling} and \emph{placement} decisions taken for vehicle $v$ at the time $t_v$ it arrives.
%Constraints~\eqref{eq:admission-control} specify an admission control $N_p^{\max}$ on every PoP $p$.
Constraint \eqref{eq:to-one-pop} imposes that the \gls{cv2n} tasks of each vehicle are placed on a single \gls{pop} while  \eqref{eq:limit-cpus} describes the processing capacity constraints (number of CPUs) at a \gls{pop}.
%The admission control parameter is set such that the total load  \mbox{$\rho_{p,v}=\frac{\lambda N_{p,v}}{\mu(C_{p,v})} $} on \gls{pop} $p$ is smaller than 1, so that the processing delay does not grow to infinity.

%and \eqref{eq:limit-cpus} impose that the tasks of each vehicle are steered to a single \gls{pop} and that the number of \glspl{cpu} at a \gls{pop} is bounded; respectively. The admission control parameter is set such that the total load  $\rho_{p,v}=\frac{\lambda N_{p,v}}{\mu(C_{p,v})} $ on \gls{pop} $p$ is smaller than 1, so that the processing delay does not grow to infinity.
%\todo[inline]{*by Luca, which is the relationship between $N_{\max}$ and $\rho_v(t)$ ? Danny: see added sentence.}.

Since we aim to meet the delay constraint $d_{tgt}$ of the \gls{cv2n} service, and minimize the number of \glspl{cpu} that we employ at the \glspl{pop}; we resort to the following reward function:
\begin{equation}
 R(C_{p,v}, p'_v)= 
 \frac{d_v}{d_{tgt}}
 \cdot\ \exp\left(- \frac{1}{2}\left[\left(\frac{d_v}{d_{tgt}}\right)^2 - 1\right] \right)
 \label{eq:reward}
\end{equation}
with $d_v$ 
dependent on both the \emph{placement}~$p'_v$ and \emph{scaling} $C_{p, v}$ decisions, as they both impact transmission and processing delay based on 
formula \eqref{eq:total-delay}.

Fig.~\ref{fig:reward}~illustrates how the reward function~\eqref{eq:reward} aims at meeting the target delay $d_{tgt}$ without leading to over- or under-provisioning of computing resources (\glspl{cpu}). Namely, over-provisioning resources  $C_{p,v}$ at the PoP will result in a delay lower than the target one $d_{tgt}$ as we use more \glspl{cpu} than necessary. Conversely, under-provisioning computing resources  %the number of \glspl{cpu} $C_{p,v}$ 
%at the \gls{pop} where the vehicle is assigned 
results in exceeding the target delay, decreasing the reward. The goal of the reward function is to push $d_{t}$ toward $d_{tgt}$, while having its range bounded $R(\cdot) \in [0, 1]$ that provides effective signals for training stability. Without a bounded range, rewards could go to plus/minus infinity when the latency is too small/large, which can hinder the learning of DRL agents by incurring extreme gradients during backpropagation. Moreover, the placement decision $p'_v$ also impacts the reward, for processing the vehicle $v$ tasks at a \gls{pop} $p'_v \neq p_v$ results in a high transmission latency $l_{p_v,p'_v}$. As a result, the total delay may increase beyond the target $d_{tgt}$ thus reducing the reward. Ideally,  \emph{placement} and \emph{scaling} decisions should lead to a total delay as close as possible to the target $d_{tgt}$.

Notice that the reward function of~\eqref{eq:reward} is nonlinear in the decision variables $C_{p,v}$ and $p'_v$ (via equation~\eqref{eq:total-delay}). Taking logarithms on~\eqref{eq:reward} would not suffice, as it would result to a %for still we would have the 
processing delay term in the power of two. Moreover,  when we substitute the processing delay with its average according to formula~\eqref{eq:avg-delay}, we  have a non-linear dependency with the number of \glspl{cpu} $C_{p,v}$ and the number of vehicles that a~\gls{pop} processes.
%even if we unroll the processing delay with its average~\eqref{eq:avg-delay}, it appears a non-linear dependency with the number of \glspl{cpu} $C_{p,v}$ and the number of vehicles that a~\gls{pop} processes -- see~the denominator of \eqref{eq:avg-delay}. Hence, we cannot linearize the reward function~\eqref{eq:reward}. 
Additionally, the reward function~\eqref{eq:reward} is not convex/concave in the decision variables, for the second derivative with respect to the delay
with $d=d_v/d_{tgt}$, does not have the same sign for all $d$:
\begin{equation}
    \frac{\partial^2}{\partial d^2}R(\cdot) = d(d^2-3)\exp\left(-\frac{d^2-1}{2}\right)
     \label{eq:deriv}
\end{equation}
Therefore, we cannot drive the search of the optimal solution using the chain rule on the \emph{task placement} $p_v'$ and \emph{scaling}~$C_{p,v}$ variables -- note both variables determine the total delay $d_v$ -- as the search may be trapped in local optima. Achieving a global optimal solution for Problem~\ref{problem} requires exhaustive search. However, in the following lemma we show that the problem is $\mathcal{NP}$-hard thus computationally intractable.

\begin{figure}[t]
    \centering
    \begin{tikzpicture}[xscale=.03, yscale=2]
  \tikzmath{\T=100;};
  
  % Axis
  \draw[->] (-5, 0) -- (240, 0) node[right] {$d$};
  \draw[->] (-5, 0) -- (-5, 1.9) node[above] {reward};
  
  % Target delay
  \draw (\T, 0.05) -- (\T, -0.05);
  \node[below] at (\T, 0) {$d_{tgt}$};
  
  % Function
  \draw[domain=0:230, line width=3, smooth, variable=\x,  color=DodgerBlue4] plot ({\x}, {(\x/\T)*exp(-1/2*( (\x/\T)^2 - 1))});
  
  % Target delay limit
  \draw[dashed] (\T, 0) -- (\T, 1.5);
  
  % Overprovisioning / high propagations
  \draw[->] (\T+20, 1.3) -- (200, 1.3) node[align=center,above,midway] {\small Under-provisioning} node[below,midway] {$d>d_{tgt}$};
  
  % Underprovisioning / low propagation
  \draw[->] (\T-10, 1.3) -- (5, 1.3) node[above,midway,align=center] {\small Over-provisioning} node[below,midway]{$d<d_{tgt}$};
  
  % % Sweet-point
  % \node [align=center,anchor=south] (sweet) at (\T+40,.1) {Optimal scaling\\ \& placement};
  % \filldraw [black] (\T,1) ellipse (80pt and 2pt);
\end{tikzpicture}
    \caption{Reward as function of the service delay $d$.}
    \label{fig:reward}
    \vspace{-3mm}
\end{figure}

%Therefore, achieving an optimal solution of Problem~\ref{problem} requires an exhaustive search. However, in the following lemma we show that the problem is NP-hard and issuing an exhaustive search may lead to unfeasible computation times.

% After defining the reward function $R(\cdot)$, it is evident that the admission control constraint~\eqref{eq:admission-control} is not strictly necessary, for having $\lambda N_p(t)>\mu(C_p(t))$ results in infinite processing delays -- see~\eqref{eq:avg-delay} -- and rewards that tend to zero -- see~\eqref{eq:reward} and Figure~\ref{fig:reward}. Nevertheless, we keep constraint~\eqref{eq:admission-control} for it may be of relevant for the admission control procedure required by some infrastructure providers, and it also allows us to define the Problem~\ref{problem} complexity:
% {\color{red}@Jorge: above we should say that the
%admission control is necessary to avoid going to
%infty delay. So remove the sentence saying that it
%is not necessary.}

\begin{lemma}\label{lemma:complex}
    Problem~\ref{problem} is $\mathcal{NP}$-hard.
\end{lemma}

\noindent The proof is provided in Appendix A.
\vspace{1mm}
%\todo[inline]{*by Luca, if I am not wrong, to prove that the problem is NP-hard we need to find a linear transformation from our problem to a known NP-hard problem. Is the transformation we are making linear ?}

%\todo[inline]{Jorge: use Oracle rather than clairvoyant.% And say it is a global optimal solution.}

% Explain we implement a clairvoyant solution
Despite the $\mathcal{NP}$-hardness of Problem~\ref{problem} we use an ``oracle" to get the optimal solution. % as the oracle knows the arrival process of vehicles. 
The oracle knows in advance the arrival time $\{t_1,t_2,\ldots,t_V\}$
of every vehicle $\{1,2,\ldots,V\}$; and plugs such values in the optimization problem to obtain the
best \emph{placement}~$p'_v$ and \emph{scaling}~$C_{p,v}$ decisions. Due to the problem's complexity, computing the oracle solution is only feasible for small instances of the problem. % e.g., up to six vehicle arrivals. 
However we use it to derive optimality gaps to our proposed approach % (Section~\ref{sec:ddpg}) 
and other solutions we adapt from the state of the art.%(Section~\ref{ScalingApproaches}). %To get the oracle solution we model Problem~\ref{problem} using AMPL~\cite{ampl}.

Lastly, we replace the $d_{tgt}$ term in
the reward function by $\tfrac{d_{tgt}}{K(\kappa)}$ to
meet the $\kappa$-percentile of the delay.
%Lastly, we foster meeting
%the $\kappa$-percentile delay although
%the reward function~\eqref{eq:reward}
%is computed based on the average processing delay
%-- see~\eqref{eq:total-delay} expression for $d_v$.
%To do so, it is enough to replace the $d_{tgt}$ term in the reward function as stated in Lemma~\ref{lemma:kappa}.

\begin{lemma}[$\kappa$-percentile reward maximum]
    Replacing $d_{tgt}$ by $\tfrac{d_{tgt}}{K(\kappa)}$
    in~\eqref{eq:reward} ensures that the maximum reward
    is achieved before $d_v^\kappa$ reaches $d_{tgt}$.
    \label{lemma:kappa}
\end{lemma}
\begin{proof}
    Replacing the $d_{tgt}$ term in the
    $\tfrac{d_v}{d_{tgt}}$ ratio from~\eqref{eq:reward}
    results into the following inequality
    %% \begin{multline}
    %%     \frac{d_v}{\frac{d_{tgt}}{K(\kappa)}}=
    %%     \frac{K(\kappa)d_v}{d_{tgt}}>
    %%     \frac{l_{p_v,p_v'}+K(\kappa)\mathbb{E}[d_{p_v',v}]}{d_{tgt}}\\
    %%     >\frac{l_{p_v,p_v'}+d_{p_v',v}^{\kappa}}{d_{tgt}}
    %%     =\frac{\overline{d_v^{\kappa}}}{d_{tgt}}
    %%     >\frac{d_{v}^{\kappa}}{d_{tgt}}
    %% \end{multline}
    \begin{equation}
        \frac{d_v}{\frac{d_{tgt}}{K(\kappa)}}=
        \frac{K(\kappa)d_v}{d_{tgt}}>
        \frac{l_{p_v,p_v'}+K(\kappa)\mathbb{E}[d_{p_v',v}]}{d_{tgt}}
        =\frac{\overline{d_v^{\kappa}}}{d_{tgt}}
        >\frac{d_{v}^{\kappa}}{d_{tgt}}
    \end{equation}
    with the first inequality holding
    due to~\eqref{eq:total-delay}
    and $K(\kappa)\geq1$, and
    the latest equality given
    by the definition of $\overline{d_v^\kappa}$
    provided in~\eqref{eq:bound}.
    Consequently, if we replace the $d_{tgt}$
    term as specified in
    the Lemma statement, the maximum reward is achieved
    before the $\kappa$-percentile of the experienced delay
    reaches the target.
\end{proof}

Thanks to Lemma~\ref{lemma:kappa}, we can select
the $d_{tgt}$ term, use the average
processing delay from an M/G/1-PS queue~\eqref{eq:avg-delay}, solve the optimization Problem~\ref{problem}, and ensure that the solution will foster meeting the $\kappa$-percentile delay requirement.
Specifically, in Appendix~\ref{app:latency} we show that setting the target delay to $d_{tgt}/2$ and using the average delay formula of the M/G/1-PS, ensures meeting the $d_{tgt}$ delay requirement 99\% of the time.
%Moreover, Fig.16 shows setting the target delay to d_tgt/2 leads to an accurate approximation of the 99-percentile of the latency, thus resulting into no/negligible over-provisioning

%Appendix~\ref{app:latency} shows with simulations that setting $K(99)=2$ ensures the delay experienced by Advanced Driving tasks is under $100$\,\textrm{ms} the 99 percent of the time However, Lemma~\ref{lemma:kappa} ensures the vehicle target delay $d_v$ is bounded by $d_{tgt}/K(\kappa)$. That is, for a given placement and scaling decision $p_v',C_{p,v}$ the actual delay experienced by the vehicle may be smaller than $d_{tgt}$ and having $C_{p,v}-1$ CPUs may still meet the target delay. In brief, using Lemma~\ref{lemma:kappa} may lead to slight CPU over-provisioning. Nevertheless, the results in Appendix~\ref{app:latency} show that setting $K(99)=2$ leads to a tight approximation of $d_v$ that results in no/negligible over-provisioning.

% \begin{algorithm}
% \caption{Interaction with Environment}\label{alg:env}
% \begin{algorithmic}
% \State $env \gets Environment()$
% \State $arrivalTime, arrivalStation, State, Stop \gets env.step()$
% \While{$\neg Stop$}
%     \State $mappedStation \gets placementAlgorithm(arrivalStation, state)$
%     \State $env.placeCar(mappedStation)$
%     \State $nrOfCPUs \gets scalingAlgorithm(state)$
%     \State $env.setCPUs(nrOfCPUs)$
%     \State $arrivalTime, arrivalStation, State, Stop \gets env.step()$
% \EndWhile
% \end{algorithmic}
% \end{algorithm}

\subsection{\gls{mdp} Formulation}
\label{subsec:mdp}

%{\color{red}@Xi + @Chrysa: to review}

% Action space and transition probabilities
In reality, at the time placement and potentially scaling decisions are taken for vehicle $v$, the arrival times of future vehicles are not known. Therefore we reformulate Problem~\ref{problem} (which assumes this knowledge) as an ~\gls{mdp}.
%In particular, we define what the state and action spaces, as well as the transition 
In particular, we define the state and action spaces, as well as the transition probabilities and reward of the considered~\gls{mdp} $(\mathcal{S},\mathcal{A}, \mathbb{P}, R)$.% With such tuple, our~\gls{mdp} is completely defined.

The state at the arrival time $t_v$ of vehicle $v$ is represented by the number of vehicles and \glspl{cpu} at each~\gls{pop}:
\begin{equation}
    s_v=(p_v, N_{1,v}, C_{1,v}, \ldots, N_{P,v}, C_{P,v})
    \label{eq:state}
\end{equation}
where $p_v$ denotes the PoP at which vehicle $v$ arrives, and we define $s_{v,p}=(N_{p,v},C_{p,v})$
as the state of PoP $p$ upon the arrival of a
vehicle $v$.
Note that the state space is formally defined as $\mathcal{S}=\mathbb{N}^{2P+1}$. 
An action is then defined as:
\begin{equation}
    a_{v}=(p_v', C^+_{1,v},\ldots,C^+_{P,v})
    \label{eq:action}
\end{equation}
Therefore, the action space of our~\gls{mdp} is $\mathcal{A}=\mathbb{N}\times\mathbb{Z}^P$. The goal is to find a
policy \mbox{$\pi:\mathcal{S}\mapsto\mathcal{A}$}
that draws an action $a_{v} \in \mathcal{A}$ when the system is in state $s_{v}$.
This action for vehicle $v$ consists of deciding in which \gls{pop} its \gls{cv2n} tasks will be processed $p_v'$,
and determining the number of \glspl{cpu} to scale to at each~\gls{pop} $C_{1,v}^+,\ldots,C_{P,v}^+$. For example $C^+_{p,v}=1$ means that \gls{cpu} at \gls{pop}~$p$ is scaled up by one, while vehicle $v$ is placed to \gls{pop} $p'_v$.
% \textcolor{blue}{[actually they are decided at the same time but I think it doesn't matter here]}). Note that $C^+_{p,v}\in\mathbb{Z}$, thus, 
% It can take negative values relating to
% scaling down the number of \gls{cpu}s.
Negative values are associated with scaling down the number of \gls{cpu}s. %, and they may be
%large enough to turn off all \glspl{cpu}.
% \textcolor{blue}{[if it can take negative values then shouldn't it be $C^+_{p,v}\in\mathbb{Z}$ instead of $C^+_{p,v}\in\mathbb{N}$?]}.

%Namely, the action to take $a_{v} \in \mathcal{A}$ upon the arrival of vehicle $v$ at time $t_v$ consists of deciding in which \gls{pop} its \gls{cv2n} tasks will be processed $p_v'$,
%and of determining the number of \glspl{cpu} to scale to at each~\gls{pop} $C_{1,v}^+,\ldots,C_{P,v}^+$.

% Transition probabilities
The transition probabilities $\mathbb{P}(s_{v+1}|s_{v},a_{v})$ express how likely it is to end up in a new state $s_{v+1}$ based on the action $a_{v}$ taken in the prior state $s_{v}$.  In this context they depend upon the arrival process of the vehicles and their lingering time in the vicinity of each respective PoP. Note that the scaling decisions $C^+_{p,v}$ within the action described by formula~\eqref{eq:action} determine the number of \glspl{cpu} in the future state, namely
$C_{p,v+1}=\max\left\{0, C_{p,v}+C^+_{p,v}\right\}$. %{\color{red}  As a result, the only uncertainty in the future state $s_{v+1}$ will be the number of vehicles that each \gls{pop}~$p\in P$ will be processing, i.e., $N_{p,v+1}$, and the PoP $p_v$ where the vehicle is arriving.}
%{\color{red} The latter depends on ($i$) how many vehicles $v$ have their
%\gls{cv2n} tasks processed at \gls{pop}~$p$, i.e., $p'_v=p$; and
%($ii$) how many of such vehicles will leave in the interval $[t_v,t_{v+1})$.}
%For example, the longer it takes for vehicle $v+1$ to arrive, the more vehicles may leave, thus, $N_{p,v+1}$ will be smaller in the new state $s_{v+1}$.
%\textcolor{blue}{[is this just an example? as the N is not always smaller in the new state]}.
%Hence, the transition probabilities $\mathbb{P}(s_{v+1}|s_{v},a_{v})$ depend on the arrival process.%on the
%random variable that tells when vehicle $v+1$
%will arrive.

% Reward function
The last element to define in the \gls{mdp} tuple is the reward function $R(s_{v,p},a_{v})$.% That is, what will be the reward that \gls{pop} $p$ foresees if action $a_{v}$ is taken at its current state $s_{v,p}$. 
 As a reward function we employ the objective function used in the optimization Problem~\ref{problem}, i.e., $R(C_{p,v},p_v')$.

%In particular, the reward function matches the objective function~\eqref{eq:reward} that we used in the optimization Problem~\ref{problem}, i.e., $R(C_{p,v},p_v')$. Hence, the reward function of our~\gls{mdp} $R(s_{v,p},a_v)$ is precisely the objective function used in the optimization Problem~\ref{problem}. % -- with $s_{t,p}=s_{t,2p:2p+1}$ the state of \gls{pop}~$p$ at time $t$.

Given the definition of the state space, action space, transition probabilities, and the reward function; we formulate the \gls{mdp}.

\begin{problem}[Task placement and scaling]
\label{mdp}
    Given the 
    $(\mathcal{S},\mathcal{A}, \mathbb{P}, R)$
    tuple, find a policy $\pi$ that maximizes
    \begin{equation}
    \label{eq:mdp}
        \mathbb{E}_{a_v\sim\pi,\mathbb{P}(s_v,a_v)} \frac{1}{P}\sum_v\gamma^v \sum_p R(s_{v,p},a_v)
    \end{equation}
    with the discount factor $\gamma\in[0,1]$ .
\end{problem}

The goal of the \gls{mdp} is to maximize the expected discounted reward, averaging over the reward  obtained at each \gls{pop} over time. However, we do not make any assumption of how vehicles arrive to different roads where \glspl{pop} are located, nor the volume of road traffic over the day. Making such assumptions becomes hard when it comes to model unpredictable events as accidents, traffic jams, etc. Rather, we resort to a model-free approach to solve Problem~\ref{mdp}. %In the following Section~\ref{sec:ddpg} we propose to use a~\gls{rl} solution to draw an optimal \gls{cv2n} service provisioning policy $\pi$.

% \section{\gls{ddpg}-based \gls{cv2n}~scaling}
\section{Proposed method}
\label{sec:ddpg}

To solve Problem~\ref{mdp} we need to jointly optimize placement and scaling decision variables, that are both discrete in nature. 
In our previous work \cite{noms}, the resource scaling problem is tackled in a continuous fashion using the DDPG variant, to address  the scalability problem inherent to the high-dimensional discrete action space.
Specifically, we have deliberately employed a continuous action space for scaling decisions, to enhance robustness against increasing dimensionality.
%The approach  demonstrates robustness against increasing dimensionality, through the use of a continuous action space followed by a non-
%parametric discretization method t 
On the other hand, handling the placement problem is trivial using discrete DRL algorithms like DQN~\cite{atari}, as it is a discrete decision-making process with a low-dimension action space.
%For addressing the placement problem, which is essentially a discrete decision-making process, employing modern discrete DRL approaches, such as DQN~\cite{atari}, is a trivial solution.
In this work, we extend this approach by combining continuous scaling decisions with discrete placement actions, resulting in a hybrid action space.
However, jointly optimizing  discrete and continuous decision variables under a single DRL framework is challenging. The main reason is because discrete and continuous actions often require different policy representations. For discrete actions, a tabular or probability distribution representation is common, while for continuous actions, a Gaussian distribution or other parametric function is typically used. Reconciling these different representations can be difficult. Furthermore, defining a reward decomposition which transforms the reward signal to effectively guide both discrete and continuous action selection is a non-trivial task.
There have been a number of attempts to address this challenge, but there is no single, widely accepted solution.
Some approaches involve discretizing the continuous action space, which can lead to loss of information. Others involve using separate learned models for the discrete and continuous actions, which can make the learning process more complex~\cite{zhu2021overview}. %{\color{red}Furthermore, for the problem at hand we have deliberately employed a continuous action space for scaling decisions, to enhance robustness against increasing dimensionality~\cite{noms}. }

To this end, we  propose Deep Hybrid Policy Gradient (DHPG), which operates inherently in hybrid action spaces. DHPG follows the learning framework of DDPG~\cite{ddpg}, but with the following novel components:
($i$) joint state encoder $\mu_e$
which learns to extract compact and informative representations across various state spaces;
($ii$) specialized action heads
($\mu_p,\mu_C$)
that predict actions in different but correlated action spaces; and ($iii$) a new technique termed as probability-As-Action, which relaxes discrete actions into continuous representations, and thus allows E2E efficient learning under a unified framework.
%{\color{red}The state encoder $\mu_e$ and action head neural networks for placement and scaling decisions $\mu_p,\mu_C$, share weights $\theta^{\mu}$ to capture the correlations across action spaces.
 Fig.~\ref{fig:dhpg} illustrates the proposed framework. %, employing 2 action heads as an example.

\begin{figure}
    \resizebox{\columnwidth}{!}{%
        \begin{tikzpicture}
    % State A
    \node[draw,rectangle,
        fill=gray!20]
        (stateA)
        at (0,0) {State A};
    % State B
    \node[draw,rectangle,anchor=north,
        fill=gray!20]
        (stateB)
        at ($(stateA.south)-(0,.1)$)
        {State B};
    % State bounding box
    \draw
        ($(stateB.south west)-(.1,.1)$)
        --
        ($(stateB.south east)+(.1,-.1)$)
        --
        node[pos=.5] (state) {}
        ($(stateA.north east)+(.1,.1)$)
        --
        node[pos=.5] (stateN) {}
        ($(stateA.north west)+(-.1,.1)$)
        --
        ($(stateB.south west)-(.1,.1)$)
        ;

    % State encoder
    \node[trapezium,draw,align=center,
        anchor=east,
        shape border rotate=270,
        fill=Burlywood2]
        (enc)
        at ($(state.east)+(1.75,0)$)
        {State\\Encoder};

    % Joint feature
    \node[draw,align=center,
        fill=gray!20]
        (feat)
        at ($(enc.east)+(1,0)$)
        {Joint\\Feature};

    % Action heads
    \node[trapezium,draw,align=center,
        shape border rotate=90,
        trapezium right angle=80,
        trapezium left angle=80,
        fill=Burlywood2]
        (dhead)
        at ($(feat.east) + (1.25,.6)$)
        {\ \ \ Discrete\ \ \ \\Actor};
    \node[trapezium,draw,align=center,
        shape border rotate=90,
        trapezium right angle=80,
        trapezium left angle=80,
        fill=Burlywood2]
        (chead)
        at ($(feat.east) + (1.25,-.6)$)
        {Continuous\\Actor};

    % Actor bounding box
    \draw[dashed]
        ($(enc.south west |- chead.south east)-(.1,.1)$)
        --
        node[pos=0,below,
        anchor=north west] {Joint Actor}
        ($(chead.south east)+(.3,-.1)$)
        --
        ($(dhead.north east)+(.3,.1)$)
        --
        ($(enc.north west |- dhead.north east)+(-.1,.1)$)
        --
        ($(enc.south west |- chead.south east)-(.1,.1)$)
        ;

    % Joint action
    \node[draw,align=center,fill=gray!20]
        (action)
        at
        ($(feat -| chead.east) + (1,0)$)
        {Joint\\Action};

    % Joint Critic
    \node[trapezium,draw,align=center,
        shape border rotate=270,
        trapezium right angle=80,
        trapezium left angle=80,
        fill=Burlywood2]
        (critic)
        at
        ($(action.east) + (1,0)$)
        {Joint\\Critic};

    % Environment
    \node[draw, align=center,inner sep=5,
        fill=DodgerBlue3!30]
        (env)
        at ($(feat) + (0,2.5)$)
        {Environment};

    %%%%%%%%%%%%
    %% ARROWS %%
    %%%%%%%%%%%%
    % State to encoder
    \draw[->] (state.east) -- (enc.west);

    % encoder to joint feature
    \draw[->] (enc.east) -- (feat.west);

    % joint feture to dhead
    \draw[->] (feat.east) -- (dhead.west);
    % joint feture to chead
    \draw[->] (feat.east) -- (chead.west);

    % dhead to joint action
    \draw[->] (dhead.east) -- (action.west);
    % chead to joint action
    \draw[->] (chead.east) -- (action.west);

    % joint action to critic
    \draw[->] (action.east) -- (critic.west);

    % action to env
    \draw[->] (action.north) |-
        node[pos=0.8,below] {action}
        ($(env.east)-(0,.1)$);

    % env to critic
    \draw[->]  ($(env.east)+(0,.1)$) -|
        node[pos=0.3,above] {reward}
        (critic.north);

    % env to state
    \draw[->] (env.west) -|
        node[pos=.3,above] {state}
        (stateN);

    % critic to actor
    \draw[->] (critic.east) --
        ($(critic.east)+(.25,0)$)
        |-
        node[pos=.75,below] {update}
        ($(feat |- chead.south east)-(0,.5)$)
        --
        (feat |- chead.south east)
        ;

\end{tikzpicture}%
    }%
    \caption{Framework of DHPG. The states collected from the environment are first concatenated and sent to the state encoder to extract the joint feature. The feature is then fed to different actors to generate corresponding actions which are concatenated into a joint action. Finally the joint action is returned to the (i) environment and (ii) joint critic, along with the reward for learning. The details of the use of replay buffer and target networks are omitted for simplicity.} %DHPG is trained using the actor-critic method, which involves alternating between training the actor to maximize the expected cumulative reward and training the critic to improve its prediction accuracy. Note that the number of states/action heads can be an arbitrary number. This figure uses 2 action heads as an example.}
    \label{fig:dhpg}
    \vspace{-3mm}
\end{figure}

\noindent \textbf{Joint state encoder.} Traditional policy networks in DRL typically take a single state representation as input and output a probability distribution over a set of actions. This approach has been successful in a variety of tasks, but it can be limited in its ability to capture the complex interactions between multiple state components, particularly in tasks with hierarchical or heterogeneous structures. The joint state encoder $\mu_e$ takes a concatenated state representation $s_v$, which is a vector of state information from multiple components of the environment, as input. It then applies a series of neural network layers to extract a
joint feature representation
$z_v=\mu_e(s_v|\ \theta^{\mu})$
that captures the essential information about the state. Note that we adopt one-hot encoding for discrete states.

\noindent \textbf{Specialized action heads.} The joint feature representation is then passed to multiple dedicated actors, each responsible for generating actions for a specific component of the environment.
In the case of Problem~\ref{mdp}, the
dedicated actors $\mu_p,\mu_C$ take the
placement and scaling decisions,
respectively.
These actors are also neural network models
that take the joint feature representation
$z_v$ as input, and output:
($i$) a probability distribution
representing the chances of placing
a task to a specific \gls{pop}
$\mu_p(z_v|\ \theta^\mu)\in[0,1]^P$; 
and ($ii$) real values
over the corresponding continuous action space to
decide the number of \gls{cpu}s at
each \gls{pop} $\mu_s(z_v|\ \theta^\mu)\in\mathbb{R}^P$.
It is worth mentioning that the last
layer of the discrete actor $\mu_p$ is
a softmax layer
$\sigma:\mathbb{R}^P\to[0,1]^P$, thus
$\mu_p(z_v|\ \theta^\mu)\in[0,1]^P$.
The actions generated by the actors are concatenated into a joint action vector
$\hat{a}_v=[\mu_p(z_v|\ \theta^\mu),\mu_s(z_v|\ \theta^\mu)]$.
% and sent to the environment for execution.
The joint action $\hat{a}_v$
is passed to the joint critic which estimates the expected cumulative reward given the joint action and state.

\noindent \textbf{Probability-as-action.} The traditional approach handling discrete action spaces relies on sampling an action from a probability distribution, while continuous action spaces involve generating a real-valued action from a parametric function. This difference leads to incompatible approaches in optimization. To address these limitations, we propose a novel approach called probability-as-action (PAA),
which takes the softmax-normalized vector directly as the action, rather than using a sampled action.
% within the joint action $\hat{a}_v$.
This continuous relaxation of the discrete action allows for more nuanced policy representation and seamless integration with continuous action space algorithms.
In PAA, the discrete actor produces a softmax-normalized vector
$\mu_p(z_v|\ \theta^\mu)\in[0,1]^P$
representing the likelihood of each discrete action
-- e.g. the likelihood of placing the
vehicle task at \gls{pop} $1,\ldots,P$.
This vector is directly used as the action representation
passed to the critic.
However, the actual action $a_v$ sent to
the environment is randomly
selected according to the softmax-normalized vector.
Hence, the same joint action $\hat{a}_v$ may
result in different actual actions
$a_v\neq a'_v$ during
the training stage. Finally, since the reward is no longer in response to a specific sampled action but a probability distribution, the discrete actor $\mu_p$ ends up learning
to generate a probability distribution that leads to higher expected cumulative rewards.
% Since the action is now a continuous vector instead of a discrete value, the reward function also needs to be redefined accordingly. In traditional DRL with discrete actions, the reward is assigned to the chosen action.
% Here, we define the reward as the expected reward of all possible actions sampled from the probability distribution. 
% This approach is consistent with the idea that the actor is learning to generate probability distributions that lead to higher expected cumulative rewards.

\bigskip 
Overall, the \gls{dhpg} agent comprises discrete and continuous action heads
$\mu_p,\mu_C$ taking the placement and
scaling decisions jointly.
In the following we explain how the
actual action
$a_v=(p'_v,C_{1,v}^+,\ldots,C_{P,v}^+)$
is taken.
During the training stage, the placement
decision is chosen using a random sampling
function $X:[0,1]^P\to \mathbb{N}\cap[1,P]$
over the softmax-normalized vector
-- i.e. $p'_v=X(\mu_p(z_v|\ \theta^\mu))$.
In the testing stage, the placement decision
is taken based on the \gls{pop} reporting
the highest likelihood -- i.e.
$p'_v=\argmax \{\mu_p(z_v|\ \theta^\mu)\}$.
The scaling decisions are produced after applying
\gls{dod}~\cite{noms} for the output of the
continuous action head, that is, the
\gls{cpu} scaling is obtained as
$(C_{1,v}^+,\ldots,C_{P,v}^+)=
\text{DOD}(\mu_C(z_v|\ \theta^\mu))$.
Note, however, that the training stage
uses the continuous output (prior to the
\gls{dod}) to feed the critic network.
Algorithm~\ref{alg:dhpg} describes the \gls{dhpg} learning
procedure.

\begin{algorithm}[t]
\caption{Learning of DHPG}\label{alg:dhpg}
\begin{algorithmic}[1]
\State Initialize joint actor $\mu$ with parameters $\theta^\mu$ 
\State Initialize joint critic $Q$ with parameters $\theta^Q$
\State Initialize target network $\mu^\prime \gets \mu$ and $Q^\prime \gets Q$
\State Initialize replay buffer $B$
\State \textbf{for $episode = 1, E$ do}
    \State \ \ \ \ Receive initial state $s_1$
    \State \ \ \ \ \textbf{for $v = 1, V$ do}
        \State \ \ \ \ \ \ \ \ Extract joint latent representation $z_v = \mu_e(s_v|\theta^\mu)$
        %\State \ \ \ \ \ \ \ \ Predict placement probability $e^{pla}_{v} \gets \mu_{pla}(z_v|\theta^\mu)$
        \State \ \ \ \ \ \ \ \ Obtain placement probability $y_p = \mu_p(z_v|\theta^\mu)$
        % \State \ \ \ \ \ \ \ \ Predict scaling action $e^{sca}_{v} \gets \mu_{sca}(z_v|\theta^\mu)$
        \State \ \ \ \ \ \ \ \ Obtain scaling action $y_C = \mu_C(z_v|\theta^\mu)$
        \State \ \ \ \ \ \ \ \ Obtain joint action $\hat{a}_v = \left[ \sigma(y_p),y_C \right]$
        \State \ \ \ \ \ \ \ \ Obtain executable actions:
            % \State \ \ \ \ \ \ \ \ \ \ \ \ $a^{pla}_{v} \sim e^{pla}_{v}$
            % \State \ \ \ \ \ \ \ \ \ \ \ \ $a^{sca}_{v} \gets DOD(e^{sca}_{v})$
        \State \ \ \ \ \ \ \ \ \ \ \ \ $p'_v = X\left(\sigma(y_P)\right)$\label{line:randomPoP}
        \State \ \ \ \ \ \ \ \ \ \ \ \ $(C_{1,v}^+,\ldots,C_{P,v}^+)= \text{DOD}(y_C)$
        % \State \ \ \ \ \ \ \ \ Execute actions $a^{pla}_{v}$ and $a^{sca}_{v}$ 
        \State \ \ \ \ \ \ \ \ Execute actions $p'_{v}$ and $(C_{1,v}^+,\ldots,C_{P,v}^+)$ 
        \State \ \ \ \ \ \ \ \ Observe reward $r_v$ and next state $s_{v+1}$
        \State \ \ \ \ \ \ \ \ Store transition $(s_v, \hat{a}_v, r_v, s_{v+1})$ in $B$
        \State \ \ \ \ \ \ \ \ Sample a random batch $b$ of transitions in $B$
        \State \ \ \ \ \ \ \ \ Set $y_v = r_v  +\gamma Q'(s_{v+1},\mu'(s_{v+1})), v \in b$
        \State \ \ \ \ \ \ \ \ Update Critic using Bellman error:
        \State \ \ \ \ \ \ \ \ \begin{equation*}
            \quad\delta\theta^Q\!\! = \frac{1}{|b|}\sum_{v\in b}\ \!\!\nabla_{\theta^Q}\ \!\!\big[y_v -Q(s_v,a_v) \big]^2
            \end{equation*}\label{line:critic-update}
        % \State \ \ \ \ \ \ \ \ \begin{equation*}
        %     \ \qquad\delta\theta^Q\!\! =\!\!\nabla_{\theta^Q}\!\!\sum_{v\in b}\big[r_v  +\gamma Q'(s_{v+1},\mu'(s_{v+1})) -Q(s_v,a_v) \big]^2
        %     \end{equation*}\label{line:critic-update}
        \State \ \ \ \ \ \ \ \ Update Actor using policy gradient:
        \State  \ \ \ \ \ \ \ \ \begin{equation*}
            \quad \delta\theta^\mu= -\frac{1}{|b|}\sum_{v\in b} \nabla_a Q(s_v,a) \nabla_{\theta^\mu} \mu(s_v)
            \end{equation*}  
        % \State \ \ \ \ \ \ \ \ Compute cumulative rewards $y$ using target networks
        % \State \ \ \ \ \ \ \ \ Update $\theta^Q$ using mean-squared Bellman error with $y$
        % \State \ \ \ \ \ \ \ \ Update $\theta^\mu$ using the sampled policy gradient
        % \State \ \ \ \ \ \ \ \ Update Critic and Actor:
        %     \State \ \ \ \ \ \ \ \ \ \ \ \ $\theta^Q\gets\theta^Q+\delta\theta^Q$
        %     \State \ \ \ \ \ \ \ \ \ \ \ \ $\theta^\mu\gets\theta^\mu+\delta\theta^\mu$
        \State \ \ \ \ \ \ \ \ Update target networks:
            \State \ \ \ \ \ \ \ \ \ \ \ \ $\theta^{Q^\prime} \gets \tau\theta^Q + (1-\tau)\theta^{Q^\prime}$
            \State \ \ \ \ \ \ \ \ \ \ \ \ $\theta^{\mu^\prime} \gets \tau\theta^\mu + (1-\tau)\theta^{\mu^\prime}$
    \State \ \ \ \ \textbf{end for}
\State \textbf{end for}
\end{algorithmic}
\end{algorithm}

\begin{figure}
    \centering
    \subfloat[\centering Proposed \gls{dhpg} agent]{{\includegraphics[width=0.85\columnwidth]{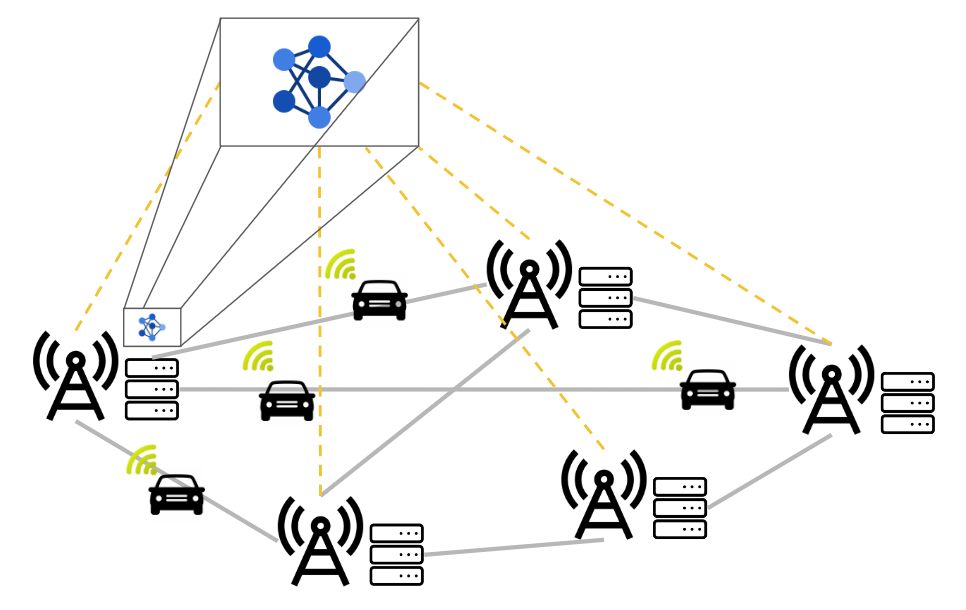}}}
    \\
    \subfloat[\centering Decentralized \gls{ddpg} agents]{{\includegraphics[width=0.85\columnwidth, clip, trim = 0cm 0cm 0cm 3cm]{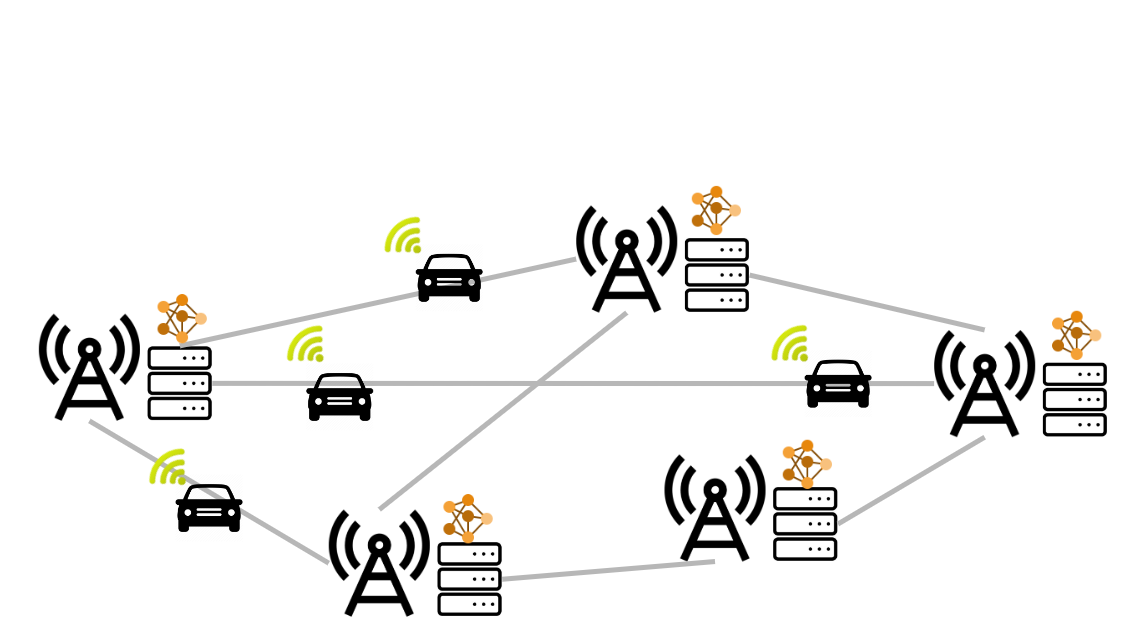}}}

    % \subfloat[\centering Proposed centralized \gls{dhpg} agent]{\input{Figures/road_graph_central}}

    % \subfloat[\centering Decentralized \gls{ddpg} agents~\cite{noms}]{\input{Figures/road_graph}}
    
    \caption{The proposed \gls{dhpg} agent vs.
    decentralized \gls{ddpg} agents~\cite{noms} for C-V2N service provisioning.
    %(a) The centralized \gls{dhpg} agent
     %decides which \gls{pop} to process
     %incoming C-V2N tasks, and scale computational resources for all \gls{pop}s.
     %(b) The decentralized \gls{ddpg} agent
     %takes scaling decisions for a single
     %\gls{pop} according to the C-V2N
     %tasks assigned to it.
    %(a) In a centralized agent architecture, a single entity is responsible for managing all C-V2N communications and services. This entity maintains a comprehensive view of the C-V2N ecosystem and makes decisions for service provisioning based on the global knowledge. (b) In a decentralized agent architecture, multiple autonomous agents operate independently based on the respective local information.
    }
    \label{fig:ddpg-n}
\end{figure}

Lastly, it is worth mentioning that
our solution DHPG comprises an end-to-end joint DRL approach
that determines the
placement decision $p'_v$, and
a scaling vector $(C_{1,v}^+,\ldots,C_{P,v}^+)$
%a scaling action $a_v$ that specifies
that specifies
the \gls{cpu} increment at each \gls{pop}
%$a_v=(C_{1,v}^+,\ldots,C_{P,v}^+)$
based on the observed state $s_v=(p_v, N_{1,v},C_{1,v},\ldots, N_{P,v},C_{P,v})$
across all \glspl{pop}.
% Overall, the \gls{ddpg}-based solution that we propose
% receives as input the state of each \gls{pop}
% $s_v=(N_{1,v},C_{1,v},\ldots, N_{1,v},C_{1,v})$,
% and outputs a scaling action $a_v$ that specifies the
% \gls{cpu} increment at each \gls{pop}
% $a_v=(C_{1,v}^+,\ldots,C_{P,v}^+)$.
%We make the hypothesis that the scaling problem becomes challenging  when we take decisions for each \gls{pop}, all at once.
%use the state of all \gls{pop}s,
%and take decisions for each \gls{pop}, all at once.
%Therefore, in the performance evaluation
Note that \gls{dhpg} differs from the
decentralized \gls{ddpg} approach
in~\cite{noms},
where a dedicated \gls{ddpg} agent is in charge of scaling decisions at each \gls{pop}.
% where vehicles' tasks
% are processed at the \gls{pop} with the minimum average experienced delay (as described in formula~\eqref{eq:greedy-assign}),
%are processed at the least loaded \gls{pop},
% with a dedicated \gls{ddpg} agent in charge of scaling decisions.
Fig.~\ref{fig:ddpg-n} highlights the
difference between the proposed
\gls{dhpg} solution, which utilizes
%centralized information aggregation with decentralized processing, 
global state information aggregated at each agent/PoP running the \gls{dhpg} approach,
versus the purely decentralized \gls{ddpg} scaling
agent from~\cite{noms} coupled with greedy-based placement agent.
In Section~\ref{sec:evaluation} we
compare the performance of both approaches.

\section{V2N Application Service Simulation}
\label{sec:simulation}

%{\color{red}@Kote + @Jorge + @Cyril: to check first}

%{\color{red}@Danny + @Chrysa + @Luca: to review afterwards}

% Intro paragraph
Following, we describe the simulation environment that we implemented to evaluate the performance of our approach, and describe the dataset used to drive the simulations. %The environment is also used by the alternative agents presented in Section~\ref{sec:other-solutions}.

\subsection{Simulation Environment}
\label{subsec:environment-logic}

%% \begin{figure}[t]
%%     \centering
%%     \input{Figures/stacked-environment.tex}
%%     \caption{The \emph{environment} class reads from the dataset
%%     that vehicle 5 arrived to \gls{pop}~1 at time $t_5$,
%%     and informs the agents that \gls{pop}~1 already has
%%     $C_{1,4}=1$~CPUs processing the \gls{cv2n} tasks of
%%     $N_{1,4}=3$~vehicles.
%%     The agents place vehicle~5 tasks
%%     to \gls{pop}~1 and 
%%     scale up $C_{1,5}^+=1$ CPU to get the
%%     corresponding reward $R(\cdot)$.}
%%     % \caption{The \emph{environment} class retrieves from the dataset the arrival of vehicle $v=5$ to \gls{pop}~1 at time $t_{5}$. Then, each {\em \gls{pop} queue} reports its status $s_{5,p}$, so that the agent can take action $a_{v_5}$ based on the global state $s_{v_5}$. The    \emph{environment} forwards the scaling $C_p^+$ and steering
%%     % $p_{v_5}'$ decisions to the {\em \gls{pop} queues}, and returns
%%     % the corresponding reward $R(s_{v_5},a_{v_5})$ to the agent.\todo[inline]{Jorge: put stacked agents
%%     % on top of each PoP, and they should send the
%%     % placement \& scaling action. Think how to
%%     % illustrate that both things are done. On top,
%%     % write down the infor. exchanged, namely, 
%%     % number of CPus and number of attended vehicle.s
%%     % }}
%%     \label{fig:environment}
%% \end{figure}

The simulation environment~\cite{5growth-scaling}
%\footnote{\url{https://github.com/MartinPJorge/5growth-scaling/blob/master/FiftyStations/clean0326/FiftyStationsEnvironment_v3.py}}
implements the \gls{sarsa}~\cite{sutton2018rl} logic
inherent to a~\gls{mdp}, i.e., it implements the functions required to perform actions~$a_v$, to monitor the state~$s_v$,
and to report the rewards~$R(s_v,a_v)$ as the simulation progress. %Thanks to the developed environment, we also have a common frame to compare the performance of the proposed \gls{ddpg} agent against state of the art solutions of Section~\ref{sec:other-solutions}. 
It is implemented in Python and consists of two main
classes: ($i$) a class simulating the operations within the MEC node at each \gls{pop}, named~\emph{\gls{pop} queue}; and ($ii$) a class simulating and monitoring the metro environment, named \emph{environment}.
%{\color{red} For the sake of simplicity, in our implementation the placement and scaling agents are not part of the PoP classes and are called independently by the environment. However they are considered part of each PoP.}
% Furthermore we simulate the selected \gls{cv2n} application, in order to be able to derive the corresponding processing delay and workload at each PoP.
% As mentioned in Section \ref{sec:problem} we employ an \textit{Advanced
% Driving} \gls{v2n} application~\cite{3GPPv2xrequirements,v2x-tutorial} where each vehicle $v\in V$ produces a video sequence to be processed at the edge. Specifically, each frame (task) is decoded and analysed at a \gls{pop} $p_v\in P$ and the result of the analysis is sent back to the vehicle
% -- e.g., pressing the break pedal.
% According to~\cite{3GPPv2xrequirements,v2x-tutorial}
% Advanced Driving services require that
% the latency requirement (between 10~ms and 100~ms)
% is met 99\% of the times.
% For the reward function in~\eqref{eq:reward}
% resorts to the average processing delay,
% the DDPG agents would foster meeting the
% average delay if we just set $d_{tgt}=100$\,ms.
% Rather, we rely on Lemma~\ref{lemma:kappa} and
% run our experiments replacing the target delay
% $d_{tgt}$ in~\eqref{eq:reward} by
% $\tfrac{d_{tgt}}{K(99)}=
% \tfrac{100}{2}=50$\,ms. As a result, our
% experiments foster DDPG agents 
% meeting the $\kappa=99$ delay requirement 
% of 100\,ms.
In the following, we explain in detail the classes of the simulation environment and their interactions -- see Algorithm~\ref{alg:env}.

\begin{table}[t]
    \centering
    \caption{Time to decode $\tau_d$~\cite{Alvarez-Mesa2012}
    and analyze $\tau_a$~\cite{SHUSTANOV2017718} a frame (f)
    using H.265/HEVC streams~\cite{ETSI.TR.126.985}, and $C_p$ Intel Xeon CPUs.
    Both contributions yield the \gls{v2n}
    frame processing rate~$\mu$.}
    \begin{tabular}{c l l l l l}
        \toprule
        &  $\bm{C_p=1}$ & $\bm{C_p=2}$ &$\bm{C_p=3}$ &$\bm{C_p=4}$ &$\bm{C_p=5 }$\\\midrule
        $\tau_d$ & 8.47~ms/f & 4.41~ms/f & 3.05~ms/f & 2.37~ms/f & 2.03~ms/f\\
        $\tau_a$ & 37~ms/f & 18.50~ms/f & 12.33~ms/f & 9.25~ms/f & 7.40~ms/f\\
        $\mu$ & 0.02~f/ms & 0.04~f/ms & 0.06~f/ms & 0.09~f/ms & 0.11~f/ms \\\bottomrule
    \end{tabular}
    \label{table:processing-rates}
\end{table}

\begin{algorithm}[t]
\caption{Simulation Environment}\label{alg:env}
\begin{algorithmic}[1]
\State Initialize environment env
\State Initialize total reward $r_t \gets 0$
\State $s_v, r_v, \text{stop} \gets \text{env.step}()$
\While{not stop}
    % \State $p_v' \gets A_p(p_v, s_v)$ \Comment{placement}
    % \State $a_v \gets A_s(s_v)$  
    \State $r_t \gets r_t + r_v$
    \State $p_v', a_v \gets \text{DHPG}(s_v)$
    % \State $r_t \gets r_t + r_v$
    % \State $env \gets env.update(p_v', a_v)$
    \State $s_v, r_v, \text{stop} \gets \text{env.step}(p_v', a_v)$
\EndWhile

\State \Return $r_t$
\end{algorithmic}
\end{algorithm}

% \begin{algorithm}
%     \DontPrintSemicolon 
%     \SetKwBlock{DoParallel}{do in parallel}{end}
%     \KwIn{Some inputs}
%     \KwOut{The ouput}
%     \DoParallel{
%         Compute a \\;
%         Compute b \;
%         Compute c \;
%     }
%     \DoParallel{
%         a1\;
%         b1\;
%         c1\;
%     }
%     \Return{the solution}\;
%     \caption{Parallel Algo}
%     \label{algo:parallelAlgorithm}
% \end{algorithm}

% Explain how the environment works
\vspace{1mm}
\noindent \textbf{\gls{pop} queue class.} The \textbf{\emph{\gls{pop} queue}} class simulates the operations at \gls{pop}~$p$ level. Upon the arrival of vehicles $v\in V$ it keeps track of the number of active CPUs $C_{p,v}$, how many vehicles $N_{p,v}$ are attended by the \gls{pop}, what is its current load~$\rho_{p,v}$, and what is the resulting processing latency~$d_{p,v}$. 
The \emph{\gls{pop} queue} class derives the processing delay and load as follows:
\begin{itemize}
    \item \textit{Processing delay}: to compute $d_{p,v}$ using formula~\eqref{eq:avg-delay}, we have to obtain the processing rate of the \gls{pop} $\mu(C_{p,v})$. This rate is derived considering the video decoding $\tau_d$ and analysis contributions $\tau_a$. Specifically, we estimate the processing rate as $\mu(C_{p,v}) = \tfrac{1}{\tau_d(C_{p,v}) + \tau_a(C_{p,v})}$, with $\tau_d$ set using the empirical evidence presented in~\cite{Alvarez-Mesa2012}, and $\tau_a$ as presented in~\cite{SHUSTANOV2017718}. Table~\ref{table:processing-rates} summarizes the processing rates $\mu$, as well as the decoding and analysis contributions.
    \item \textit{Load}: the current load per PoP is estimated using the number of served vehicles~$N_{p,v}$ and active CPUs~$C_{p,v}$. In particular, the load is set to \mbox{$\rho_{p,v}=\lambda N_{p,v}/\mu(C_{p,v})$}, where $\lambda=29.5$~fps per H.265/HEVC stream (ETSI~\cite{ETSI.TR.126.985} \gls{v2n} services).
\end{itemize}

\noindent \textbf{Environment class.} The \textbf{\emph{environment}} class keeps track of the set $P$ of \emph{\gls{pop} queue class instances} and interacts with them to monitor the system state~$s_v$, take actions~$a_v$, and report the obtained rewards~$R(s_v,a_v)$. All such interactions are driven by the instants at which vehicles arrive
$\{t_1,t_2,\ldots,t_V\}$ according to the used dataset -- see Section~\ref{subsec:dataset}. In the following we enumerate the \emph{environment} interactions:
\begin{enumerate}[label=\roman*)]
    \item Upon the arrival of a vehicle $t_v$, the \emph{environment} class iterates over  \emph{\gls{pop} classes} and removes those vehicles $v'$ that already left each \gls{pop} $p$, i.e., those with $T_{v'}<t_v$.
    \item The \emph{environment} reports the system state $s_{v}$ in~\eqref{eq:state} by retrieving from each \emph{\gls{pop} queue} the number of vehicles it is serving~$N_{p,v}$, and the number of active \glspl{cpu} $C_{p,v}$.% -- see~Fig.~\ref{fig:environment} continuous-line arrow.
 % \item   {\color{red}   The placement and scaling agents per PoP take the state~$s_{v}$ and send the \emph{environment} class an action $a_{v}$ that contains the placement $p_v'$ and scaling decisions $C_{p,v}^+, \forall p\in P$.}
 \item   The DHPG agent takes the state~$s_{v}$ across all PoPs and returns an action $a_{v}$ that contains the placement $p_v'$ and scaling decisions $C_{p,v}^+, \forall p\in P$ to the \emph{environment} class.% -- see~\eqref{eq:action} and Fig.~\ref{fig:environment} dashed-line arrow; 
    \item The \emph{environment} class assigns the vehicle $v$ and places its corresponding tasks to the \emph{\gls{pop} queue} $p_v'$ and sets its departure time as $T_v=t_v+30k$, with $k\sim \text{Exp}(\Lambda)$ drawn from a exponential distribution with rate \mbox{$\Lambda=1$}, i.e., on average vehicles linger for 30~sec in the road. Then, the \emph{environment} iterates over each \emph{\gls{pop} queue} to scale its \glspl{cpu} according to the received action $C_{p,v}^+$. In case the vehicle $v$ is processed by another \gls{pop} (i.e., $p_v'\neq p_v$) the \emph{environment} accounts for the transmission delay  that vehicle $v$  will experience; the delay overhead is  set to
    %{\color{red}
    $l_{p_v,p_v'}=20$~ms, which equals to two times the one-way E2E latency
    %}
    (i.e., UL+DL) for 5G networks \cite{3GPPv2xrequirements}~\cite{5G_Latency} 
(as shown in formula~\eqref{eq:total-delay}).
\item The \emph{environment} iterates over each \emph{\gls{pop} queue} and computes the reward per \gls{pop} $R(s_{v,p},a_{v})$ using formula \eqref{eq:reward} and vehicles' experienced delay given by formula \eqref{eq:total-delay}. Then, the average reward by all~\glspl{pop} is sent to the agents $\tfrac{1}{P}\sum_{p\in P} R(s_{v,p},a_{v})$.
%-- denoted as $r_v$ for simplicity over the dotted-line arrow in Fig.~\ref{fig:environment}.
Maximizing this average reward is the goal of the \gls{mdp} described in~Problem~\ref{mdp}.
\end{enumerate}
% The steps above are sketched in Fig.~\ref{fig:environment} and Algorithm~\ref{alg:env}, where we see that the \emph{environment} is fed with the arrival times of the vehicles in the area covered by the \glspl{pop}.
The steps above are summarized in Algorithm~\ref{alg:env}.
In the next section, we detail the dataset used to obtain the arrival time of vehicles $t_v$,
and each vehicle's associated \gls{pop} $p_v$.

\subsection{Vehicular Mobility Dataset }
\label{subsec:dataset}

\begin{figure}
    \centering
    \includegraphics[width=.8\columnwidth]{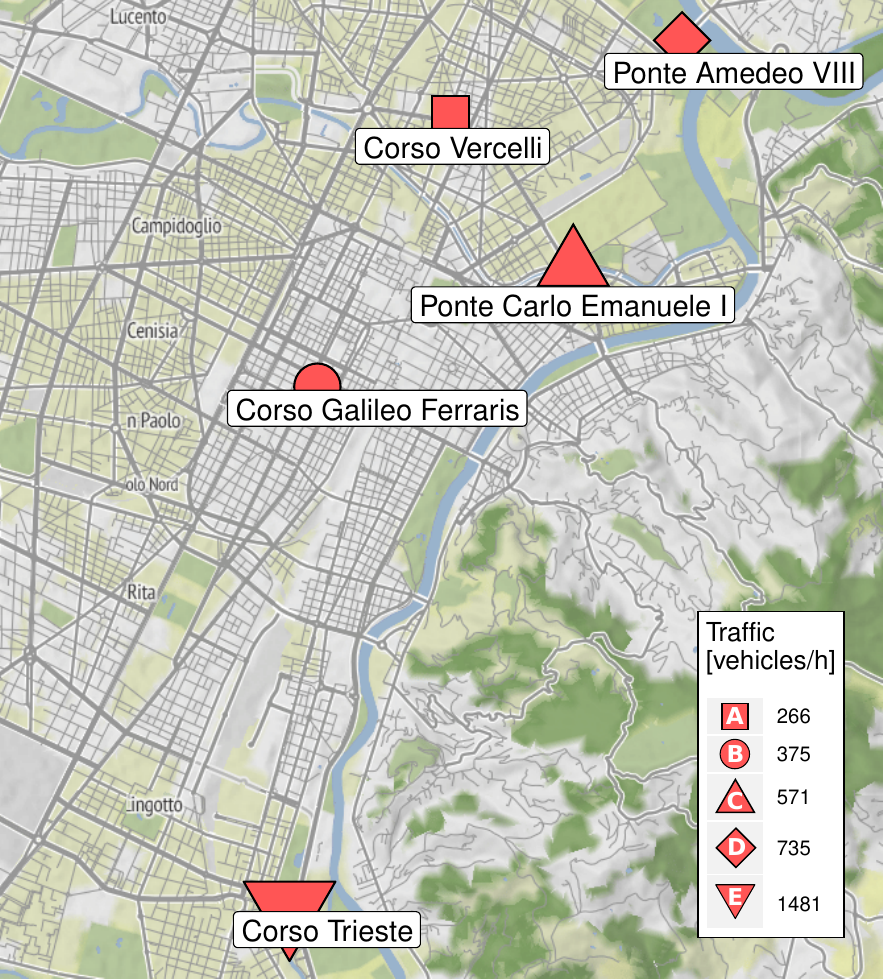}
    \caption{ \glspl{pop} across Turin with % in the environment dataset (Turin city).
    the average car traffic load.}
    \label{fig:torinoMap}
\end{figure}

In order to feed the environment with realistic traffic we collected a dataset spanning from January to October 2020 of vehicle traffic gathered in Turin (Italy). There are over 100 measuring locations distributed throughout Turin where the number of vehicles that pass each particular measuring point is counted over 5~minute intervals. At the end of every interval, the traffic intensity (i.e., the number of vehicles arriving in that 5~minute interval divided by 5~minutes) is expressed as a number of vehicles per hour.
% Not all measuring points report traffic intensities for each 5~minute interval.
Not all measuring points do report traffic intensities for each 5~minute interval,
hence we
interpolate the missing measurements via expectation maximization.
Additionally, we consider
$P=5$ \glspl{pop} to find optimal solutions
of Problem~\ref{problem} using commodity
solvers in Section~\ref{subsec:optimality-gap}.
For $P>5$, finding optimal solutions becomes intractable due to the fast growth
of the search space in
Problem~\ref{problem}.
%We selected the $P=5$ stations with the maximum amount of measurements and interpolated the missing measurements (via expectation maximization).
Fig.~\ref{fig:torinoMap} shows how the measuring points (which we consider to be as \glspl{pop}) are distributed across the city.

This dataset provides us the traffic intensity
\mbox{$\Lambda_p(t)$~[vehicles/hour]} at each
\gls{pop} $p \in \{1, \ldots, P\}$ every 5~minutes.
We use these time-varying $\Lambda_p(t)$ values as the arrival rates of $P$ independent Poisson processes that model vehicle arrivals on instants $t_v$ at the $P$ \glspl{pop}. Namely, at each time window $[t,t+5\text{min}]$ we iterate over each \gls{pop}
$p$ and randomly generate its car arrivals as
\begin{equation}
    \{t_1, t_2, \ldots\} =\{t_i\}_i =  \left\{t + \sum_{j=0}^i k_j \right\}_i,\quad k_j\sim \text{Exp}\left(\Lambda_p(t)\right)
    \label{eq:car-arrivals}
\end{equation}
with $\text{Exp}\left(\Lambda_p(t)\right)$ denoting the exponential distribution with rate $\Lambda_p(t)$. By repeating this random generation of arrivals %in~\eqref{eq:car-arrivals} 
 for every \gls{pop} $p\in P$, and for every time window of 5~min; we generate a dataset with samples
$\{ (t_1, p_1), (t_2, p_2),\ldots \}$ (time $t_v$ of vehicle arrival at \gls{pop} $p_v$).
%This dataset drives the simulation -- see $(t_v,p_v)$ rows Fig.~\ref{fig:environment}.
To gain statistical significance,
we repeat 40 times the random generation of arrivals, using formula~\eqref{eq:car-arrivals} with different seeds.
As a result, we obtain 40 different traces that express how vehicles arrive to each \glspl{pop} in Fig.~\ref{fig:torinoMap} between
the 28\textsuperscript{th} of January 2020,
and the 1\textsuperscript{st} of October 2020.

Remark that the considered \glspl{pop} exhibit increased traffic intensity $\Lambda_p(t)$ during morning and afternoon rush hours, a pattern also observed in datasets from other cities, such as PeMSD4 (San Francisco Bay Area), PeMSD7 (District 7 of California), and PeMSD8 (San Bernardino)~\cite{fiorevehdata,gao2023spatial,wavelets-traffic}. Consequently, the performance of \gls{dhpg} evaluated in Section~\ref{sec:evaluation} is expected to generalize to these cities.
% \gls{dhpg} would still capture
% different traffic trends
% -- as in the PeMSD7(L)
% dataset~\cite{wavelets-traffic}
% -- as long as it is trained using the
% corresponding traffic
% data during its training stage, i.e.
% upon the execution of Algorithm~\ref{alg:dhpg}.
Specifically, \gls{dhpg} can capture diverse traffic trends, as seen in the PeMSD7(L) dataset~\cite{wavelets-traffic}, provided it is trained on the corresponding traffic data during the training phase.
% i.e., when executing Algorithm~\ref{alg:dhpg}.

\section{Performance evaluation} 
\label{sec:evaluation}

\begin{table}[t]
\caption{Training and testing traces.}
\label{table:traces}
\resizebox{\columnwidth}{!}{%
\begin{tabular}{ c l l l l }
 \toprule
 \textbf{trace} & \textbf{start} & \textbf{end} & \textbf{duration} & \textbf{arrivals} \\
 \midrule
 training   & Jan 28 17:55 & Jan 29 07:13 & 13.3 hrs & 28352\\
 testing   & Jan 29 07:13 & Jan 29 12:43 & 5.5 hrs & 28353\\
 \bottomrule
\end{tabular}
}
\end{table}
In this section we evaluate the performance of the proposed approach on the task placement and scaling problem. % and compare it against the solutions that are coupled with existing scaling approaches ( Section~\ref{sec:other-solutions}). 
% {\color{red}Specifically, while employing the same greedy placement strategy as defined in formula (\ref{eq:greedy-assign}), we study the performance of the decentralized approach \gls{ddpg} as well as the centralized DHPG (Fig.~\ref{fig:ddpg-n}) and compare them against state-of-the-art scaling approaches. For the sake of simplicity, we denote the solutions using their associate scaling methods, as they share the same placement strategy. }
%{\color{red}
Specifically, we study the performance of the DHPG as well as the decentralized scaling approach \gls{ddpg} (as shown in Fig.~\ref{fig:ddpg-n}), and compare them against other state-of-the-art scaling approaches. For the sake of simplicity, we denote the solutions using their associate scaling methods, as they share the same greedy-based placement strategy defined in formula (\ref{eq:greedy-assign}).
%}

\subsection{Comparison Solutions}
\label{ScalingApproaches}
We compare DHPG to fully or partially decoupled solutions, where the placement decision is followed by scaling, as presented in \cite{noms}. Specifically, we employ a greedy placement algorithm placing vehicle $v$ tasks
to the \gls{pop} $p'_v$ that minimizes the average
experienced delay, i.e., the transmission and
processing latency:
\begin{equation}
    p'_v = \argmin_{p\in P} \left( \mathbb{E}[l_{p_v,p}] + \mathbb{E}[d_p] \right)
    \label{eq:greedy-assign}
\end{equation}

\noindent Scaling is tackled with the following approaches \cite{noms},\cite{DeVleeschauwer2021} \cite{v2n-access}:
\setlist{nolistsep}
\begin{itemize}[noitemsep,leftmargin=*]
    \item \textbf{Constant (CNST)} approach as baseline in which the number of active CPUs is constant over time. Exhaustive search is used to determine the number of CPUs per \gls{pop} that brings the highest reward on the training set $V'$:
    \begin{equation}
        \argmax_{(C_1, \ldots C_P)} \sum_{v\in V'}R(C_p, p_v', t_v)
        \label{eq:cnst}
    \end{equation}
    Note that $C_{p,v}=C_p,\ \forall v\in V'$ with $C_p$ a constant.
\item \textbf{\gls{pi}} controller \cite{Ang05pidcontrol}, with parameters $\alpha$ and $\beta$, 
    that aims to keep the most loaded CPU below a selected threshold of $\rho$ \cite{DeVleeschauwer2021}; %Therefore, it first calculates \mbox{$\delta(t) = \alpha (\rho(t) - \rho_{tgt}) + \beta (\rho(t) - \rho(t-1))$} and then adjusts the number of CPUs for the next slot to \mbox{$N(t+1) =  N(t) + \delta(t)$}, suitably bounded below by a minimum number of CPUs and above by a maximum number of CPUs. Via simulations on the training trace we concluded that values $\rho_{tgt}=0.6$, $\alpha=\beta=5.0$ bring the highest average reward. 
\item \textbf{\gls{tes}} scaling algorithm, that is an adaptation of the \emph{n-max} solution in~\cite{v2n-access}, that creates an equally spaced time series $f_{p,t-S},\ldots,f_{p,t-1}, f_{p,t}$ corresponding to the flow (i.e., the number of vehicles) that each \gls{pop} $p$ has received in the last $S$ slots, each of $m=5$ minutes. Based on this flow time series, the algorithm uses \gls{tes}~\cite{Winters1960TES} to predict the maximum traffic flow in a window of $W\cdot m$ minutes, with $W$ being the number of
time slots to forecast.
Then, the number of \glspl{cpu} at each \gls{pop} $p$ is scaled according to the \gls{tes} forecast.

\item \textbf{Deep Deterministic Policy Gradient (\gls{ddpg})} is a model-free, off-policy reinforcement learning algorithm that operates in environments with continuous action spaces~\cite{ddpg}. It utilizes an actor-critic architecture, where the actor learns a deterministic policy that maps states to actions, while the critic estimates the expected cumulative future reward given a state and action. Authors in~\cite{noms} proposed DDPG-DOD, a variant of DDPG, which exploits the underlying structure of the discrete action space, enabling more efficient learning and improving scalability in resource scaling problems. DDPG-DOD is referred to hereafter as DDPG for simplicity.
    
\end{itemize}
%\bigskip

\subsection{Experimental Setup}
\label{sec:expsetup}

All solutions are evaluated with the simulation environment presented in Section~\ref{sec:simulation}, where we use Python 3.9.12, complemented with the PyTorch 1.11.0 library. We employ the data-set described in Section~\ref{subsec:dataset} in an area of 5 \gls{pop}s. 
% The details of the traces selected for training and testing are summarized in Table~\ref{table:traces}.
% Note that we consider a trace that is long enough to cover various traffic patterns, including peak/off-peak hours.
% To have statistical significance in our results, we use 40 different traces for each experiment that span through the training or testing period; each containing the arrival time $t_v$ and \gls{pop} $p_v$ of every vehicle.
To run the experiments we use a server with Intel Core i7-10700K \gls{cpu}, 32 GB of RAM.

\subsubsection{Hyperparameters}
\label{subsec:parameter}
In the following, we give the detailed parameter settings for the algorithms mentioned in Section~\ref{sec:ddpg} and~\ref{ScalingApproaches}. %Note that the placement agent has no learnable parameters.

\noindent \textbf{CNST.} We perform a grid search to find the optimal number of \gls{cpu}s for the selected \glspl{pop} (five in total) in formula (\ref{eq:cnst}), namely sweeping over $C_p$ %$ \in \{0, 1, 2, 3, 4, 5\}$, 
where $p \in \{1, 2, 3, 4, 5\}$, resulting in $6^5=7776$ combinations. The best found combination over the training set is $(C_A, C_B, C_C, C_D, C_E)=(1, 5, 5, 1, 5)$.

\noindent \textbf{PI.} We perform a grid search to find the optimal parameters $\alpha$, $\beta$ and $\rho_{tgt}$ within the ranges suggested by the authors in~\cite{DeVleeschauwer2021}. The best found result over the training set is $(\alpha, \beta, \rho_{tgt})=(4, 0, 0.7)$.

\noindent \textbf{TES.} \gls{tes} is an adaptive algorithm that does not have a training phase. We set $W \cdot m = 1$ second.

% \noindent \textbf{DDPG.} \gls{ddpg}-1 and \gls{ddpg}-5 have similar hyper-parameter settings: the architectures of actor and critic are both a 3-layer perceptron. Each layer contains 64 neurons for \gls{ddpg}-1 and 256 neurons for \gls{ddpg}-5, followed by \gls{elu}~\cite{elu} activation functions. The size of replay buffer consists of \num{1e6} entries; the discount factor is set to $\gamma=0.99$; the soft update for Polyak averaging is set to $\tau=\num{1e-3}$. Adam optimization~\cite{adam} is employed with different learning rates $\alpha_{a}=\num{1e-4}$ and $\alpha_{c}=\num{1e-3}$ for the actor and critic, respectively, following the settings in~\cite{ddpg}. We notice that setting $\alpha_{c}$ larger than $\alpha_{a}$ is crucial to have the training more robust. Furthermore, a noise perturbation drawn from Gaussian distribution with $\mu=0$ and $\sigma=0.1$ is introduced to the action value during training, to encourage exploration~\cite{td3}. We train both \gls{ddpg}-1 and \gls{ddpg}-5 for 100 episodes.

\noindent \textbf{DDPG.} The architectures of actor and critic are both a 3-layer perceptron. Each layer contains 64 neurons, followed by \gls{elu}~\cite{elu} activation functions. The size of replay buffer consists of \num{1e6} entries; the discount factor is set to $\gamma=0.99$; the soft update for Polyak averaging is set to $\tau=\num{1e-3}$. Adam optimization~\cite{adam} is employed with different learning rates $\alpha_{a}=\num{1e-4}$ and $\alpha_{c}=\num{1e-3}$ for the actor and critic, respectively, following the settings in~\cite{ddpg}.
% We notice that setting $\alpha_{c}$ larger than $\alpha_{a}$ is crucial to have the training more robust.
Furthermore, a noise perturbation drawn from Gaussian distribution with $\mu=0$, $\sigma=0.1$ is introduced to the action value during training, to encourage exploration~\cite{td3}. The model converges after 20 training episodes.

\noindent \textbf{DHPG.} The architectures
of the actor $\mu$, critic $Q$ and the hyper-parameters are similar to that of \gls{ddpg} with minor differences: each layer contains 256 neurons, and the last layer of the actor split into two parallel action heads.
In particular, the actor is composed of
($i$) a 3-layer perceptron as the joint state state encoder $\mu_e$,
followed in parallel by
($ii$) a 2-layer perceptron as continuous head
$\mu_C$; and
($iii$) a 2-layer perceptron acting as
discrete head $\mu_p$.
%Given an input state $s_v$, the placement decision is taken as $p'_v=\argmax \{\mu_p\left( \mu_e(s_v|\ \theta^\mu) |\ \theta^\mu\right)\}$, and the scaling decisions are obtained as $(C_{1,v}^+,\ldots,C_{P,v}^+)=\text{DOD} \left(\mu_C\left( \mu_e(s_v|\ \theta^\mu) |\ \theta^\mu\right)\right)$.
The model converges after 30 episodes of training.

\subsubsection{Evaluation scenarios and metrics}
\label{metr}
Our goal is to maximize the long
term reward as described in formula~\eqref{eq:mdp}
that aims at meeting the delay target without over-provisioning resources, by finding an adequate strategy for task placement and scaling over time. 

\noindent \textbf{Convergence of \gls{dhpg}.} We show the
evolution of the long-term \textit{average reward} as we
increase the training episodes of the \gls{dhpg} scaling agents.

\noindent \textbf{Comparison Analysis.}
We run the approaches
specified in Section~\ref{ScalingApproaches}
%We run all scaling agents
%(coupled with the greedy placement agent described in Section~\ref{sec:ddpg})
using the traces summarized in Table~\ref{table:traces}, and compare them on the basis of  the($i$) \textit{average reward}, ($ii$) \textit{number of active CPUs}, and
%(iii) \textit{delay violations ($>$50ms)}.
($iii$) \textit{experienced delay}.
Note that we consider a trace that is long enough to cover various traffic patterns, including peak/off-peak hours. To have statistical significance in our results, we use 40 different traces (see Section~\ref{subsec:dataset}) for each experiment that span through the training or testing period.
% Before the actual comparisons of the approaches, we look into learning curves of \gls{ddpg}-based approaches. 

\noindent \textbf{\gls{dhpg} vs \gls{ddpg} per PoP.}
We closely examine the behavior of the \gls{dhpg} and \gls{ddpg} algorithms at each \gls{pop} to analyze how the agents
distribute the \emph{workload}.
%{\color{red} manage \textit{workload} distribution.}
%In the following, we look more closely into the behaviour of the \gls{dhpg}
%and \gls{ddpg}} approaches at each \gls{pop} to understand how the \rev{\gls{drl}} agents distribute the load.

\noindent \textbf{Optimality Gaps.} We show the optimality gap with regards to the global optimal solution for Problem~\ref{problem}.

\noindent \textbf{Computational Complexity.} Finally, we investigate the computational complexity of the proposed approaches and compare them against the state-of-the-art solutions.

  % Note that we consider a trace that is long enough to cover various traffic patterns, including peak/off-peak hours. To have statistical significance in our results, we use 40 different traces (see Section~\ref{subsec:dataset}) for each experiment that span through the training or testing period. Particularly, we look into learning curves of \gls{ddpg}-based agents and their behaviors at each \gls{pop}. Furthermore, we show the optimality gap with regards to the global optimal solution. Finally, we give the complexity analysis for all considered approaches in this paper.
% Convergence of DDPG-N
% Comparison analysis
% Behaviour of DDPG-N at each PoP
% Startup Optimality gap
% Complexity Analysis

% \subsection{Learning curves of \gls{ddpg}-N}
\subsection{Results}
\subsubsection{Convergence of \gls{dhpg}}
\label{subsec:training}

\begin{figure}[t]
    \input{Figures/learning-dhpg.tex}
    \caption{\gls{dhpg} learning curves over the training and testing sets.}
    \label{fig:learning}
\end{figure}

Fig.~\ref{fig:learning} shows the learning
curve
%of \gls{ddpg}-1 and \gls{ddpg}-5
of \gls{dhpg}
over 30 episodes. We report average rewards and the corresponding $95\%$ confidence intervals by repeating the experiment 10 times.

We can see that
\gls{dhpg} converges in around 5 episodes, and
the subsequent episodes lead to rather small
increase in the achieved reward.
Fig.~\ref{fig:learning} also shows that
the training reward is smaller than the
testing reward, as \gls{dhpg} placement
decisions are taken by the
discrete action head $\mu_p$, via sampling actions based on the a probability distribution.
That is, during the
training stage \gls{dhpg} forwards the
vehicle task to a stochastic \gls{pop} obtained
through $p'_v=X(\mu_p(z_v|\ \theta^\mu))$
-- see line~\ref{line:randomPoP}
from Algorithm~\ref{alg:dhpg}.
Whereas during the testing stage,
the \gls{dhpg} agent selects the
\gls{pop} with the maximum likelihood.

%% \gls{ddpg}-1 converges faster than \gls{ddpg}-5, while having lower variance. We argue that this difference is mainly caused by their reward signals. Due to the nature of their design, \gls{ddpg}-1 receives a reward by taking an action, while \gls{ddpg}-5 receives a reward by taking a set of actions (for multiple \glspl{pop}) jointly. The reward for the latter is identical to the average of rewards received from all \glspl{pop}, thus the value is  averaged out. % (some rewards are large yet others are small). 
%% This fact results in smaller gradients in optimization. Therefore, learning takes longer to converge.

In Fig.~\ref{fig:learning} we also observe how the confidence interval of the reward grows at episode 20. In the testing traces we observe that in one out of ten runs, the 
\gls{dhpg} exhibits a sudden reward
drop/increase. Such effect may be caused by ($i$) outliers in task departure times $T_v$, which are randomly generated at each run based on the process specified in
Section~\ref{subsec:environment-logic};
or ($ii$) not having a sufficiently small learning
rate $\alpha_a$. Nevertheless, the reward at the training and testing traces remain stable in the long run.

% Moreover, no overfitting is observed during 10 epochs. We note that the final reward obtained from the testing set is higher than that of the training set due to the Gaussian noise imposed upon the training phase.

% {\color{red}@Cyril:
% Specify the hours+date of the training
% and testing trace.
% Mention that the
% actor critic follow asymmetric learning
% (specify lr values),
% briefly explain how it work, and how
% that affects the learning in
% Fig.~\ref{fig:learning}. Also justify
% that 10 epochs was enough to reach convergence.
% On top, talk about how long is the training
% and testing trace, and that both span
% through off-peak/peak hours. Provide
% parameters for the discounted reward,
% $\tau$, and elaborate a bit why those
% values (e.g., a reference) and what they
% mean, and if possible try to draw some
% comment on how that impacts the learning
% curves. Finally, try to say that the training
% approach aimed at generalizing the \gls{ddpg}
% proper behaviour at both peak/off-peak
% hours.}

\subsubsection{Comparison Analysis}
\label{subsec:comparison}
% What we do in this section
We compare the
%\gls{ddpg}-1 and \gls{ddpg}-5 agents against the solutions in Section~\ref{sec:other-solutions}, since the placement approach is the same.
\gls{dhpg} agent with the solutions in
Section~\ref{ScalingApproaches}.
The comparison is done over the 40 randomly generated periods of 5.5h in the testing trace -- Formula~\eqref{eq:car-arrivals} and Table~\ref{table:traces}, respectively.
% % Introduce error bars, and abbreviations
% \textcolor{blue}{
% \st{Fig.~\mbox{\ref{fig:avg-reward}} illustrates the reward that each solution attained averaged over the 40 versions of the testing trace, while the error bars show three times the associated standard deviation of the reward. We refer as ``CNST'' to the constant provisioning explained in Section\mbox{\ref{subsec:cnst}}, and we denote as ``PI'' the PI provisioning specified in Section~\mbox{\ref{subsec:pi}}. Similarly, ``TES'' refers to the existing~\mbox{\gls{tes}} forecast-based provisioning~\mbox{\cite{v2n-access}} that we adapted as
% described in Section~\mbox{\ref{subsec:tes}}, using a granularity of \mbox{$W\cdot m=1$~sec}. That is, \mbox{\gls{tes}} updates its forecast about the road traffic every second.}}
\begin{figure}[t]
    \begin{tikzpicture}[gnuplot]
%% generated with GNUPLOT 5.2p8 (Lua 5.3; terminal rev. Nov 2018, script rev. 108)
%% mié 17 ene 2024 17:13:40
\path (0.000,0.000) rectangle (9.375,3.500);
\gpcolor{color=gp lt color border}
\gpsetlinetype{gp lt border}
\gpsetdashtype{gp dt solid}
\gpsetlinewidth{1.00}
\draw[gp path] (1.320,0.952)--(1.500,0.952);
\node[gp node right] at (1.136,0.952) {$0.5$};
\draw[gp path] (1.320,1.512)--(1.500,1.512);
\node[gp node right] at (1.136,1.512) {$0.6$};
\draw[gp path] (1.320,2.071)--(1.500,2.071);
\node[gp node right] at (1.136,2.071) {$0.7$};
\draw[gp path] (1.320,2.631)--(1.500,2.631);
\node[gp node right] at (1.136,2.631) {$0.8$};
\draw[gp path] (1.320,3.191)--(1.500,3.191);
\node[gp node right] at (1.136,3.191) {$0.9$};
\node[gp node center] at (2.070,0.308) {DHPG};
\node[gp node center] at (3.571,0.308) {DDPG~\cite{noms}};
\node[gp node center] at (5.071,0.308) {TES~\cite{v2n-access}};
\node[gp node center] at (6.571,0.308) {PI~\cite{5growth-scaling}};
\node[gp node center] at (8.072,0.308) {CNST};
\draw[gp path] (1.320,3.191)--(1.320,0.616)--(8.822,0.616);
\node[gp node center,rotate=-270] at (0.292,1.903) {average reward};
\gpfill{color=DodgerBlue3} (1.620,0.616)--(2.521,0.616)--(2.521,2.690)--(1.620,2.690)--cycle;
\gpcolor{rgb color={0.000,0.000,0.000}}
\draw[gp path] (1.620,0.616)--(1.620,2.689)--(2.520,2.689)--(2.520,0.616)--cycle;
\gpfill{color=Firebrick3} (3.120,0.616)--(4.022,0.616)--(4.022,2.037)--(3.120,2.037)--cycle;
\draw[gp path] (3.120,0.616)--(3.120,2.036)--(4.021,2.036)--(4.021,0.616)--cycle;
\gpfill{color=Gold3} (4.621,0.616)--(5.522,0.616)--(5.522,1.878)--(4.621,1.878)--cycle;
\draw[gp path] (4.621,0.616)--(4.621,1.877)--(5.521,1.877)--(5.521,0.616)--cycle;
\gpfill{color=Green4} (6.121,0.616)--(7.023,0.616)--(7.023,1.301)--(6.121,1.301)--cycle;
\draw[gp path] (6.121,0.616)--(6.121,1.300)--(7.022,1.300)--(7.022,0.616)--cycle;
\gpfill{color=DeepPink3} (7.622,0.616)--(8.523,0.616)--(8.523,1.140)--(7.622,1.140)--cycle;
\draw[gp path] (7.622,0.616)--(7.622,1.139)--(8.522,1.139)--(8.522,0.616)--cycle;
\draw[gp path] (2.070,2.684)--(2.070,2.695);
\draw[gp path] (1.980,2.684)--(2.160,2.684);
\draw[gp path] (1.980,2.695)--(2.160,2.695);
\draw[gp path] (3.571,2.023)--(3.571,2.049);
\draw[gp path] (3.481,2.023)--(3.661,2.023);
\draw[gp path] (3.481,2.049)--(3.661,2.049);
\draw[gp path] (5.071,1.869)--(5.071,1.885);
\draw[gp path] (4.981,1.869)--(5.161,1.869);
\draw[gp path] (4.981,1.885)--(5.161,1.885);
\draw[gp path] (6.571,1.274)--(6.571,1.327);
\draw[gp path] (6.481,1.274)--(6.661,1.274);
\draw[gp path] (6.481,1.327)--(6.661,1.327);
\draw[gp path] (8.072,1.088)--(8.072,1.189);
\draw[gp path] (7.982,1.088)--(8.162,1.088);
\draw[gp path] (7.982,1.189)--(8.162,1.189);
\gpcolor{color=gp lt color border}
\node[gp node center] at (2.070,2.997) {{0.81}};
\node[gp node center] at (3.571,2.344) {{0.69}};
\node[gp node center] at (5.071,2.185) {{0.67}};
\node[gp node center] at (6.571,1.608) {{0.56}};
\node[gp node center] at (8.072,1.447) {{0.53}};
\draw[gp path] (1.320,3.191)--(1.320,0.616)--(8.822,0.616);
%% coordinates of the plot area
\gpdefrectangularnode{gp plot 1}{\pgfpoint{1.320cm}{0.616cm}}{\pgfpoint{8.822cm}{3.191cm}}
\end{tikzpicture}
%% gnuplot variables
    \vspace{-1em}
    \caption{
        Average reward over 40 traces (generated with different random seeds) of a 5.5h interval. Number annotations represent the average value, and error bars represent three times the standard deviations.}
    \label{fig:avg-reward}
    \vspace{-3mm}
\end{figure}
% Comment the error bars results
Fig.~\ref{fig:avg-reward} illustrates the reward for each approach averaged over the 40 versions of the testing trace, while the error bars show three times the associated standard deviation of the reward.

%In Fig.~\ref{fig:avg-reward}
%we see that DDPG-1 and DDPG-5 increase  the average reward by $3$\% and $1.5$\% compared to the one attained by TES, respectively;
We can see that \gls{dhpg} and \gls{ddpg}
obtain higher average rewards by
$13$\% and $2$\% compared to the one attained by \gls{tes}, respectively;
while PI and CNST remain below \gls{tes}
with a decrease in average reward
% of $16.4$\% and $20.9$\%, respectively.
of $11$\% and $14$\%, respectively.
Moreover, the behaviour of almost every solution is fairly stable (see the error bars in Fig.~\ref{fig:avg-reward}) thanks to the training approach used; the parameters of every solution are selected to maximize the reward for a training trace with both off-peak and rush hours during the day.
Hence, every solution generalizes adequately despite the traffic variations of the 40 different versions of the testing trace.

% Check snapshot to gain intuition
To understand the rewards reported in Fig.~\ref{fig:avg-reward} we take a look at the behaviour of every solution during rush hours in the city of Turin. In Fig.~\ref{fig:snapshot}, we omit the CNST case, which we used as a benchmark in Fig.~\ref{fig:avg-reward}, to render the figure more readable. To further increase its readability, we smooth the traces by using Bezier curves to capture the main trends without being distracted by the small fluctuations inherent to the traces. % due to the high granularity of our environment (which is much below 1s).

% Why PI is so bad
We first focus on the PI approach. Primarily, PI obtains a small reward due to its failure to distribute the \gls{cv2n} traffic among~\glspl{pop}, since there is almost always a \gls{pop} with 5 active \glspl{cpu} ($\max_p\{C_p\}=5$ during long periods in Fig.~\ref{fig:snapshot}). A second reason is that PI has the highest number of \glspl{cpu} ($\sum_p C_p$) across \glspl{pop}. %$\sum_p C_p$ is the highest -- see Fig.~\ref{fig:snapshot}. 
And the third reason is that PI leads to the largest increase of \glspl{cpu} upon traffic peaks. Fig.~\ref{fig:snapshot} shows that as road traffic peaks around 7:30, it leads to the highest number of active \glspl{cpu} ($\sum_p C_p>20$). 
PI is over-provisioning computing resources, as a conservative solution
%{\color{red}
that aims at keeping load around 0.7 -- the best found target load for the $d_{tgt}$ parameter, as described in Section~\ref{subsec:parameter}.
%}
%This turns out to be a low value, leading to a delay lower than the target one. %Hence PI fails to obtain the best possible reward (Fig.~\ref{fig:avg-reward}).

\begin{figure}[t]
    \input{Figures/snapshot.tex}
    \caption{Behaviour of DHPG and other solutions as the number of vehicles change over time (bottom). From top to down we show the maximum/minimum CPUs each \gls{pop} has ($\max_p\{C_p\}$ and $\min_p\{C_p\}$, respectively), and the sum of \gls{pop} CPUs
    ($\sum_p^5 C_p$).}
    \label{fig:snapshot}
    \vspace{-3mm}
\end{figure}

% What can we tell of TES & DDPGs from
% the snapshot?
Although Fig.~\ref{fig:snapshot} conveys the flaws of PI, it does not readily reveal why
% DDPG-1 and DDPG-5
\gls{dhpg} and \gls{ddpg}
outperform \gls{tes}. Actually, the three solutions use almost the same number of \glspl{cpu} across \glspl{pop} $\sum_p C_P$. The main difference
% , although not that pronounced,
is that
% DDPG-5
\gls{dhpg}
is prone to consolidating workloads at certain \glspl{pop}. We observe that the number of \glspl{cpu} for the most loaded \gls{pop} ($\max_p \{C_p\}$) remains higher than for
% DDPG-1
\gls{ddpg}
and \gls{tes} (see Fig.~\ref{fig:snapshot} top). At the same time, %Fig.~\ref{fig:snapshot} also shows that 
% DDPG-5
\gls{dhpg}
tends to have less \glspl{cpu} in the least loaded \gls{pop} ($\min_p \{C_p\}$) compared to
% DDPG-1
\gls{ddpg}
and \gls{tes}. %Specifically, it provisions the least loaded \gls{pop} with less \glspl{cpu} than DDPG-1 and TES.
% Despite both facts provide insights of \gls{dhpg}, \gls{ddpg} and \gls{tes} behaviour, the reason behind \gls{dhpg}'s superior performance remains unclear.
% Both facts give insights on why
% DDPG-5 is slightly worse than DDPG-1, however it is difficult to draw conclusions regarding the reward. % as it compensates having a loaded \gls{pop} by reducing further the load on  the least loaded one.

% To really understand what happens,
% we check the PDFs and CDFs

%%\begin{figure}[t]
%%   \input{Figures/all-delays-dens.tex}
%%   \caption{
%%   \gls{epdf} of the experienced
%%   total delay.}
%%   \label{fig:all-delays-dens}
%%\end{figure}
\begin{figure}[t]
   \input{Figures/all-delays99-cdf.tex}
   \caption{
   \acrshort{ecdf} for the upper bound of the
   99 delay percentile.}
   \label{fig:all-delays-dens}
   \vspace{-3mm}
\end{figure}

\begin{figure}[t]
   \begin{tikzpicture}[gnuplot]
%% generated with GNUPLOT 5.2p8 (Lua 5.3; terminal rev. Nov 2018, script rev. 108)
%% mié 17 ene 2024 17:27:24
\path (0.000,0.000) rectangle (8.750,6.125);
\gpcolor{color=gp lt color border}
\gpsetlinetype{gp lt border}
\gpsetdashtype{gp dt solid}
\gpsetlinewidth{1.00}
\draw[gp path] (1.320,2.192)--(1.500,2.192);
\node[gp node right] at (1.136,2.192) {$25$};
\draw[gp path] (1.320,3.258)--(1.500,3.258);
\node[gp node right] at (1.136,3.258) {$50$};
\draw[gp path] (1.320,4.324)--(1.500,4.324);
\node[gp node right] at (1.136,4.324) {$75$};
\draw[gp path] (1.320,5.390)--(1.500,5.390);
\node[gp node right] at (1.136,5.390) {$100$};
\draw[gp path] (1.320,1.165)--(1.320,0.985);
\node[gp node center] at (1.320,0.677) {$10$};
\draw[gp path] (2.695,1.165)--(2.695,0.985);
\node[gp node center] at (2.695,0.677) {$12$};
\draw[gp path] (4.071,1.165)--(4.071,0.985);
\node[gp node center] at (4.071,0.677) {$14$};
\draw[gp path] (5.446,1.165)--(5.446,0.985);
\node[gp node center] at (5.446,0.677) {$16$};
\draw[gp path] (6.822,1.165)--(6.822,0.985);
\node[gp node center] at (6.822,0.677) {$18$};
\draw[gp path] (8.197,1.165)--(8.197,0.985);
\node[gp node center] at (8.197,0.677) {$20$};
\draw[gp path] (1.320,5.816)--(1.320,1.165)--(8.197,1.165)--(8.197,5.816)--cycle;
%\node[gp node left] at (1.492,3.044) {\shortstack{median\\$\leq14$ CPUs}};
\node[gp node center,rotate=-270] at (0.292,3.490) {eCDF [\%]};
\node[gp node center] at (4.758,0.215) {$\sum_p^5 C_p$ [CPUs]};
\node[gp node right] at (6.729,2.423) {DHPG};
\gpcolor{color=DodgerBlue3}
\gpsetlinewidth{7.00}
\draw[gp path] (6.913,2.423)--(7.829,2.423);
\draw[gp path] (1.320,1.296)--(2.008,1.548)--(2.695,2.109)--(3.383,2.776)--(4.071,3.400)%
  --(4.759,4.080)--(5.446,4.602)--(6.134,4.950)--(6.822,5.130)--(7.509,5.270)--(8.197,5.323);
\gpcolor{color=gp lt color border}
\node[gp node right] at (6.729,2.115) {DDPG~\cite{noms}};
\gpcolor{color=Firebrick3}
\gpsetlinewidth{5.00}
\draw[gp path] (6.913,2.115)--(7.829,2.115);
\draw[gp path] (1.320,1.250)--(2.008,1.423)--(2.695,1.786)--(3.383,2.556)--(4.071,3.294)%
  --(4.759,4.264)--(5.446,4.777)--(6.134,5.114)--(6.822,5.288)--(7.509,5.345)--(8.197,5.377);
\gpcolor{color=gp lt color border}
\node[gp node right] at (6.729,1.807) {PI~\cite{5growth-scaling}};
\gpcolor{color=Green4}
\gpsetdashtype{gp dt 1}
\draw[gp path] (6.913,1.807)--(7.829,1.807);
\draw[gp path] (1.320,1.165)--(2.008,1.228)--(2.695,1.415)--(3.383,1.805)--(4.071,2.370)%
  --(4.759,3.124)--(5.446,3.942)--(6.134,4.524)--(6.822,4.832)--(7.509,5.114)--(8.197,5.233);
\gpcolor{color=gp lt color border}
\node[gp node right] at (6.729,1.499) {TES~\cite{v2n-access}};
\gpcolor{color=Gold3}
\gpsetdashtype{gp dt solid}
\draw[gp path] (6.913,1.499)--(7.829,1.499);
\draw[gp path] (1.320,1.208)--(2.008,1.407)--(2.695,1.759)--(3.383,2.392)--(4.071,3.060)%
  --(4.759,3.898)--(5.446,4.546)--(6.134,4.961)--(6.822,5.181)--(7.509,5.279)--(8.197,5.344);
\gpcolor{color=gp lt color border}
\gpsetlinewidth{1.00}
\draw[gp path] (1.320,5.816)--(1.320,1.165)--(8.197,1.165)--(8.197,5.816)--cycle;
\gpcolor{rgb color={0.000,0.000,0.000}}
\gpcolor{color=gp lt color border}
\gpsetdashtype{gp dt solid}
% \draw[gp path,->](3.039,3.258)--(3.968,3.258);
%% coordinates of the plot area
\gpdefrectangularnode{gp plot 1}{\pgfpoint{1.320cm}{1.165cm}}{\pgfpoint{8.197cm}{5.816cm}}
\end{tikzpicture}
%% gnuplot variables
   \caption{
   \acrshort{ecdf} of CPUs used by all 
   \gls{pop}s $\sum_p C_p$.}
   \label{fig:all-cpus-dens}
   \vspace{-3mm}
\end{figure}

% What ePDF+eCDF tells about DDPG-5
To gain a deeper understanding on the behaviour of DHPG and other benchmarks, we plot 
%{\color{red}
($i$) the \gls{ecdf} of the delay $\overline{d_v^{99}}$
experienced by vehicles  in Fig.~\ref{fig:all-delays-dens}
and ($ii$) the \gls{ecdf} of the total number of active \glspl{cpu} across the five \glspl{pop} in Fig.~\ref{fig:all-cpus-dens}.
% For the former we resort to kernel density estimates to derive the \gls{epdf} illustration.
For the former we compute the average delay experienced over
time in the experiments, and then compute the 99-percentile
bound from formula~\eqref{eq:bound}.
%}

%Fig.~\ref{fig:all-delays-dens} shows that the \gls{epdf} peak of DDPG-5 is the closest to the target delay of $d_{tgt}=50$~ms. Hence, vehicles are most likely to experience delays near the target 50~ms, which translates into a higher reward (see~Fig.~\ref{fig:reward}). But at the same time,

Fig.~\ref{fig:all-delays-dens} shows that
with \gls{dhpg} the $99^{th}$ percentile of the delay remains close to
the target of 100~ms, violating the
delay constraint  at approximately 10\% of the times
(similar as \gls{tes} and \gls{ddpg}), while \gls{ddpg}
and \gls{tes} result in smaller task delays.
Inline with the previous figure, Fig.~\ref{fig:all-cpus-dens} shows that \gls{dhpg} uses less CPUs to ensure
the target delay. In particular,
\gls{dhpg} has higher chances than
\gls{tes} and \gls{ddpg} to use less than 14~CPUs, thus  acieving a higher reward in Fig.~\ref{fig:avg-reward}.

Overall, Fig.~\ref{fig:all-delays-dens}
and Fig.~\ref{fig:all-cpus-dens} show
a rather similar behaviour for \gls{tes} and
\gls{ddpg}.% with the latter using less resources.
 The proposed \gls{dhpg} ends up outperforming all the
solutions because it is more likely
to use less CPUs while ensuring
that $99^{th}$ percentile of the delay remains below the 100~ms target. Both facts suggest that \gls{dhpg} actor $\mu$
has inherently learned the traffic trends
at all \glspl{pop}, hence, placing new
task arrivals at non-saturated \glspl{pop} to avoid
turning on additional \gls{cpu}s.

%In the next section, we further investigate how \gls{dhpg} differs from \gls{ddpg} regarding the placement
%of tasks and scaling
%decisions among \glspl{pop}.

%% %\bigskip
%% % Concluding remark on DDPGs and TES
%% Overall, Fig.~\ref{fig:all-delays-dens} and Fig.~\ref{fig:all-cpus-dens} show that the DDPG-1 actor $\theta^\pi$ learned a policy with a very similar behaviour to the forecast-based solution of TES. This suggests that DDPG-1 has (implicitly) learned to foresee how the traffic patterns will evolve in \gls{cv2n} communications (without having to provide it with the seasonality as in the case of TES). However, the DDPG-5 scaling agent learned a more conservative policy that uses less \glspl{cpu} to support the \gls{cv2n} workload.
%% % As a result, most vehicles experience delays $d_v$ near the target of $d_{tgt}=50~ms$ and achieve high rewards, but also have higher risk of experiencing delays larger than that (thus decreasing the reward accordingly).
%% {\color{red}
%% As a result, DDPG-5 is less likely to meet
%% the 99-percentile delay requirement of 100\,ms.}
%% Below we check in more  detail DDPG-1 and DDPG-5 to gain a better understanding on the differences between the top two solutions
%% (Fig.~\ref{fig:avg-reward}).

\subsubsection{\gls{dhpg} vs \gls{ddpg}
per \gls{pop}}
\label{subsec:ddpg-behaviour}

% Recap
%So far we know that \gls{dhpg} and \gls{ddpg} differ by a $13\%$ in the obtained average reward (Fig.~\ref{fig:avg-reward}).

%%% repetition!!!
%As shown in Fig.~\ref{fig:avg-reward}, \gls{dhpg} achieves a $13\%$ higher average reward compared to \gls{ddpg}.
%We also know that \gls{dhpg} is more conservative than \gls{ddpg} in terms of allocating \glspl{cpu}  (Fig.~\ref{fig:all-cpus-dens}) and tasks experience larger delays. Nevertheless, \gls{dhpg} is slightly less prone than \gls{ddpg} to violate the 99 delay percentile requirement -- see~Fig.~\ref{fig:all-delays-dens}.

In Fig.~\ref{fig:ddgp1-vs-ddpg5} we compare the behaviour of \gls{dhpg} and \gls{ddpg} per~\gls{pop} on
% (i) the \gls{epdf} of the total delay experienced by vehicles $d_v$,
($i$) the \gls{epdf}  of the the $99^{th}$ percentile latency;
($ii$)
the \gls{epmf} of
the number of \glspl{cpu} $C_p$; and ($iii$)  \gls{epmf} of the number of assigned vehicles $N_p$. As before, we use kernel density estimates to derive the \gls{epdf} and \gls{epmf}s.

%% % Recap
%% So far we know that DDPG-1 and DDPG-5 differ by just $0.01$ in the obtained average reward (Fig.~\ref{fig:avg-reward}). We also know that DDPG-5 is more conservative in terms of allocating \glspl{cpu} than DDPG-1 (Fig.~\ref{fig:all-cpus-dens}), hence risking more delay violations, yet remaining near the target delay (see Fig.~\ref{fig:all-delays-dens}).
%% % What we illustrate per PoP
%% In Fig.~\ref{fig:ddgp1-vs-ddpg5} we compare the behaviour of DDPG-1 and DDPG-5 per~\gls{pop} by looking at
%% % (i) the \gls{epdf} of the total delay experienced by vehicles $d_v$,
%% {\color{red}
%% (i) the \gls{epdf} of the bound for the 99-percentile delay,
%% }
%% (ii) the number of \glspl{cpu} $C_p$, and (iii)  the number of assigned vehicles $N_p$. As before, we resort to kernel density estimates to derive the \gls{epdf}s.

\begin{figure}
    \centering
    %\subfloat{\input{Figures/delays-dens.tex}}\\
    \subfloat{\input{Figures/delays99-dens.tex}}\\
    \subfloat{\input{Figures/cpus-dens.tex}}\\
    \subfloat{\input{Figures/assigned-dens.tex}}
    \caption{
    Comparison of DHPG vs. DDPG over all PoPs.
    We compare the \gls{epdf} of the
99 delay percentile bound $\overline{d_v^{99}}$ (top), \gls{epmf} of the number of CPUs $C_p$ at each \gls{pop} (middle), and number of vehicles assigned to each \gls{pop} $N_p$ (bottom). }
    \label{fig:ddgp1-vs-ddpg5}
    \vspace{-3mm}
\end{figure}

% DDPG-1 distributes more assigned vehicles
 Fig.~\ref{fig:ddgp1-vs-ddpg5} shows that \gls{ddpg} distributes the number of vehicles $N_p$ per \gls{pop} more evenly than
\gls{dhpg}, as the highest \gls{epmf} peak is near $N_p=8$ for every \gls{pop}. In comparison, \gls{dhpg} assigns less vehicles to
\glspl{pop} A and E, and more vehicles to
\glspl{pop} B and C
-- see
Fig.~\ref{fig:ddgp1-vs-ddpg5}~(middle).
% -- see how the formers' \gls{epdf} are skewed to the left, and the latters' to the right.
% More assigned vehicles result in unbalanced CPUs
%The second evidence is a consequence of the first.
% As a result \gls{dhpg} uses (i) a smaller number of \glspl{cpu} in \glspl{pop} A and E - as evident by the {\color{red} \gls{epdf} in Fig.~\ref{fig:ddgp1-vs-ddpg5}~(middle) that skews to the left - and a larger set of resources in \glspl{pop} B and C  (approximately $C_p=5$). }
% In other words, as \gls{dhpg} assigns less vehicles to \glspl{pop} A and E, it also uses less \glspl{cpu}.
%Conversely, as \gls{dhpg} assigns the excess of vehicles to \glspl{pop} B and C, it also provisions more \glspl{cpu}, around $C_p>4$ to be precise. 
%This explains why on Fig.~\ref{fig:snapshot}~(top) the maximum loaded~\gls{pop} 
%(either \gls{pop}~B or \gls{pop}~C)
%has more active \glspl{cpu}
Furthermore, we can see in Fig.~\ref{fig:ddgp1-vs-ddpg5}~(top) that  
%The third evidence we
%draw from Fig.~\ref{fig:ddgp1-vs-ddpg5}~(top)
 C-V2N tasks experience a larger delay
 at \glspl{pop} where \gls{dhpg}
dispatches fewer tasks -- i.e. \glspl{pop}
A and E. This is due to the significantly smaller number of CPUs used in \glspl{pop} A and E
(Fig.~\ref{fig:ddgp1-vs-ddpg5}~(middle)).
With the use of \gls{ddpg}, tasks experience a rather
equally distributed latency
(Fig.~\ref{fig:ddgp1-vs-ddpg5}~(top))
as the approach distributes equally the load across \glspl{pop}.
Finally, in terms of violating the 100~ms delay  bound (Fig.~\ref{fig:ddgp1-vs-ddpg5}~(top)), the two approaches behave quite similar, as already noted for Fig.~\ref{fig:all-delays-dens}.

%Fig.~\ref{fig:all-delays-dens}, the results
%in Fig.~\ref{fig:ddgp1-vs-ddpg5}~(top) evidence
%that \gls{dhpg} and \gls{ddpg} are very
%close in terms of violating the 99 delay
%percentile requirement of 100~ms.

%%% % How it is the latency
%%% The third evidence we draw from Fig.~\ref{fig:ddgp1-vs-ddpg5}~(top) is that DDPG-5 is more prone to exceed the target delay of $d_{tgt}=50$~ms, as we discussed in Fig.~\ref{fig:all-delays-dens}.
%%% In particular, although DDPG-5 assigns few vehicles to \gls{pop}~B the \gls{cpu} provisioning should not be as low, for that results in a peak 
%%% {\color{red}to the right of the target delay of 100\,ms.}
%%% % on the right side of the target delay $d_{tgt}$~\gls{epdf}.
%%% That is, DDPG-5 assigned too little \glspl{cpu} to \gls{pop}~B and this results in high probability of violating the target delay of
%%% 50~ms,
%%% {\color{red}100~ms,}
%%% as shown in the peak of Fig.~\ref{fig:ddgp1-vs-ddpg5}~(top).
%%% Conversely, DDPG-1 keeps similar \gls{epdf}
%%% % for the vehicle total delay $d_v$ across \glspl{pop}, always having a higher mass of probability below 50~ms.
%%% {\color{red}for the vehicle 99 delay percentile across \glspl{pop}, always having a higher mass of probability below 100~ms.}

% Draw conclusions of DDPG-1 and DDPG-5

Overall, Fig.~\ref{fig:ddgp1-vs-ddpg5}
highlights
that decentralized \gls{ddpg} leads
to a more even load distribution across
\glspl{pop}.
While \gls{dhpg} manages to
significantly ease the load
at \glspl{pop} A and E, at the expense of
loading more \glspl{pop} B and C.
Still, the load ease at \glspl{pop} A and
E is so significant 
-- see~Fig.~\ref{fig:ddgp1-vs-ddpg5} --
that the total number of used \glspl{cpu}
turns out to be smaller with \gls{dhpg}
than with \gls{ddpg}.
Thus, \gls{dhpg} outperforms
\gls{ddpg} regarding average reward
in Fig.~\ref{fig:avg-reward}.

\subsubsection{Optimality Gap}
\label{subsec:optimality-gap}

% OPT gap, we need oracle, still NP-hard, reduce
We study how far \gls{dhpg} is with respect to the performance of optimal solution for Problem~\ref{problem}.
To find an optimal solution we resort to an
oracle that knows \emph{a posteriori} the
vehicle arrival times $\{t_1,t_2,\ldots\}$.
However, an oracle cannot solve
Problem~\ref{problem} in polynomial time
-- see Lemma~\ref{lemma:complex}.
Hence, it is a must to reduce the problem
size to find a feasible solution.
% }
% Problem size
% \rev{
% The problem
% size is governed by the number of vehicle
% arrivals $|V|$ considered, for the arrival
% of each vehicle $v\in V$ is associated
% with the corresponding placement
% $p_v'$ and scaling $C_{p,v}$ decisions.
% In particular, there are
% $\left(P (C_p^{\max})^P \right)^{|V|}$
% possible decisions for a problem with
% $|V|$ vehicle arrivals, i.e. a vast
% search space to find an optimal solution.
% For example, an
% oracle solution using AMPL~\cite{ampl}
% and the KNITRO~\cite{knitro} solver
% could not find a solution in 65\,\textrm{h}
% for $|V|=81$ vehicle arrivals.
% }
The problem
size is governed by the number of vehicle
arrivals $|V|$ considered, for the arrival
of each vehicle $v\in V$ is associated
with the corresponding placement
$p_v'$ and scaling $C_{p,v}$ decisions.
In particular, there are
$\left(P (C_p^{\max})^P \right)^{|V|}$
possible decisions for a problem with
$|V|$ vehicle arrivals, i.e. a vast
search space to find an optimal solution.
For example, an
oracle solution using AMPL~\cite{ampl}
and the KNITRO~\cite{knitro} solver
could not find a solution in 65\,\textrm{h}
for $|V|=81$ vehicle arrivals.

% Small scale scenario
In order to find an optimal solution through
an oracle we study a small scale instance of
Problem~\ref{problem}. The considered small
scale scenario concerns $|V|=6$ vehicle
arrivals $\{t_{v'},\ldots,t_{v'+6}\}$ and
a common starting system state
$s_{v'}$. The system state corresponds to the
one achieved by \gls{dhpg} by the time
$t_{v'}$ vehicle $v'$ arrived.

\begin{table}[t]
\caption{Small scale evaluation}
\label{table:og}
\centering
%\resizebox{\columnwidth}{!}{%
\begin{tabular}{ c l l l  }
 \toprule
 \textbf{Algorithm} & \textbf{avg gap} & \textbf{avg $\#$CPUs} & \textbf{avg load} \\
 \midrule
 Oracle & - & 3.20 & 86.6\% \\
 \gls{dhpg}   & 5.2\% & 3.23 & 90\%  \\
 \gls{pi}       & 10.4\% & 3.04 & 89\%  \\
 \gls{tes}   & 11.3\% & 3.04 & 89\%  \\
 \gls{ddpg}   & 12.1\% & 3.02 & 89.2\%  \\
 CNST     & 77.7\% & 3.40 & 77.4\%  \\
%  DZ       & 0.77 & 0.47 & 0\%  \\
%  TES-1s   & 0.63 & 0.5 & 10\%  \\
%  TES-5m   & 0.33 & 0 & 67\%  \\
%  GA       & 0.26 & 2.73 & 0\%  \\
 \bottomrule
\end{tabular}
%}
\vspace{-3mm}
\end{table}

% Discuss results
Table~\ref{table:og} shows the optimality
gap achieved by \gls{dhpg} and the
benchmarks in the aforementioned small scale
scenario. The optimality gap is measured
as the relative difference between
the oracle reward $R^*$ and the
reward achieved by the algorithm $R$, i.e.
we compute it as
$\tfrac{R^*-R}{R^*}$.
Clearly, \gls{dhpg} is the one
staying the closest to the optimal solution,
with just a 5.2\% optimality gap. Whereas,
\gls{ddpg} and other benchmarks have
optimality gaps above a 10\%.
Although \gls{tes} and \gls{pi} outperform
\gls{ddpg} in the considered small scale
scenario, the latter achieves larger
long-term reward -- see Fig.~\ref{fig:avg-reward}.

\begin{figure}
    \input{Figures/processing-delay.tex}
    \caption{Experienced
    average processing delay $\mathbb{E}[d_{p,v}]$ at a \gls{pop} as the number of assigned vehicles $N_p$ and \glspl{cpu} (lines) increase.}
    \label{fig:processing-delay}
\end{figure}

% Explain why the avg#CPUs=3
Table~\ref{table:og} also details that
all solutions use around three CPUs.
Such result is because
with three CPUs it is possible to achieve task
processing delays close to the target
delay $d_{tgt}$ (see~Fig.~\ref{fig:processing-delay}),
thus leading to higher reward.
It is worth to mention that \gls{dhpg}
fosters task delays \emph{close} to the target $\min d_v-d_{tgt}$, whereas
the benchmarks foster task delays \emph{below}
the target latency $d_v<d_{tgt}$
due to their greedy
placement~\eqref{eq:greedy-assign}
strategy.
As a result, \gls{dhpg} may lead to
under-provisioning of \gls{pop} yet with high rewards.
This aligns with the fact that \gls{dhpg} has
higher load and \glspl{cpu}
than the benchmarks in
Table~\ref{table:og}.

% Take-away
% \rev{
% Overall, we conclude that \gls{dhpg} halves
% the competitors optimality gap because it
% considers the placement dimension, rather
% than only the scaling. Consequently, it
% exploits the placement and scaling
% inter-dependency to maximize the reward.
% }

Overall, we conclude that by leveraging a data-driven approach, \gls{dhpg} learns a more flexible strategy in placement dimension, rather than using greedy method. The design of joint state encoder effectively exploits the inter-dependency between placement and scaling to maximize the reward.

\subsubsection{Computational Complexity}
\label{subsec:complexity}

\begin{table}[t]
    \caption{Computational complexity of algorithms.}
    \centering
    \begin{tabular}{ l l l l}
         \toprule
        \textbf{\gls{dhpg}} & \textbf{\gls{ddpg}} & \textbf{\gls{pi}} & \textbf{\gls{tes}} \\ \midrule
          $\mathcal{O}(\eta^2H)$ & $\mathcal{O}(\eta^2H)$ & $\mathcal{O}(P)$ & $\mathcal{O}(P)$\\
         \bottomrule
    \end{tabular}
    \label{tab:complexity}
\end{table}

% \begin{figure}[t]
%     \centering
%     \input{Figures/avg-inference-error-bars.tex}
%     \caption{Average run time of each algorithm
%     when performing the placement and scaling
%     in the considered 5~\gls{pop}s upon the
%     arrival of a vehicle.}
% \end{figure}

%% % Introduce section 
%% In this section  we derive the worst-case
%% runtime analysis of the proposed
%% \gls{dhpg} approach
%% and further compare their average runtime
%% to \gls{ddpg}, \gls{pi}, and \gls{tes}.
%% %Specifically, we derive the worst-case run time analysis of the proposed \gls{ddpg}, \gls{pi}, and \gls{tes}.

% Introduce section 
In this section,  we derive the worst-case
computational complexity analysis of the considered
approaches.
 
 \gls{pi}, \gls{tes} and \gls{ddpg} take the
placement decisions based
on~\eqref{eq:greedy-assign}. Hence, the
placement has computational complexity
$\mathcal{O}(P)$. Then, 
\gls{tes} performs $W$ $m$-minutes
predictions at
each \gls{pop} to decide upon scaling, having an overall computational complexity
$\mathcal{O}(WP)$.
\gls{pi} performs a fixed
set of operations at each \gls{pop}
to derive the scaling decision, thus having
a runtime complexity $\mathcal{O}(P)$.
% %\subsubsection{Proposed \gls{ddpg} algorithm}
%%% %\label{subsec:ddpg-complexity}
%%% %\noindent\textbf{DDPG-based approach.} 
%%% After completion of the training stage,
%%%  Algorithm~\ref{alg:pns} places vehicles' application tasks using
%%% $\argmin_{p\in P}(l_{p_v,p}+d_p)$. %, with
%%% %$d_p$ derived using the total average delay
%%% %from \eqref{eq:avg-delay}, and $l_{p_v,p}$
%%% %the measured average delay between
%%% %the \gls{pop} where the vehicle arrived
%%% %$p_v$ and the \gls{pop} where the demand is
%%% %placed $p$. 
%%% Overall, the placement stage
%%% has runtime complexity $\mathcal{O}(P)$.
\gls{ddpg} uses its trained \gls{ddpg}
actor $\mu$ to scale resources, as described in\cite{noms}.
Each fully connected layer of the
actor network 
performs a vector-matrix
multiplication $\mu_i\cdot h_{i-1}$
during the forward pass, where
$\mu_i\in \mathbb{R}^{\eta_i\times \eta_i}$
are the weights of the $i$\textsuperscript{th}
hidden layer with $h_i$ neurons,
$\eta_i$ being the number of neurons at the
$i$\textsuperscript{th} hidden layer, and
$h_{i-1}\in\mathbb{R}^{\eta_i}$ being the output of the previous layer. %the output
%vector of the $i-1$\textsuperscript{th}
%hidden layer. 
Let
$\eta = \max_i \{\eta_i\}$ denote the maximum number of neurons in any hidden layer and
$H$ for the total number of hidden layers.
Then a forward pass of the
\gls{ddpg} actor network
$\mu$  has a computational complexity of $\mathcal{O}(\eta^2 H)$. 
Since the forward pass dominates the placement operation, which has complexity $\mathcal{O}(P)$, this also determines the overall complexity of the \gls{ddpg} agent. % that has $\eta_i=64$ neurons per layer and $H=3$ hidden layers.
%consequently so is the run time complexity for the \gls{ddpg} agent -- for the forward pass dominates the  \mathcal{O}(P)$ placement operation.
%{\color{blue}It is worth mentioning
%the \gls{ddpg} actor has
%$\eta_i=64$ neurons per layer and $H=3$ hidden layers.}

%

%\subsubsection{\gls{dhpg} complexity Analysis}
%\label{subsec:dhpg-complexity}

%We now discuss the runtime complexity of the
%\gls{dhpg} agent at both the training and
%inference stages. 
The \gls{dhpg} framework constitutes of
a joint actor and a joint critic, as illustrated in Fig.~\ref{fig:dhpg}.
To obtain the computational complexity
of \gls{dhpg} at  \emph{inference stage},
we look into the operations performed
by the joint actor $\mu$.
The input state vector $s_v$
traverses 5 fully connected
hidden layers, each with $\eta_i$ neurons. 
Every layer has a weight matrix
$\mu_i\in\mathbb{R}^{\eta_i\times\eta_{i-1}}$
and performs a vector-matrix multiplication
$\mu_i\cdot h_{i-1}$, with
$h_i\in\mathbb{R}^{\eta_i}$ being the size
of a hidden layer. 
Hence, the computational complexity of a forward pass in the joint actor
corresponds to one matrix
multiplication % $\mu_i\cdot h_{i-1}$
per layer, i.e., $\mathcal{O}(\eta^2H)$ with
$\eta=\max_i \{\eta_i\}$.
%, $H=5$ layers, and $\eta_i=256$ neurons in each hidden layer. 
The last layer of the
joint actor %, i.e., the last layer of
%both the discrete and continuous action
%heads, 
has as many neurons
as \glspl{pop}.
As a result, an increasing number of \glspl{pop} $P$ will lead to a longer inference
time. %Specifically, the number
%of entries increases %in the weight matrix of 
% in the last layer of the action head
%-- with size
%$\mu_5\in\mathbb{%R}^{P\times\eta_4}$.
%Thus, the corresponding matrix multiplication
%$\mu_5\cdot h_4$ in the forward pass 
%increases the number of operations
%at pace $\mathcal{O}(P\eta_4)$.

To obtain the computational complexity
of \gls{dhpg} at the \emph{training stage},
we look into the operations performed by the
joint critic $Q$ and recall the learning
Algorithm~\ref{alg:dhpg}. 
The joint critic $Q$ has three fully
connected hidden layers with the last layer
 estimating the $Q$-value. % of the actor's decision.
The first hidden layer receives the
state-action tuple $(s_v,a_v)$ and has a
weight matrix
$\theta_1^Q\in\mathbb{R}^{ \eta_1\times(3P+2)}$,
with $P$ being the number of \glspl{pop}
and $\eta_1$ being the number of neurons of
the layer.
Hence, a forward pass of the first
layer has computational complexity
$\mathcal{O}(\eta_1P)$. %and increasing the number of \glspl{pop} increases the runtime of the first forward pass.
Performing
a forward pass across all five layers
involves the matrix multiplication
of the weights associated to each layer,
with the inner layers having 
weight matrix
$\theta_i^Q\in\mathbb{R}^{\eta_i\times \eta_{i-1}}$.
Therefore, %a forward pass of
the computational complexity of critic $Q$ is $\mathcal{O}(\eta^2 H)$ with $\eta=\max\{\eta_i,3P+2\}$. %=256$ and $H=4$.
Considering   Algorithm~\ref{alg:dhpg}, the dominant
operation at the training
stage is the update of the critic
$\delta\theta^Q$ and actor
$\delta\theta^\mu$ weights using the
batch of $b$ transitions of the replay
buffer.
To compute the critic
update $\delta\theta^Q$ we calculate the
mean square error between the critic $Q$
and target critic $Q'$ for each sample
within the batch, i.e., we perform
$2b$ forward passes for the critic
and target critic
-- that is
$\mathcal{O}(b\eta^2H)$ operations.
Then, we run a back propagation
to compute the associated gradient descend
$\nabla_{\theta^Q}$
in line~\ref{line:critic-update}. As the complexity of the back propagation is defined by a matrix multiplication
for each layer~\cite{backprop},
computing $\delta\theta^Q$ has a
total complexity of
$\mathcal{O}(b\eta^2H)$.
Similarly, the complexity of
the critic update $\delta\theta^\mu$ is
$\mathcal{O}(b\eta^2H)$.
As a result, the overall
complexity of \gls{dhpg} at its
training stage is
$\mathcal{O}(EVb\eta^2H)$ with
$E$ being the number of episodes and
$V$ being the number of vehicle arrivals
in the training dataset.

% With the aforementioned,
% we conclude the \gls{dhpg} runtime
% complexity is dominated by the number
% of hidden layers, their size, so as the
% training hyper-parameters:
% episodes, batch size and size of the
% training dataset. Increasing the number
% of \glspl{pop} leads to a linear increase
% of the runtime of the forward pass of the
% last and
% former layer of he joint actor and critic,
% respectively.
In summary, the computational complexity of the \gls{dhpg} is determined by the number and size of hidden layers, as well as training hyper-parameters, including episodes, batch size, and training dataset size. Increasing the number of \glspl{pop} will linearly increase the runtime of the forward pass for the last layer of the actor and the first layer of the critic, respectively.

\begin{figure}[t]
    \centering
    \begin{tikzpicture}[gnuplot]
%% generated with GNUPLOT 5.2p8 (Lua 5.3; terminal rev. Nov 2018, script rev. 108)
%% jue 11 ene 2024 12:23:11
\path (0.000,0.000) rectangle (9.375,3.500);
\gpcolor{color=gp lt color border}
\gpsetlinetype{gp lt border}
\gpsetdashtype{gp dt solid}
\gpsetlinewidth{1.00}
\draw[gp path] (1.504,0.616)--(1.684,0.616);
\node[gp node right] at (1.320,0.616) {$10$};
\draw[gp path] (1.504,1.004)--(1.594,1.004);
\draw[gp path] (1.504,1.230)--(1.594,1.230);
\draw[gp path] (1.504,1.391)--(1.594,1.391);
\draw[gp path] (1.504,1.516)--(1.594,1.516);
\draw[gp path] (1.504,1.618)--(1.594,1.618);
\draw[gp path] (1.504,1.704)--(1.594,1.704);
\draw[gp path] (1.504,1.779)--(1.594,1.779);
\draw[gp path] (1.504,1.845)--(1.594,1.845);
\draw[gp path] (1.504,1.904)--(1.684,1.904);
\node[gp node right] at (1.320,1.904) {$100$};
\draw[gp path] (1.504,2.291)--(1.594,2.291);
\draw[gp path] (1.504,2.518)--(1.594,2.518);
\draw[gp path] (1.504,2.679)--(1.594,2.679);
\draw[gp path] (1.504,2.803)--(1.594,2.803);
\draw[gp path] (1.504,2.905)--(1.594,2.905);
\draw[gp path] (1.504,2.992)--(1.594,2.992);
\draw[gp path] (1.504,3.066)--(1.594,3.066);
\draw[gp path] (1.504,3.132)--(1.594,3.132);
\draw[gp path] (1.504,3.191)--(1.684,3.191);
\node[gp node right] at (1.320,3.191) {$1000$};
\node[gp node center] at (2.236,0.308) {CNST};
\node[gp node center] at (3.699,0.308) {PI~\cite{5growth-scaling}};
\node[gp node center] at (5.163,0.308) {TES~\cite{v2n-access}};
\node[gp node center] at (6.627,0.308) {DDPG~\cite{noms}};
\node[gp node center] at (8.090,0.308) {DHPG};
\draw[gp path] (1.504,3.191)--(1.504,0.616)--(8.822,0.616);
\node[gp node center,rotate=-270] at (0.292,1.903) {runtime [$\mu$s]};
\gpfill{color=DeepPink3} (1.797,0.616)--(2.676,0.616)--(2.676,0.827)--(1.797,0.827)--cycle;
\gpcolor{rgb color={0.000,0.000,0.000}}
\draw[gp path] (1.797,0.616)--(1.797,0.826)--(2.675,0.826)--(2.675,0.616)--cycle;
\gpfill{color=Green4} (3.260,0.616)--(4.139,0.616)--(4.139,0.884)--(3.260,0.884)--cycle;
\draw[gp path] (3.260,0.616)--(3.260,0.883)--(4.138,0.883)--(4.138,0.616)--cycle;
\gpfill{color=Gold3} (4.724,0.616)--(5.603,0.616)--(5.603,1.246)--(4.724,1.246)--cycle;
\draw[gp path] (4.724,0.616)--(4.724,1.245)--(5.602,1.245)--(5.602,0.616)--cycle;
\gpfill{color=Firebrick3} (6.188,0.616)--(7.067,0.616)--(7.067,2.153)--(6.188,2.153)--cycle;
\draw[gp path] (6.188,0.616)--(6.188,2.152)--(7.066,2.152)--(7.066,0.616)--cycle;
\gpfill{color=DodgerBlue3} (7.651,0.616)--(8.530,0.616)--(8.530,2.729)--(7.651,2.729)--cycle;
\draw[gp path] (7.651,0.616)--(7.651,2.728)--(8.529,2.728)--(8.529,0.616)--cycle;
\draw[gp path] (2.236,0.819)--(2.236,0.833);
\draw[gp path] (2.146,0.819)--(2.326,0.819);
\draw[gp path] (2.146,0.833)--(2.326,0.833);
\draw[gp path] (3.699,0.877)--(3.699,0.889);
\draw[gp path] (3.609,0.877)--(3.789,0.877);
\draw[gp path] (3.609,0.889)--(3.789,0.889);
\draw[gp path] (5.163,1.241)--(5.163,1.250);
\draw[gp path] (5.073,1.241)--(5.253,1.241);
\draw[gp path] (5.073,1.250)--(5.253,1.250);
\draw[gp path] (6.627,2.150)--(6.627,2.153);
\draw[gp path] (6.537,2.150)--(6.717,2.150);
\draw[gp path] (6.537,2.153)--(6.717,2.153);
\draw[gp path] (8.090,2.714)--(8.090,2.742);
\draw[gp path] (8.000,2.714)--(8.180,2.714);
\draw[gp path] (8.000,2.742)--(8.180,2.742);
\gpcolor{color=gp lt color border}
\node[gp node center] at (2.236,1.134) {{14.55}};
\node[gp node center] at (3.699,1.191) {{16.12}};
\node[gp node center] at (5.163,1.553) {{30.83}};
\node[gp node center] at (6.627,2.460) {{155.87}};
\node[gp node center] at (8.090,3.036) {{436.96}};
\draw[gp path] (1.504,3.191)--(1.504,0.616)--(8.822,0.616);
%% coordinates of the plot area
\gpdefrectangularnode{gp plot 1}{\pgfpoint{1.504cm}{0.616cm}}{\pgfpoint{8.822cm}{3.191cm}}
\end{tikzpicture}
%% gnuplot variables
    \caption{
    Average runtime of each algorithm
    when performing the placement and scaling
    in the considered 5~\gls{pop}s upon the
    arrival of a vehicle.}
    \label{fig:run-time}
    % \vspace{-3mm}
\end{figure}

Overall, the \gls{ddpg}
and \gls{dhpg}
solutions have higher
computational complexity, as shown in
Fig.~\ref{fig:run-time}, which
illustrates the average runtime of each
algorithm for the experiments conducted in 
Sub-sections~\ref{subsec:training}-\ref{subsec:ddpg-behaviour}.
Nevertheless, they still
achieve runtimes in the sub-msec
range. Thus, the proposed
% \gls{ddpg}
\gls{dhpg} solution is suitable
for \gls{cv2n} services with
strict latency requirements. Note that taking
$\leq500\ \mu$s to perform the placement
and scaling is fast enough for \gls{cv2n}
services
with down to $d_{tgt}=10$ ms latency requirements -- not
to mention the considered service in
Section~\ref{sec:evaluation},
with $d_{tgt}=100$ ms.

\subsection{Discussion on Reward Function}
\label{subsec:main-flaws}

\label{sec:discussion}
\begin{figure}[t]
   \input{Figures/all-delays99-dens-overloads.tex}
   \caption{
   Tail of the \gls{epdf} for vehicles' 99-percentile latency
   bound $\overline{d_v^{99}}$.
   Prominent peaks on DDPG
   are due to CPU overload \mbox{$\rho_p>1$} on
   multiple \gls{pop}s.}
   \label{fig:all-delays-dens-overloads}
\end{figure}

\begin{figure}
    \begin{tikzpicture}[gnuplot]
%% generated with GNUPLOT 5.2p8 (Lua 5.3; terminal rev. Nov 2018, script rev. 108)
%% jue 20 abr 2023 10:38:58
\path (0.000,0.000) rectangle (8.750,3.500);
\gpcolor{color=gp lt color border}
\gpsetlinetype{gp lt border}
\gpsetdashtype{gp dt solid}
\gpsetlinewidth{1.00}
\draw[gp path] (0.768,2.575)--(0.768,0.523)--(8.197,0.523);
\node[gp node left,rotate=90] at (3.675,0.780) {$d_{tgt}$};
\gpcolor{rgb color={0.200,0.200,0.200}}
\gpsetdashtype{gp dt 2}
\draw[gp path](3.998,0.523)--(3.998,2.250);
\gpcolor{color=gp lt color border}
\node[gp node center,rotate=-270] at (0.292,1.549) {reward};
\node[gp node center] at (4.482,0.215) {$d$};
\node[gp node left] at (1.393,3.166) {$R(d)$};
\gpcolor{color=DodgerBlue4}
\gpsetdashtype{gp dt solid}
\gpsetlinewidth{5.00}
\draw[gp path] (0.293,3.166)--(1.209,3.166);
\draw[gp path] (0.768,0.523)--(0.843,0.588)--(0.918,0.654)--(0.993,0.719)--(1.068,0.784)%
  --(1.143,0.848)--(1.218,0.912)--(1.293,0.975)--(1.368,1.038)--(1.443,1.100)--(1.518,1.161)%
  --(1.593,1.220)--(1.668,1.279)--(1.744,1.337)--(1.819,1.393)--(1.894,1.448)--(1.969,1.501)%
  --(2.044,1.553)--(2.119,1.603)--(2.194,1.652)--(2.269,1.699)--(2.344,1.744)--(2.419,1.788)%
  --(2.494,1.829)--(2.569,1.869)--(2.644,1.906)--(2.719,1.942)--(2.794,1.976)--(2.869,2.007)%
  --(2.944,2.037)--(3.019,2.064)--(3.094,2.090)--(3.169,2.113)--(3.244,2.134)--(3.319,2.153)%
  --(3.394,2.170)--(3.469,2.185)--(3.544,2.198)--(3.620,2.209)--(3.695,2.217)--(3.770,2.224)%
  --(3.845,2.229)--(3.920,2.232)--(3.995,2.233)--(4.070,2.232)--(4.145,2.230)--(4.220,2.225)%
  --(4.295,2.219)--(4.370,2.211)--(4.445,2.202)--(4.520,2.191)--(4.595,2.179)--(4.670,2.165)%
  --(4.745,2.150)--(4.820,2.133)--(4.895,2.115)--(4.970,2.097)--(5.045,2.077)--(5.120,2.055)%
  --(5.195,2.033)--(5.270,2.010)--(5.345,1.987)--(5.421,1.962)--(5.496,1.937)--(5.571,1.911)%
  --(5.646,1.884)--(5.721,1.857)--(5.796,1.830)--(5.871,1.802)--(5.946,1.773)--(6.021,1.745)%
  --(6.096,1.716)--(6.171,1.687)--(6.246,1.658)--(6.321,1.629)--(6.396,1.600)--(6.471,1.570)%
  --(6.546,1.541)--(6.621,1.512)--(6.696,1.483)--(6.771,1.455)--(6.846,1.426)--(6.921,1.398)%
  --(6.996,1.370)--(7.071,1.342)--(7.146,1.315)--(7.221,1.288)--(7.297,1.262)--(7.372,1.236)%
  --(7.447,1.210)--(7.522,1.185)--(7.597,1.161)--(7.672,1.137)--(7.747,1.113)--(7.822,1.090)%
  --(7.897,1.068)--(7.972,1.046)--(8.047,1.024)--(8.122,1.004)--(8.197,0.983);
\gpsetpointsize{10.00}
\gppoint{gp mark 6}{(0.768,0.523)}
\gppoint{gp mark 6}{(1.518,1.161)}
\gppoint{gp mark 6}{(2.269,1.699)}
\gppoint{gp mark 6}{(3.019,2.064)}
\gppoint{gp mark 6}{(3.770,2.224)}
\gppoint{gp mark 6}{(4.520,2.191)}
\gppoint{gp mark 6}{(5.270,2.010)}
\gppoint{gp mark 6}{(6.021,1.745)}
\gppoint{gp mark 6}{(6.771,1.455)}
\gppoint{gp mark 6}{(7.522,1.185)}
\gppoint{gp mark 6}{(0.751,3.166)}
\gpcolor{color=gp lt color border}
\node[gp node left] at (1.393,2.858) {$R^*(d;\sigma=30,K=75)$};
\gpcolor{color=DodgerBlue4!75}
\gpsetlinewidth{3.00}
\draw[gp path] (0.293,2.858)--(1.209,2.858);
\draw[gp path] (0.768,0.523)--(0.843,0.588)--(0.918,0.654)--(0.993,0.719)--(1.068,0.784)%
  --(1.143,0.848)--(1.218,0.912)--(1.293,0.975)--(1.368,1.038)--(1.443,1.100)--(1.518,1.161)%
  --(1.593,1.220)--(1.668,1.279)--(1.744,1.337)--(1.819,1.393)--(1.894,1.448)--(1.969,1.501)%
  --(2.044,1.553)--(2.119,1.603)--(2.194,1.652)--(2.269,1.699)--(2.344,1.744)--(2.419,1.788)%
  --(2.494,1.829)--(2.569,1.869)--(2.644,1.906)--(2.719,1.942)--(2.794,1.976)--(2.869,2.007)%
  --(2.944,2.037)--(3.019,2.064)--(3.094,2.090)--(3.169,2.113)--(3.244,2.134)--(3.319,2.153)%
  --(3.394,2.170)--(3.469,2.185)--(3.544,2.198)--(3.620,2.209)--(3.695,2.217)--(3.770,2.224)%
  --(3.845,2.229)--(3.920,2.232)--(3.995,2.233)--(4.070,2.226)--(4.145,2.211)--(4.220,2.187)%
  --(4.295,2.153)--(4.370,2.110)--(4.445,2.058)--(4.520,2.000)--(4.595,1.936)--(4.670,1.866)%
  --(4.745,1.792)--(4.820,1.715)--(4.895,1.635)--(4.970,1.555)--(5.045,1.475)--(5.120,1.396)%
  --(5.195,1.319)--(5.270,1.244)--(5.345,1.172)--(5.421,1.104)--(5.496,1.040)--(5.571,0.981)%
  --(5.646,0.925)--(5.721,0.875)--(5.796,0.829)--(5.871,0.787)--(5.946,0.749)--(6.021,0.716)%
  --(6.096,0.687)--(6.171,0.661)--(6.246,0.639)--(6.321,0.619)--(6.396,0.603)--(6.471,0.589)%
  --(6.546,0.577)--(6.621,0.567)--(6.696,0.558)--(6.771,0.551)--(6.846,0.546)--(6.921,0.541)%
  --(6.996,0.537)--(7.071,0.534)--(7.146,0.532)--(7.221,0.530)--(7.297,0.528)--(7.372,0.527)%
  --(7.447,0.526)--(7.522,0.525)--(7.597,0.525)--(7.672,0.524)--(7.747,0.524)--(7.822,0.524)%
  --(7.897,0.524)--(7.972,0.523)--(8.047,0.523)--(8.122,0.523)--(8.197,0.523);
\gpsetpointsize{8.00}
\gppoint{gp mark 1}{(0.768,0.523)}
\gppoint{gp mark 1}{(1.518,1.161)}
\gppoint{gp mark 1}{(2.269,1.699)}
\gppoint{gp mark 1}{(3.019,2.064)}
\gppoint{gp mark 1}{(3.770,2.224)}
\gppoint{gp mark 1}{(4.520,2.000)}
\gppoint{gp mark 1}{(5.270,1.244)}
\gppoint{gp mark 1}{(6.021,0.716)}
\gppoint{gp mark 1}{(6.771,0.551)}
\gppoint{gp mark 1}{(7.522,0.525)}
\gppoint{gp mark 1}{(0.751,2.858)}
\gpcolor{color=gp lt color border}
\node[gp node left] at (5.437,3.166) {$R^*(d;\sigma=10,K=25)$};
\gpcolor{color=DodgerBlue4!50}
\draw[gp path] (4.337,3.166)--(5.253,3.166);
\draw[gp path] (0.768,0.523)--(0.843,0.588)--(0.918,0.654)--(0.993,0.719)--(1.068,0.784)%
  --(1.143,0.848)--(1.218,0.912)--(1.293,0.975)--(1.368,1.038)--(1.443,1.100)--(1.518,1.161)%
  --(1.593,1.220)--(1.668,1.279)--(1.744,1.337)--(1.819,1.393)--(1.894,1.448)--(1.969,1.501)%
  --(2.044,1.553)--(2.119,1.603)--(2.194,1.652)--(2.269,1.699)--(2.344,1.744)--(2.419,1.788)%
  --(2.494,1.829)--(2.569,1.869)--(2.644,1.906)--(2.719,1.942)--(2.794,1.976)--(2.869,2.007)%
  --(2.944,2.037)--(3.019,2.064)--(3.094,2.090)--(3.169,2.113)--(3.244,2.134)--(3.319,2.153)%
  --(3.394,2.170)--(3.469,2.185)--(3.544,2.198)--(3.620,2.209)--(3.695,2.217)--(3.770,2.224)%
  --(3.845,2.229)--(3.920,2.232)--(3.995,2.233)--(4.070,2.187)--(4.145,2.061)--(4.220,1.870)%
  --(4.295,1.641)--(4.370,1.402)--(4.445,1.178)--(4.520,0.985)--(4.595,0.832)--(4.670,0.719)%
  --(4.745,0.640)--(4.820,0.590)--(4.895,0.559)--(4.970,0.541)--(5.045,0.532)--(5.120,0.527)%
  --(5.195,0.525)--(5.270,0.524)--(5.345,0.523)--(5.421,0.523)--(5.496,0.523)--(5.571,0.523)%
  --(5.646,0.523)--(5.721,0.523)--(5.796,0.523)--(5.871,0.523)--(5.946,0.523)--(6.021,0.523)%
  --(6.096,0.523)--(6.171,0.523)--(6.246,0.523)--(6.321,0.523)--(6.396,0.523)--(6.471,0.523)%
  --(6.546,0.523)--(6.621,0.523)--(6.696,0.523)--(6.771,0.523)--(6.846,0.523)--(6.921,0.523)%
  --(6.996,0.523)--(7.071,0.523)--(7.146,0.523)--(7.221,0.523)--(7.297,0.523)--(7.372,0.523)%
  --(7.447,0.523)--(7.522,0.523)--(7.597,0.523)--(7.672,0.523)--(7.747,0.523)--(7.822,0.523)%
  --(7.897,0.523)--(7.972,0.523)--(8.047,0.523)--(8.122,0.523)--(8.197,0.523);
\gppoint{gp mark 2}{(0.768,0.523)}
\gppoint{gp mark 2}{(1.518,1.161)}
\gppoint{gp mark 2}{(2.269,1.699)}
\gppoint{gp mark 2}{(3.019,2.064)}
\gppoint{gp mark 2}{(3.770,2.224)}
\gppoint{gp mark 2}{(4.520,0.985)}
\gppoint{gp mark 2}{(5.270,0.524)}
\gppoint{gp mark 2}{(6.021,0.523)}
\gppoint{gp mark 2}{(6.771,0.523)}
\gppoint{gp mark 2}{(7.522,0.523)}
\gppoint{gp mark 2}{(4.795,3.166)}
\gpcolor{color=gp lt color border}
\gpsetlinewidth{1.00}
\draw[gp path] (0.768,2.575)--(0.768,0.523)--(8.197,0.523);
%% coordinates of the plot area
\gpdefrectangularnode{gp plot 1}{\pgfpoint{0.768cm}{0.523cm}}{\pgfpoint{8.197cm}{2.575cm}}
\end{tikzpicture}
%% gnuplot variables
    \vspace{-1em}
    \caption{Candidate rewards for future
    work $R^*(d;\sigma,K)$, and the reward
    $R(d)$ used in our experiments.
    $R^*(d;\sigma,K)$ alike rewards will
    punish more the under-provisioning, i.e.
    $d>d_{tgt}$.}
    \label{fig:piece-reward}
\end{figure}

% why the gap
We hereby discuss the implications of the
evaluation of the proposed approach for task placement and scaling of edge resources. 
Namely, we discuss how the reward
function impacts the optimality gap, and
how to reshape it to further boost
\gls{dhpg} performance.

 %The fact that we couple task placement and scaling decisions, instead of addressing it as a joint problem, leads to sub-optimal solutions as we observed in Section~\ref{subsec:optimality-gap}.
%%A DRL approach (i.e., DDPG) that would take both placement and scaling decisions -- formula \eqref{eq:action} -- would account for their correlations, hence would avoid under/over-provisioning computational resources, addressing performance requirements. 
  %as opposed to the proposed approach where we  solve the problems simultaneously yet independently and couple them during  
% Furthermore, the reward function
The optimality gap of \gls{dhpg} may be attributed to the design of the reward function in formula \eqref{eq:reward},
which does not decay fast when the experienced delay exceeds the target -- see $d>d_{tgt}$ values in Fig.~\ref{fig:reward}.
Although the \gls{dhpg} agent learns
to meet the target 99 delay percentile
$d=d_{tgt}$, we can still observe extreme
values for the 99 delay percentile (with very small probability) -- as the prominent peaks indicate in  Fig.~\ref{fig:all-delays-dens-overloads}.
% which is showing the full delay range of  Fig.~\ref{fig:all-delays-dens} for DDPG-1.
We argue that a faster decay in the
reward for $d>d_{tgt}$ may reduce the chances of overloading, thus preventing delay violations.
%Although the considered scenario is elementary it signifies the main drawback of our reward
%function: its \emph{slow decay}.
%  Specifically, DDPG-1/-5 may benefit from faster
% decays as $d\to\infty$ in order to avoid under-provisioning i.e., $R(d)\to-\infty$
% as $d\to\infty$. 
%In particular, we can prevent 
%switching off CPUs when vehicles are served by a PoP if $R(d)\to-\infty$
%as $d\to\infty$.
In this direction the following reward function is provided as an example, also illustrated in Fig.~\ref{fig:piece-reward}.
\begin{equation}
    R^*(d;\sigma,K)=
    \begin{cases}
        R(d), & d<d_{tgt}\\
        \frac{K}{\sigma}\frac{\phi\left(\frac{d-d_{tgt}}{s}\right)}{\Phi\left(\frac{b-d_{tgt}}{\sigma} \right) - \Phi\left(\frac{a-d_{tgt}}{\sigma} \right)}, & d\geq d_{tgt}
    \end{cases}
    \label{eq:piece-reward}
\end{equation}
with $d\geq d_{tgt}$ being a $K$-scaled
truncated normal
distribution~\cite{continuous-univariate}
in the interval $[a,b]=\mathbb{R}^+$,
centered at
$d_{tgt}$ with standard deviation $\sigma$;
$\phi(x)=1/\sqrt{2\phi}\exp(-\tfrac{1}{2}x^2)$;
and
$\Phi(x)=\tfrac{1}{2}(1+\erf(x/\sqrt{2\pi}))$.
Note that the $K$ scaling is set to ensure
the reward continuity, i.e.
$\lim_{\varepsilon\to0}R^*(d+\varepsilon;\sigma,K)=\lim_{\varepsilon\to0}R^*(d-\varepsilon;\sigma,K)$.
Moreover, $R^*(d;\sigma,K)$ is differentiable
at $d=d_{tgt}$ to help learning stability.

\section{Conclusion} 
\label{sec:conclusion}

 In this paper we studied \gls{cv2n} service provisioning supported by the edge cloud. %, where the \gls{cots} servers in \glspl{pop}, on top of providing connectivity, also perform the tasks brought by the vehicles. 
%Since we want to exploit the available (shared) resources as good as possible 
More specifically, we propose a DRL approach for \gls{cv2n} application task placement and scaling of edge computing resources.
% In each \gls{pop} a placement and scaling agent is running: the former is responsible for placing the tasks associated to a vehicle to the most appropriate \gls{pop} (discrete), while the latter is dealing with allocating \glspl{cpu} resources in each \gls{pop} (continuous), to efficiently support the \gls{cv2n} application.
Task placement refers to the discrete problem of placing application tasks associated to each vehicle to the most appropriate \gls{pop}. Resource scaling is dealing with discrete problem of scaling \glspl{cpu} resources in each \gls{pop}, to efficiently support the \gls{cv2n} application tasks.
As shown however in \cite{noms}, we transfer the resource scaling problem to a continuous action space, to address the scalability issues inherent to the high-dimensional discrete action spaces.
% To promote cost-effective use of the edge resources and avoid overbooking them at \gls{pop} level, application tasks of a vehicle arriving in the vicinity of one \gls{pop} can be processed at a nearby PoP in the service area, taking into account the additional transmission delay. % which we refer to as placement (the tasks to another \gls{pop}).
To optimize the cost-efficient utilization of edge resources and meet the delay constraints of the \gls{cv2n} service, application tasks associated with a vehicle entering a service area can be offloaded to any nearby \gls{pop} taking also into account the additional transmission latency.
%Although this involves  additional transmission latency, it allows to tap into the (possible plentiful) resources of that other \gls{pop}. 
%Moreover, we try to foresee in each of the \glspl{pop} just enough resources, which we refer to as scaling (the resources in the \glspl{pop}). 
%The decisions are taken based on the service delay that the tasks experience.
To tackle the two interdependent problems with heterogeneous action spaces, we propose \gls{dhpg}, a deep reinforcement learning approach designed to handle hybrid action spaces. This enables \gls{dhpg} to make holistic decisions that consider both task placement and resource scaling concurrently, leading to improved performance in C-V2N service provisioning.
%joint scaling decisions for all \glspl{pop} together (\gls{ddpg}-5) and one that makes scaling decisions in each \gls{pop} separately (\gls{ddpg}-1).

% We compare the performance of these scaling approaches against state-of-the-art ones that base their decisions on the load that is observed rather than on the delays.
We compare the performance of \gls{dhpg} against state-of-the-art scaling algorithms, including a traditional \gls{pi} controller, a forecast-based n-max approach (\gls{tes}), and a \gls{ddpg}-based approach, coupled with a greedy-based placement agent.
% We found that the \gls{pi} approach behaves conservatively, keeping the load (thus delay) quite low and consequently demanding more edge resources.
%The \gls{tes} approach performs slightly worse than the \gls{ddpg} ones, but it relies on being provided with the exact period of seasonality.
% In other words, the \gls{dhpg} is capable of capturing the seasonality without requesting it as an additional input parameter.
%Our evaluations reveal that the \gls{pi} controller exhibits a more conservative behavior, resulting in lower delays and higher utilization of edge resources.  
\gls{dhpg} demonstrates its capability to learn and exploit the inherent seasonality in the workload.
% This enables \gls{dhpg} to adopt a more aggressive policy, achieving comparable or lower delays while utilizing edge resources efficiently.
This enables \gls{dhpg} to adopt a more aggressive policy, resulting in performance improvements ranging from 17.4\% to 52.8\% against the considered state-of-the-art algorithms.

% Comparing the centralized version of the \gls{ddpg} (\gls{ddpg}-5) to the decentralized one (\gls{ddpg}-1), we conclude that the latter performs slightly better, as is distributes the loads more evenly over the various \glspl{pop} and in turn results in higher average reward and less delay violations.
%Furthermore, DDPG-1 requires no synchronization of the information regarding the workload and resource dynamics for the rest of the PoPs.
% Furthermore, we compared \gls{dhpg} to the oracle solution that employs full knowledge of all vehicles' arrivals in the future. Our proposed \gls{dhpg} presents a 5.2\% optimality gap. Although the comparison is somewhat unfair as our solution is only based on causal information, it hints the fact that there is still room for improvement.
We further benchmarked DHPG against an oracle solution assuming knowledge of all future vehicle arrivals. Our proposed \gls{dhpg} achieves an optimality gap of 5.2\%, despite the fact that it  relies solely on causal information. % it suggests potential for further improvement in \gls{dhpg}'s performance.
%In future work we plan to tackle the identified gaps of the proposed \gls{dhpg} agents by tuning the shape of the reward function around $d_{tgt}$ for better control the degree of allowed SLA violation;%imposing a drastic reward drop to the reward function as$d>d_{tgt}$;
In future work, we will explore various reward functions associated with the target delay.
%Moreover, we will explore extensions of \gls{dhpg} that can handle correlated \gls{pop}s at a larger scale.
Moreover, extensions of \gls{dhpg} will be explored for handling correlated \gls{pop}s on a larger scale.

%o address the limitations identified in the proposed \gls{dhpg} agents by shaping the reward function around the target delay $d_{tgt}$ to achieve finer-grained control over allowable SLA violations. 
\section*{Acknowledgement}
This work has been performed in the framework of the (i) Horizon SNS JU DESIRE6G (No. 101096466); (ii) National Growth Fund through the Dutch 6G flagship project ``Future Network Services''; % 5Growth (Grant Agreement No. 856709)) 
and
(iii) 
Remote Driver Project of the Spanish Ministry of Economic Affairs and Digital Transformation under Grant TSI-065100-2022-003;


\bibliographystyle{IEEEtran}
\bibliography{References/references.bib}

\begin{thebibliography}{10}
\providecommand{\url}[1]{#1}
\csname url@samestyle\endcsname
\providecommand{\newblock}{\relax}
\providecommand{\bibinfo}[2]{#2}
\providecommand{\BIBentrySTDinterwordspacing}{\spaceskip=0pt\relax}
\providecommand{\BIBentryALTinterwordstretchfactor}{4}
\providecommand{\BIBentryALTinterwordspacing}{\spaceskip=\fontdimen2\font plus
\BIBentryALTinterwordstretchfactor\fontdimen3\font minus \fontdimen4\font\relax}
\providecommand{\BIBforeignlanguage}[2]{{%
\expandafter\ifx\csname l@#1\endcsname\relax
\typeout{** WARNING: IEEEtran.bst: No hyphenation pattern has been}%
\typeout{** loaded for the language `#1'. Using the pattern for}%
\typeout{** the default language instead.}%
\else
\language=\csname l@#1\endcsname
\fi
#2}}
\providecommand{\BIBdecl}{\relax}
\BIBdecl

\bibitem{3GPPv2xrequirements}
``{Technical Specification Group Services and System Aspects;Enhancement of 3{G}{P}{P} support for V2X scenarios;},'' 3GPP, Technical Specification 22.186.v17.0.0, April 2022.

\bibitem{tutorial-v2x}
M.~H.~C. Garcia, A.~Molina-Galan, M.~Boban, J.~Gozalvez, B.~Coll-Perales, T.~Şahin, and A.~Kousaridas, ``{A Tutorial on 5G NR V2X Communications},'' \emph{IEEE Communications Surveys Tutorials}, vol.~23, no.~3, pp. 1972--2026, 2021.

\bibitem{rammohan2023revolutionizing}
A.~Rammohan \emph{et~al.}, ``Revolutionizing intelligent transportation systems with cellular vehicle-to-everything ({C-V2X}) technology: Current trends, use cases, emerging technologies, standardization bodies, industry analytics and future directions,'' \emph{Vehicular Communications}, p. 100638, 2023.

\bibitem{3GPPV2X1}
{3GPP TS 23.28}, ``{Architecture enhancements for V2X services (Release 16)},'' Tech. Rep., March 2019.

\bibitem{AHMED20224135}
M.~Ahmed, S.~Raza, M.~A. Mirza, A.~Aziz, M.~A. Khan, W.~U. Khan, J.~Li, and Z.~Han, ``A survey on vehicular task offloading: Classification, issues, and challenges,'' \emph{Journal of King Saud University - Computer and Information Sciences}, vol.~34, no.~7, pp. 4135--4162, 2022.

\bibitem{noor20226g}
M.~Noor-A-Rahim, Z.~Liu, H.~Lee, M.~O. Khyam, J.~He, D.~Pesch, K.~Moessner, W.~Saad, and H.~V. Poor, ``{6G} for vehicle-to-everything ({V2X}) communications: Enabling technologies, challenges, and opportunities,'' \emph{Proceedings of the IEEE}, vol. 110, no.~6, pp. 712--734, 2022.

\bibitem{noms}
C.~S.-H. Hsu, J.~Martín-Pérez, C.~Papagianni, and P.~Grosso, ``{V2N} service scaling with deep reinforcement learning,'' in \emph{NOMS 2023-2023 IEEE/IFIP Network Operations and Management Symposium}, 2023, pp. 1--5.

\bibitem{DeVleeschauwer2021}
D.~De~Vleeschauwer, J.~Baranda, J.~Mangues-Bafalluy, C.~F. Chiasserini, M.~Malinverno, C.~Puligheddu, L.~Magoula, J.~Martín-Pérez, S.~Barmpounakis, K.~Kondepu, L.~Valcarenzhi, X.~Li, C.~Papagianni, and A.~Garcia-Saavedra, ``{5Growth Data-Driven AI-Based Scaling},'' in \emph{2021 Joint European Conference on Networks and Communications 6G Summit (EuCNC/6G Summit)}, June 2021, pp. 383--388.

\bibitem{v2n-access}
J.~Martín-Pérez, K.~Kondepu, D.~De~Vleeschauwer, V.~Reddy, C.~Guimarães, A.~Sgambelluri, L.~Valcarenghi, C.~Papagianni, and C.~J. Bernardos, ``{Dimensioning {V}2{N} Services in 5{G} Networks Through Forecast-Based Scaling},'' \emph{IEEE Access}, vol.~10, pp. 9587--9602, 2022.

\bibitem{10.1007/s10586-021-03265-9}
S.~Verma and A.~Bala, ``{Auto-Scaling Techniques for IoT-Based Cloud Applications: A Review},'' \emph{Cluster Computing}, vol.~24, no.~3, p. 2425–2459, sep 2021.

\bibitem{8422788}
S.~Rahman, T.~Ahmed, M.~Huynh, M.~Tornatore, and B.~Mukherjee, ``{Auto-Scaling VNFs Using Machine Learning to Improve QoS and Reduce Cost},'' in \emph{2018 IEEE International Conference on Communications (ICC)}, 2018, pp. 1--6.

\bibitem{8806631}
T.~Subramanya and R.~Riggio, ``{Machine Learning-Driven Scaling and Placement of Virtual Network Functions at the Network Edges},'' in \emph{2019 IEEE Conference on Network Softwarization (NetSoft)}, 2019, pp. 414--422.

\bibitem{8540003}
R.~Li, Z.~Zhao, Q.~Sun, C.-L. I, C.~Yang, X.~Chen, M.~Zhao, and H.~Zhang, ``Deep reinforcement learning for resource management in network slicing,'' \emph{IEEE Access}, vol.~6, pp. 74\,429--74\,441, 2018.

\bibitem{Winters1960TES}
P.~R. Winters, ``{Forecasting sales by exponentially weighted moving averages},'' \emph{Management science}, vol.~6, no.~3, pp. 324--342, 1960.

\bibitem{LEE201166}
Y.-S. Lee and L.-I. Tong, ``{Forecasting time series using a methodology based on autoregressive integrated moving average and genetic programming},'' \emph{Knowledge-Based Systems}, vol.~24, no.~1, pp. 66--72, 2011.

\bibitem{lossleap}
A.~Collet, A.~Banchs, and M.~Fiore, ``{LossLeaP: Learning to Predict for Intent-Based Networking},'' in \emph{IEEE INFOCOM 2022 - IEEE Conference on Computer Communications}, 2022, pp. 2138--2147.

\bibitem{https://doi.org/10.1049/iet-its.2016.0208}
Z.~Zhao, W.~Chen, X.~Wu, P.~C.~Y. Chen, and J.~Liu, ``{LSTM network: a deep learning approach for short-term traffic forecast},'' \emph{IET Intelligent Transport Systems}, vol.~11, no.~2, pp. 68--75, 2017.

\bibitem{aztec}
D.~Bega, M.~Gramaglia, M.~Fiore, A.~Banchs, and X.~Costa-Perez, ``{AZTEC: Anticipatory Capacity Allocation for Zero-Touch Network Slicing},'' in \emph{IEEE INFOCOM 2020 - IEEE Conference on Computer Communications}, 2020, pp. 794--803.

\bibitem{sdgnet}
Y.~Fang, S.~Ergüt, and P.~Patras, ``{SDGNet: A Handover-Aware Spatiotemporal Graph Neural Network for Mobile Traffic Forecasting},'' \emph{IEEE Communications Letters}, vol.~26, no.~3, pp. 582--586, 2022.

\bibitem{8542668}
C.~Zhang, H.~Zhao, and S.~Deng, ``A density-based offloading strategy for {IoT} devices in edge computing systems,'' \emph{IEEE Access}, vol.~6, pp. 73\,520--73\,530, 2018.

\bibitem{8240666}
J.~Du, L.~Zhao, J.~Feng, and X.~Chu, ``Computation offloading and resource allocation in mixed fog/cloud computing systems with min-max fairness guarantee,'' \emph{IEEE Transactions on Communications}, vol.~66, no.~4, pp. 1594--1608, 2018.

\bibitem{7553459}
K.~Zhang, Y.~Mao, S.~Leng, Q.~Zhao, L.~Li, X.~Peng, L.~Pan, S.~Maharjan, and Y.~Zhang, ``Energy-efficient offloading for mobile edge computing in {5G} heterogeneous networks,'' \emph{IEEE Access}, vol.~4, pp. 5896--5907, 2016.

\bibitem{9046820}
C.~Pradhan, A.~Li, C.~She, Y.~Li, and B.~Vucetic, ``Computation offloading for {IoT} in {C-RAN}: Optimization and deep learning,'' \emph{IEEE Transactions on Communications}, vol.~68, no.~7, pp. 4565--4579, 2020.

\bibitem{ZHAO2019346}
X.~Zhao, K.~Yang, Q.~Chen, D.~Peng, H.~Jiang, X.~Xu, and X.~Shuang, ``Deep learning based mobile data offloading in mobile edge computing systems,'' \emph{Future Generation Computer Systems}, vol.~99, pp. 346--355, 2019.

\bibitem{8647611}
N.~Zhao, Y.-C. Liang, D.~Niyato, Y.~Pei, and Y.~Jiang, ``Deep reinforcement learning for user association and resource allocation in heterogeneous networks,'' in \emph{2018 IEEE Global Communications Conference (GLOBECOM)}, 2018, pp. 1--6.

\bibitem{9363256}
X.~Zhang, A.~Pal, and S.~Debroy, ``Deep reinforcement learning based energy-efficient task offloading for secondary mobile edge systems,'' in \emph{2020 IEEE 45th LCN Symposium on Emerging Topics in Networking (LCN Symposium)}, 2020, pp. 48--59.

\bibitem{8690980}
X.~Chen, H.~Zhang, C.~Wu, S.~Mao, Y.~Ji, and M.~Bennis, ``Performance optimization in mobile-edge computing via deep reinforcement learning,'' in \emph{2018 IEEE 88th Vehicular Technology Conference (VTC-Fall)}, 2018, pp. 1--6.

\bibitem{8761385}
X.~Liu, Z.~Qin, and Y.~Gao, ``Resource allocation for edge computing in {IoT} networks via reinforcement learning,'' in \emph{ICC 2019 - 2019 IEEE International Conference on Communications (ICC)}, 2019, pp. 1--6.

\bibitem{8771176}
L.~Huang, S.~Bi, and Y.-J.~A. Zhang, ``Deep reinforcement learning for online computation offloading in wireless powered mobile-edge computing networks,'' \emph{IEEE Transactions on Mobile Computing}, vol.~19, no.~11, pp. 2581--2593, 2020.

\bibitem{DDQNEC}
I.~Ullah, H.-K. Lim, Y.-J. Seok, and Y.-H. Han, ``Optimizing task offloading and resource allocation in edge-cloud networks: a {DRL} approach,'' \emph{J. Cloud Comput.}, vol.~12, no.~1, Jul. 2023.

\bibitem{10036357}
J.~Yang, Q.~Yuan, S.~Chen, H.~He, X.~Jiang, and X.~Tan, ``Cooperative task offloading for mobile edge computing based on multi-agent deep reinforcement learning,'' \emph{IEEE Transactions on Network and Service Management}, vol.~20, no.~3, pp. 3205--3219, 2023.

\bibitem{10.1109/TMC.2022.3150432}
H.~Jiang, X.~Dai, Z.~Xiao, and A.~Iyengar, ``Joint task offloading and resource allocation for energy-constrained mobile edge computing,'' \emph{IEEE Transactions on Mobile Computing}, vol.~22, no.~7, p. 4000–4015, Jul. 2023.

\bibitem{LIU20231399}
S.~Liu, J.~Tian, C.~Zhai, and T.~Li, ``Joint computation offloading and resource allocation in vehicular edge computing networks,'' \emph{Digital Communications and Networks}, vol.~9, no.~6, pp. 1399--1410, 2023.

\bibitem{li2022joint}
S.~Li, N.~Zhang, R.~Jiang, Z.~Zhou, F.~Zheng, and G.~Yang, ``Joint task offloading and resource allocation in mobile edge computing with energy harvesting,'' \emph{Journal of Cloud Computing}, vol.~11, no.~1, p.~17, 2022.

\bibitem{LAI2024110692}
P.~Lai, Y.~Tao, J.~Qin, Y.~Xie, S.~Zhang, S.~Tang, Q.~Huang, and S.~Liao, ``Joint optimization of application placement and resource allocation for enhanced performance in heterogeneous multi-server systems,'' \emph{Computer Networks}, vol. 253, p. 110692, 2024.

\bibitem{7914660}
T.~Q. Dinh, J.~Tang, Q.~D. La, and T.~Q.~S. Quek, ``Offloading in mobile edge computing: Task allocation and computational frequency scaling,'' \emph{IEEE Transactions on Communications}, vol.~65, no.~8, pp. 3571--3584, 2017.

\bibitem{MA}
W.~Zhou, W.~Fang, Y.~Li, B.~Yuan, Y.~Li, and T.~Wang, ``Markov approximation for task offloading and computation scaling in mobile edge computing,'' \emph{Mobile Information Systems}, vol. 2019, no.~1, p. 8172698, 2019.

\bibitem{chen2020decentralized}
Z.~Chen and X.~Wang, ``Decentralized computation offloading for multi-user mobile edge computing: A deep reinforcement learning approach,'' \emph{EURASIP Journal on Wireless Communications and Networking}, vol. 2020, no.~1, pp. 1--21, 2020.

\bibitem{9435782}
J.~Chen, H.~Xing, Z.~Xiao, L.~Xu, and T.~Tao, ``A {DRL} agent for jointly optimizing computation offloading and resource allocation in {MEC},'' \emph{IEEE Internet of Things Journal}, vol.~8, no.~24, pp. 17\,508--17\,524, 2021.

\bibitem{peng2020deep}
H.~Peng and X.~Shen, ``Deep reinforcement learning based resource management for multi-access edge computing in vehicular networks,'' \emph{IEEE Transactions on Network Science and Engineering}, vol.~7, no.~4, pp. 2416--2428, 2020.

\bibitem{10048752}
Z.~Nan, S.~Zhou, Y.~Jia, and Z.~Niu, ``Joint task offloading and resource allocation for vehicular edge computing with result feedback delay,'' \emph{IEEE Transactions on Wireless Communications}, vol.~22, no.~10, pp. 6547--6561, 2023.

\bibitem{WANG2023}
S.~Wang, X.~Song, H.~Xu, T.~Song, G.~Zhang, and Y.~Yang, ``Joint offloading decision and resource allocation in vehicular edge computing networks,'' \emph{Digital Communications and Networks}, 2023.

\bibitem{huang2023joint}
J.~Huang, J.~Wan, B.~Lv, Q.~Ye, and Y.~Chen, ``Joint computation offloading and resource allocation for edge-cloud collaboration in internet of vehicles via deep reinforcement learning,'' \emph{IEEE Systems Journal}, vol.~17, no.~2, pp. 2500--2511, 2023.

\bibitem{10024868}
Z.~Sun, G.~Sun, Y.~Liu, J.~Wang, and D.~Cao, ``Bargain-match: A game theoretical approach for resource allocation and task offloading in vehicular edge computing networks,'' \emph{IEEE Transactions on Mobile Computing}, vol.~23, no.~2, pp. 1655--1673, 2024.

\bibitem{islam2021survey}
A.~Islam, A.~Debnath, M.~Ghose, and S.~Chakraborty, ``A survey on task offloading in multi-access edge computing,'' \emph{Journal of Systems Architecture}, vol. 118, p. 102225, 2021.

\bibitem{vnf-sharing}
F.~Malandrino, C.~F. Chiasserini, G.~Einziger, and G.~Scalosub, ``{Reducing Service Deployment Cost Through VNF Sharing},'' \emph{IEEE/ACM Transactions on Networking}, vol.~27, no.~6, pp. 2363--2376, 2019.

\bibitem{near-optimal-placement}
R.~Cohen, L.~Lewin-Eytan, J.~S. Naor, and D.~Raz, ``{Near optimal placement of virtual network functions},'' in \emph{2015 IEEE Conference on Computer Communications (INFOCOM)}, 2015, pp. 1346--1354.

\bibitem{joint-placement-infocom}
S.~Agarwal, F.~Malandrino, C.-F. Chiasserini, and S.~De, ``{Joint VNF Placement and CPU Allocation in 5G},'' in \emph{IEEE INFOCOM 2018 - IEEE Conference on Computer Communications}, 2018, pp. 1943--1951.

\bibitem{heyman1982stochastic}
D.~P. Heyman and M.~J. Sobel, \emph{{Stochastic models in operations research. 1. Stochastic processes and operating characteristics}}.\hskip 1em plus 0.5em minus 0.4em\relax McGraw-Hill New York, NY, USA:, 1982.

\bibitem{atari}
V.~Mnih, K.~Kavukcuoglu, D.~Silver, A.~Graves, I.~Antonoglou, D.~Wierstra, and M.~Riedmiller, ``{Playing Atari with Deep Reinforcement Learning},'' 2013.

\bibitem{zhu2021overview}
J.~Zhu, F.~Wu, and J.~Zhao, ``An overview of the action space for deep reinforcement learning,'' in \emph{Proceedings of the 2021 4th International Conference on Algorithms, Computing and Artificial Intelligence}, 2021, pp. 1--10.

\bibitem{ddpg}
T.~P. Lillicrap, J.~J. Hunt, A.~Pritzel, N.~Heess, T.~Erez, Y.~Tassa, D.~Silver, and D.~Wierstra, ``{Continuous control with deep reinforcement learning},'' 2015.

\bibitem{5growth-scaling}
``{5GROWTH Scaling},'' \url{https://github.com/MartinPJorge/5growth-scaling/blob/master/FiftyStations/clean0326}, 2020.

\bibitem{sutton2018rl}
R.~S. Sutton and A.~G. Barto, \emph{{Reinforcement learning: An introduction}}.\hskip 1em plus 0.5em minus 0.4em\relax MIT press, 2018.

\bibitem{Alvarez-Mesa2012}
M.~Alvarez-Mesa, C.~C. Chi, B.~Juurlink, V.~George, and T.~Schierl, ``{Parallel video decoding in the emerging HEVC standard},'' in \emph{2012 IEEE International Conference on Acoustics, Speech and Signal Processing (ICASSP)}, 2012, pp. 1545--1548.

\bibitem{SHUSTANOV2017718}
A.~Shustanov and P.~Yakimov, ``{{C}{N}{N} Design for Real-Time Traffic Sign Recognition},'' \emph{Procedia Engineering}, vol. 201, pp. 718--725, 2017, 3rd International Conference “Information Technology and Nanotechnology", ITNT-2017, 25-27 April 2017, Samara, Russia.

\bibitem{ETSI.TR.126.985}
``{5G; Vehicle-to-everything (V2X); Media handling and interaction},'' ETSI, Sophia Antipolis - FRANCE, Standard, March 2020.

\bibitem{5G_Latency}
B.~Coll-Perales, M.~d.~C. Lucas-Estañ, T.~Shimizu, J.~Gozalvez, T.~Higuchi, S.~Avedisov, O.~Altintas, and M.~Sepulcre, ``{End-to-End V2X Latency Modeling and Analysis in 5G Networks},'' January 2022.

\bibitem{fiorevehdata}
{Uppoor, Sandesh and Trullols-Cruces, Oscar and Fiore, Marco and Barcelo-Ordinas, Jose M}, ``Generation and analysis of a large-scale urban vehicular mobility dataset,'' \emph{IEEE Transactions on Mobile Computing}, vol.~13, no.~5, pp. 1061--1075, 2013.

\bibitem{gao2023spatial}
{Gao, Haotian and Jiang, Renhe and Dong, Zheng and Deng, Jinliang and Ma, Yuxin and Song, Xuan}, ``Spatial-temporal-decoupled masked pre-training for spatiotemporal forecasting,'' \emph{arXiv preprint arXiv:2312.00516}, 2023.

\bibitem{wavelets-traffic}
{Fang, Yuchen and Qin, Yanjun and Luo, Haiyong and Zhao, Fang and Xu, Bingbing and Zeng, Liang and Wang, Chenxing}, ``When spatio-temporal meet wavelets: Disentangled traffic forecasting via efficient spectral graph attention networks,'' in \emph{2023 IEEE 39th International Conference on Data Engineering (ICDE)}, 2023, pp. 517--529.

\bibitem{Ang05pidcontrol}
{Kiam Heong Ang}, G.~{Chong}, and {Yun Li}, ``{{P}{I}{D} control system analysis, design, and technology},'' \emph{IEEE Transactions on Control Systems Technology}, vol.~13, no.~4, pp. 559--576, 2005.

\bibitem{elu}
D.-A. Clevert, T.~Unterthiner, and S.~Hochreiter, ``{Fast and Accurate Deep Network Learning by Exponential Linear Units (ELUs)},'' 2015.

\bibitem{adam}
D.~P. Kingma and J.~Ba, ``{Adam: A Method for Stochastic Optimization},'' 2014.

\bibitem{td3}
S.~Fujimoto, H.~van Hoof, and D.~Meger, ``{Addressing Function Approximation Error in Actor-Critic Methods},'' 2018.

\bibitem{ampl}
R.~e.~a. Fourer, ``{AMPL. A modeling language for mathematical programming},'' 1993.

\bibitem{knitro}
R.~A. Waltz and J.~Nocedal, ``{KNITRO 2.0 User’s Manual},'' pp. 33--34, 2004.

\bibitem{backprop}
{LeCun, Yann and Touresky, D and Hinton, G and Sejnowski, T}, ``A theoretical framework for back-propagation,'' in \emph{Proceedings of the 1988 connectionist models summer school}, vol.~1, 1988, pp. 21--28.

\bibitem{continuous-univariate}
{Johnson, Norman L and Kotz, Samuel and Balakrishnan, Narayanaswamy}, \emph{{Continuous univariate distributions}}.\hskip 1em plus 0.5em minus 0.4em\relax {John wiley \& sons}, 1995, vol. 289.

\bibitem{generalized-assignment}
T.~{\"O}ncan, ``{A survey of the generalized assignment problem and its applications},'' \emph{INFOR: Information Systems and Operational Research}, vol.~45, no.~3, pp. 123--141, 2007.

\bibitem{ciw}
G.~I. Palmer, V.~A. Knight, P.~R. Harper, and A.~L. Hawa, ``Ciw: An open-source discrete event simulation library,'' \emph{Journal of Simulation}, vol.~13, no.~1, pp. 68--82, 2019.

\end{thebibliography}

\appendices
\section{Proof of Lemma~\ref{lemma:complex}}
\label{app:proof}

\begin{proof}
    We proof the $\mathcal{NP}$-hardness of Problem~\ref{problem} (Section 2.2) by showing that an instance of our problem is equivalent to the generalized assignment   problem~\cite{generalized-assignment}.

    Let us consider an instance of Problem~\ref{problem} that we want to
    solve for a time window $T$, in particular, we take a problem
    instance with every vehicle departing after the considered
    time window, i.e, $T_v>T$ holds $\forall v$.

    When a vehicle $v$ arrives to the vicinity of \gls{pop} $p_v$
    we have to take a \emph{placement} decision to know that the
    \gls{cv2n} tasks of vehicle $v$ will be processed at \gls{pop} $p'_v$.
    Note that the \gls{cv2n} tasks of vehicle $v$ will induce a
    \gls{cpu} increase $C_{v,p'_v}^+$ at \gls{pop} $p'_v$
    to maximize the reward (i.e., to stay
    close to the target delay $d_{tgt}$ in Fig.~\ref{fig:reward}).
    Namely, the \gls{cpu} increase is expressed as:
    \begin{multline}
        C_{v,p_v'}^+ = \argmax_C \left\{R(C,p_v',N_{v-1,p_v'}+1)\right\} \\
        - \argmax_C\left\{ R(C,p_v',N_{v-1,p_v'}) \right\}
        \label{eq:cpu-increase}
    \end{multline}
    with $N_{v-1,p_v'}+1$ capturing that \gls{pop} $p_v'$ will
    accommodate the tasks of an additional vehicle, i.e., vehicle $v$.
    Note how in formula~\eqref{eq:cpu-increase} we make explicit
    the dependence on the number of vehicles $N_{v,p_v'}$
    within the reward function for the latter depends on the
    average processing delay $d_{p,v}$ experienced
    by the vehicle, which is impacted by $N_{v,p_v'}$ defined (Section 2.1) as: 
  %  -- see Section \ref{subsec:system-model}: formula (3). %~\eqref{eq:avg-delay}.
\begin{equation}
    %\label{eq:avg-delay}
    \mathbb{E}[d_{p,v}] =
    \begin{cases}
        \frac{1}{\mu(C_{p,v}) - \lambda N_{p,v}} & \mbox{if $\mu(C_{p,v}) >\lambda N_{p,v}$} \\
    \infty & \mbox{otherwise} 
    \end{cases}
\end{equation}

    With the \gls{pop} \gls{cpu} increase by each vehicle
    \eqref{eq:cpu-increase}
    we reformulate Problem~\ref{problem} as:
    \begin{align}
        \max_{p'_v} & \sum_v\sum_p R(C_{p,v},p_v',N_p) \mathds{1}_{p'_v}(p)\label{eq:max-reduction}\\
        s.t.: & \sum_v C_{v,p_v'}^+ \mathds{1}_{p_v'}(p) \leq C_p^{\max}, \quad \forall p\\
        & p'_v \in P\label{eq:in-pops}
    \end{align}

    %% Let us consider an ideal scenario where the
    %% arrival rate of \gls{cv2n} tasks $\lambda$ is
    %% so small that the processing delay in~\eqref{eq:avg-delay}
    %% is esentially $d_{p_v}=0$. Hence, the delay experienced by the vehicular services will
    %% only depend on the transmission latency towards the \gls{pop} processing the traffic $d_v=l_{p_v,p'_v}$. Let us also assume that the transmission latency $l_{p_v,p'_v}$ is constant over time. Then the reward function~\eqref{eq:reward} becomes a fixed parameter that only depends on the steering decision $R(C_{p,v},p'_v)=R_{p'_v}$. Additionally, we can also reduce our problem complexity by getting rid of the scaling decisions if we consider a system that cannot turn on/off \glspl{cpu}, so that $C_{p,v}=C_{p,v'},\ \forall p, v, v'$. Finally, we solve a problem instance over a time window $T$ satisfying that $T_v>T,\ \forall v$. As a result Problem~\ref{problem} becomes:
    %% \begin{align}
    %%     \max_{p'_v} & \sum_v\sum_p R_{p'_v} \mathds{1}_{p'_v}(p)\label{eq:max-reduction}\\
    %%     s.t.: & \sum_v 1\cdot\mathds{1}_{p'_v}(p)\leq N_p^{\max},\quad \forall p\label{eq:assign-weights}\\
    %%     %& \sum_p p'_v=1,\quad \forall p\\
    %%     %& p'_v\in \{0,1\}
    %%     & p'_v \in P\label{eq:in-pops}
    %% \end{align}

    We now indicate how the variables/parameters in the formulation above
    mimic to the notation used in the generalized assignment
    problem~\cite{generalized-assignment}.
    First, we take $x_{v,p}=\mathds{1}_{p'_v}(p)$ as the binary variable telling whether
    vehicle $v$ \gls{cv2n} tasks are processed at \gls{pop} $p$.
    Second, we write $R_{v,p}=R(C_p,p_v',N_p)$ as the profit/reward obtained by having
    \gls{pop} $p$ processing vehicle $v$ \gls{cv2n} tasks.
    Third, we denote $w_{v,p}=C_{v,p}^+$ as the \gls{cpu}
    ``weight'' that vehicle $v$ brings in to \gls{pop} $p$
    in terms of additional \gls{cpu}s.
    Finally, we note that~\eqref{eq:in-pops} means that every vehicle $v$
    \gls{cv2n} tasks must be processed at one \gls{pop} within the set $P$, i.e.,
    $\sum_p x_{v,p}=1$ must hold $\forall v\in V$.
    Overall, our reduction \mbox{in~\eqref{eq:max-reduction}-\eqref{eq:in-pops}}
    becomes:
    \begin{align}
        \max & \sum_v\sum_p  p_{v,p} \cdot x_{v,p}\\
        s.t.: & \sum_v w_{v,p}\cdot x_{v,p}\leq C_p^{\max},\quad \forall p\\
        %& \sum_p p'_v=1,\quad \forall p\\
        %& p'_v\in \{0,1\}
          & \sum_{p} x_{v,p}=1,\quad \forall p
    \end{align}
    which is the generalized assignment problem with vehicles $v$ being items that we have to place at a certain bins/\gls{pop}s $p$. In our case the profit is our
    reward function, and the bins' capacity is $C_p^{\max}$.

Therefore, we have found an instance of Problem~\ref{problem} that is equivalent to the generalized assignment problem~\cite{generalized-assignment}, which is $\mathcal{NP}$-hard. As a consequence, Problem~\ref{problem} is also $\mathcal{NP}$-hard.

\end{proof}

\section{Target latency}
\label{app:latency}

% reward funded on avg delay
In Section~\ref{sec:problem} we define the
optimization Problem~\ref{problem} to perform
task placement and scaling. The reward
function~\eqref{eq:reward} is funded on the
average processing latency -- see the
$d_v$ term defined in~\eqref{eq:avg-delay}.
However, C-V2N services required that the
target delay is met the $\kappa$-percent
of the times.

% We get a bound K to keep generic expression
To that end, in~\eqref{eq:bound} we propose
a bound for $d_{p,v}^\kappa$
that relates the average
and $\kappa$ delay percentile.
Specifically, the bound $\overline{d_{p,v}^\kappa}$
uses $K(\kappa)$, a quantity that relates
the average and $\kappa$-percentile as
$d_{p,v}^\kappa<K(\kappa)\cdot\mathbb{E}[d_{p,v}]$.
To find $K(\kappa)$ we use the discrete event
simulator CIW~\cite{ciw} as follows.

\begin{figure}[t]
    \centering
    \input{Figures/compare.tex}
    \vspace{-1.5em}
    \caption{Ratio of M/D/1-PS percentile latency 
    $d^\kappa_{p,v}$ and 
    M/G/1-PS average latency
    $\mathbb{E}[d_{p,v}]$
    as the CPU
    load $\rho$ increases.
    Results are obtained with the CIW
    simulator~\cite{ciw}.}
    \label{fig:compare-deterministic}
\end{figure}

% Service times det => simulations CIW
We run simulations for an M/D/1-PS queue
that captures the deterministic processing
times for video
decoding/recognition~\cite{Alvarez-Mesa2012,SHUSTANOV2017718}
in Advanced Driving tasks~\cite{3GPPv2xrequirements,
tutorial-v2x}.
Fig.~\ref{fig:compare-deterministic}
plots the ratio between the average delay~\eqref{eq:avg-delay}
and the $\kappa$ percentile of the delay
using the service rates $\mu$ reported in
TABLE~\ref{table:processing-rates}.
Results show that the 99-percentile of the
delay is smaller than two times the average delay, i.e. $K(99)=2$.

As a result, the experimental campaign in
Section~\ref{sec:evaluation} resorts to Lemma~\ref{lemma:kappa}
to promote that \gls{dhpg} agents meet the Advanced
Driving requirement (i.e. $d_{tgt}=100$\,ms)
the 99\% of the times.
In particular, \gls{dhpg} is trained with $d_{tgt}$ replaced by $\frac{d_{tgt}}{K(99)}=50$\,ms in
the reward function~\eqref{eq:reward}. Consequently, \gls{dhpg} demonstrably meets the 100\,ms delay 99\% of the time.

\begin{IEEEbiography}[{\includegraphics[width=1in,height=1.25in,clip,keepaspectratio]{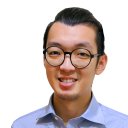}}]{Cyril Shih-Huan Hsu}
% is a Ph.D. candidate in the Informatics Institute, University of Amsterdam (UvA), The Netherlands, since 2021. He received the B.Sc. and M.Sc. degrees from National Taiwan University (NTU), in 2013 and 2015, respectively. From 2016 to 2021, he worked for several international AI startups as a machine learning researcher. He is currently a member of the Multiscale Networked Systems Group (MNS) that focuses on programmable networks and data-centric automation. His recent research focus on network resource management with AI/ML.
is a Ph.D. candidate at the Informatics Institute, University of Amsterdam (UvA), The Netherlands, where he has been pursuing his degree since 2021. He earned his B.Sc. and M.Sc. degrees from National Taiwan University (NTU) in 2013 and 2015, respectively. From 2016 to 2021, he worked as a machine learning researcher at several international AI startups. He is currently a member of the Multiscale Networked Systems (MNS) Group. His recent research centers on leveraging AI and machine learning for network resource management.
% \vspace{-3mm}
\end{IEEEbiography}
\vskip 0pt plus -1fil
\begin{IEEEbiography}[{\includegraphics[width=1in,height=1.25in,clip,keepaspectratio]{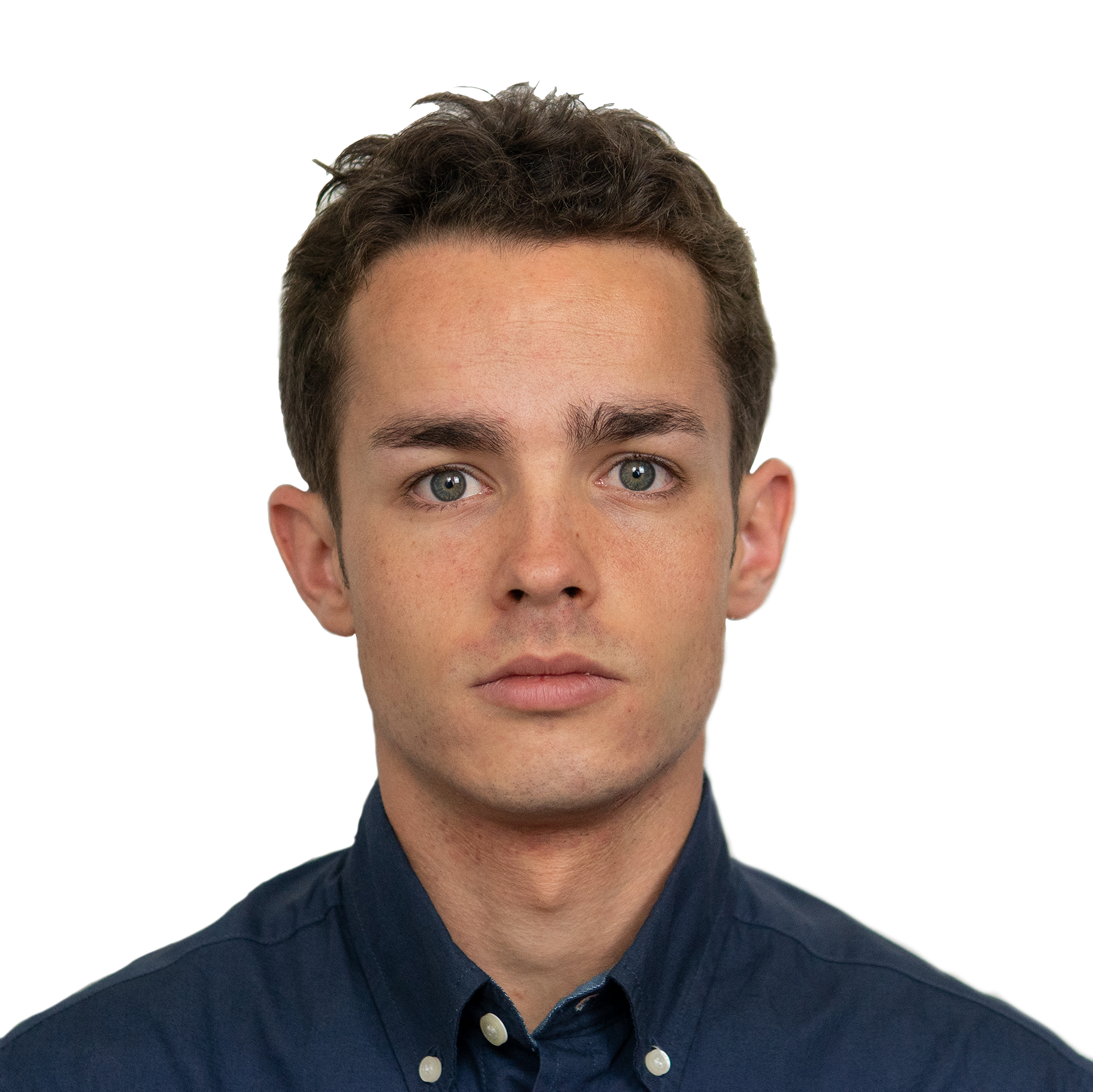}}]{Jorge Mart\'in P\'erez}
is an associate professor at the Universidad
Politécnica de Madrid (UPM), Spain.
He obtained a B.Sc in mathematics,
and a B.Sc in computer science, both
at Universidad Autónoma de Madrid (UAM) in
2016. He obtained his M.Sc. and Ph.D in
Telematics from Universidad Carlos III de
Madrid (UC3M) in 2017 and 2021, respectively.
Jorge worked as postdoc at UC3M (until 2023)
in national and EU funded projects.
His research focuses in optimal resource
allocation in networks.
% \vspace{-3mm}
\end{IEEEbiography}
\vskip 0pt plus -1fil

\begin{IEEEbiography}
[{\includegraphics[width=1in,height=1.25in,clip,keepaspectratio]{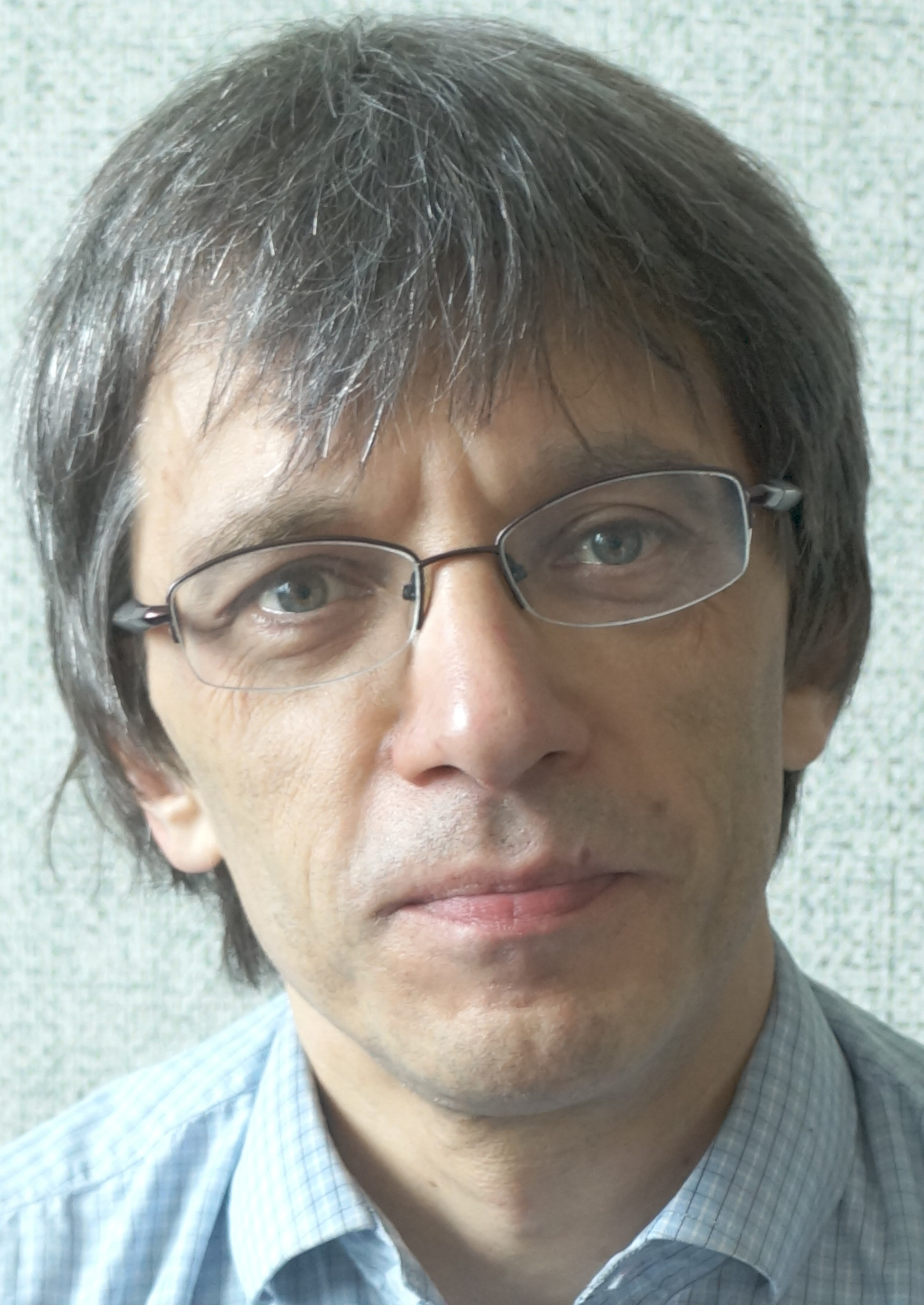}}]
{Danny De Vleeschauwer}
obtained an MSc. in Electrical Engineering and the Ph.D. degree in applied sciences from the Ghent University, Belgium, in 1985 and 1993 respectively. Currently, he is a DMTS in the access network control department of Nokia Bell Labs in Antwerp, Belgium. He was a researcher at Ghent University before joining Nokia. His early work was on image processing and on the application of queuing theory in packet-based networks. His current research focus is on the distributed control of applications over packet-based networks.
% \vspace{-3mm}
\end{IEEEbiography}
\vskip 0pt plus -1fil

%\begin{IEEEbiography}
%[{\includegraphics[width=1in,height=1.25in,clip,keepaspectratio]{img/authors/kote.jpg}}]
%{Koteswararao Kondepu}
%is an Assistant Professor at India Institute of Technology Dharwad, Dharwad, India. He obtained his Ph.D. degree in Computer Science and Engineering from Institute for Advanced Studies Lucca (IMT), Italy in July 2012. His research interests are 5G, optical networks design, energy-efficient schemes in communication networks, and sparse sensor networks.
% \vspace{-8mm}
%\end{IEEEbiography}
%\vskip 0pt plus -1fil

\begin{IEEEbiography}
[{\includegraphics[width=1in,height=1.25in,clip,keepaspectratio]{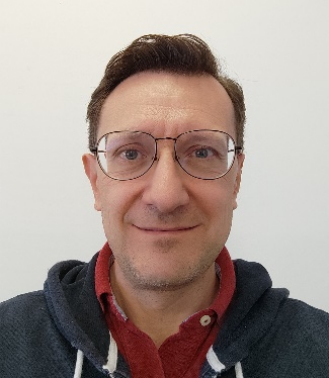}}]
{Luca Valcarenghi}
is an associate professor at the Scuola Superiore Sant'Anna of Pisa, Italy, since 2014. He published almost three hundred papers (source Google Scholar, May 2020) in International Journals and Conference Proceedings. He received a Fulbright Research Scholar Fellowship in 2009 and a JSPS "Invitation Fellowship Program for Research in Japan (Long Term)" in 2013. His main research interests are optical networks design, analysis and optimization; communication networks reliability; energy efficiency in communications networks; optical access networks; zero touch network and service management; experiential networked intelligence; 5G technologies and beyond.
% \vspace{-3mm}
\end{IEEEbiography}
\vskip 0pt plus -1fil
% Xi Li
\begin{IEEEbiography}[{\includegraphics[width=1in,height=1.25in,clip,keepaspectratio]{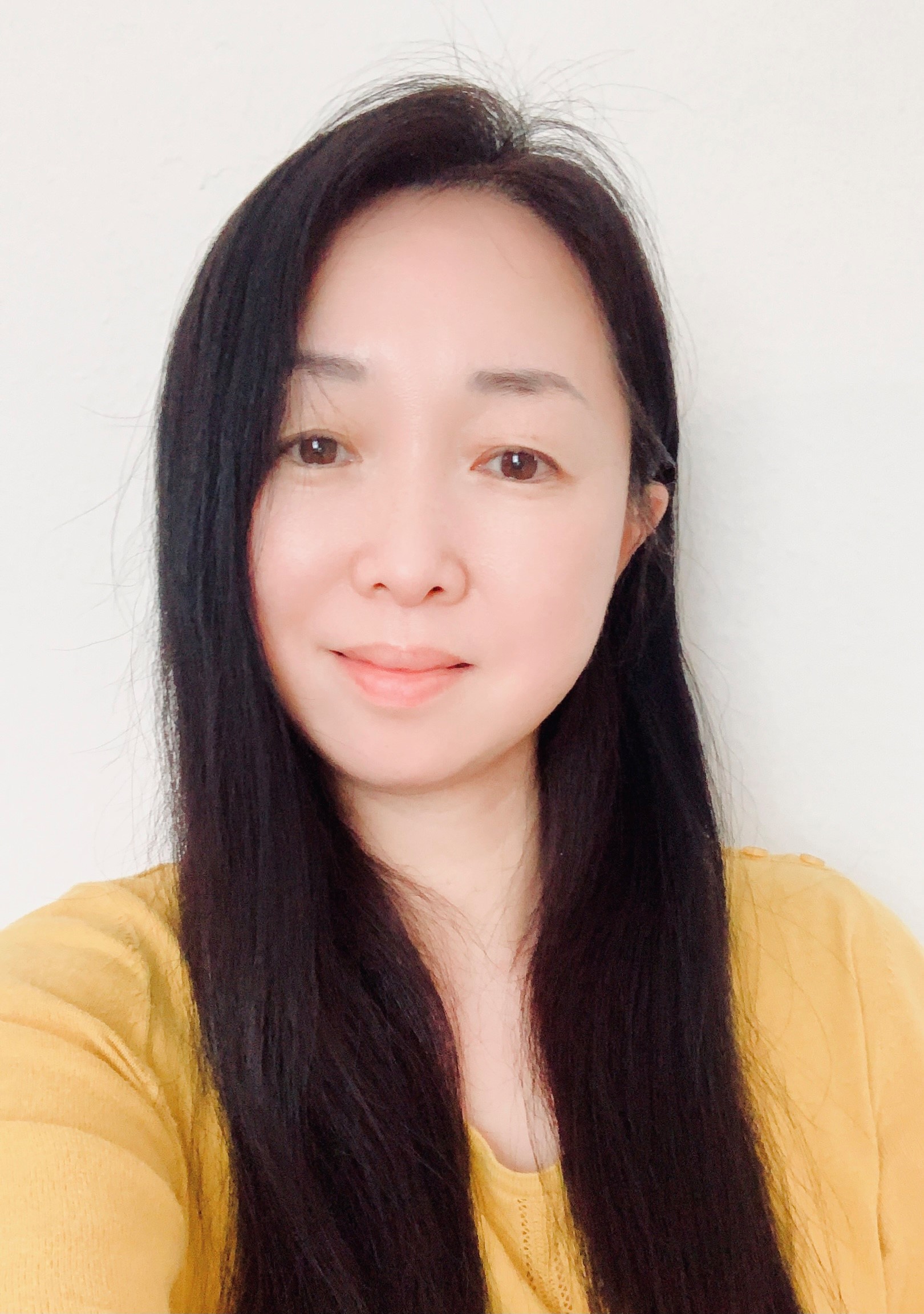}}]{Xi Li} received her M.Sc in electronics and telecommunication engineering from Technische Universität Dresden in 2002 and her Ph.D. from the University of Bremen in 2009. Between 2003 and 2014 she worked as a research fellow and lecturer at the Communication Networks Group in the University of Bremen, leading a team working on several industrial and European R\&D projects on 3G/4G mobile networks and future Internet design. From 2014 to 2015 she worked as a solution designer in Tele-fonica Germany GmbH \& Co. OHG, Hamburg. Since March 2015 she has been a senior researcher in 5G Networks R\&D at NEC Laboratories Europe.
% \vspace{-3mm}
\end{IEEEbiography}
\vskip 0pt plus -1fil
\begin{IEEEbiography}
[{\includegraphics[width=1in,height=1.25in,clip,keepaspectratio]{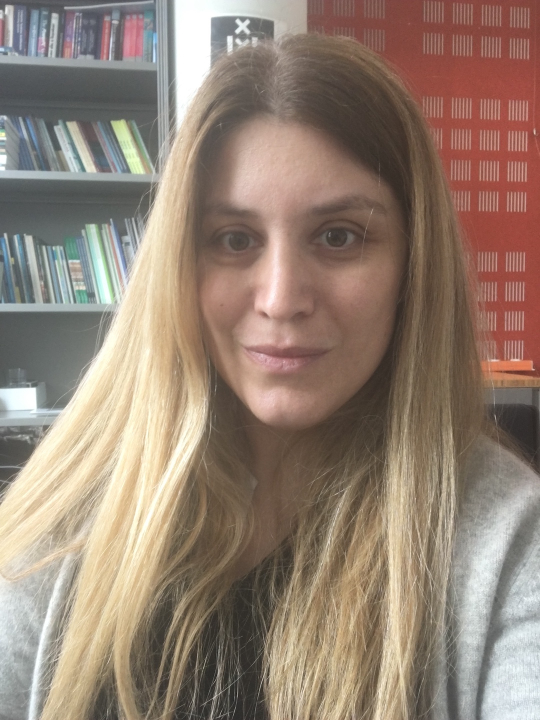}}]
{\textbf{Chrysa Papagianni}} is an associate professor at the Informatics Institute of the University of Amsterdam. She is part of the Multiscale Networked Systems group that focuses its research on network programmability and data-centric automation. Her research interests lie primarily in the area of programmable networks with emphasis on network optimization and the use of machine learning in networking. She has worked in multiple research projects funded by the European Commission, the Dutch research council and the European Space Agency, such as 5Growth, CATRIN, etc. Chrysa is currently coordinating the SNS JU DESIRE6G project on 6G system architecture. She also leads the AI-assisted networking work package in the 6G flagship project for the Netherlands on Future Network Services.
\end{IEEEbiography}

%%%%%%%%%% bib %%%%%%%%%%%%%%%%%%%
%\end{IEEEbiography}


%\newpage



%\vfill


% \clearpage

% \appendices
% \input{Appendix/np-hard-proof}
% \input{Appendix/deterministic}

\end{document}